\documentclass{article}
\usepackage[a4paper, total={7in, 9in}]{geometry}
\usepackage[english]{babel}
\usepackage[utf8]{inputenc}
\usepackage{amsmath}
\usepackage{graphicx}
\usepackage{authblk}
\usepackage[inline]{enumitem}
\usepackage[colorinlistoftodos]{todonotes}
\usepackage[font=small]{caption}
\usepackage[labelformat=empty, position=top]{subcaption}
\usepackage[export]{adjustbox}
\usepackage{hyperref}
\usepackage{isomath}
\usepackage[superscript,biblabel]{cite}
\usepackage{csquotes}
\usepackage{hyphenat}
\usepackage[ruled,vlined]{algorithm2e}

\SetCommentSty{mycommfont}

\title{Emergent Analogical Reasoning in Large Language Models\footnotetext{Published at \textit{Nature Human Behaviour} (2023) \href{https://doi.org/10.1038/s41562-023-01659-w}{https://doi.org/10.1038/s41562-023-01659-w}}}

\author[1,*]{Taylor Webb}
\author[1]{Keith J. Holyoak}
\author[1,2]{Hongjing Lu}
\affil[1]{Department of Psychology}
\affil[2]{Department of Statistics\protect\\University of California, Los Angeles, CA, USA}
\affil[*]{Correspondence to: taylor.w.webb@gmail.com}

\date{}

\begin{document}
\maketitle

\begin{abstract}

The recent advent of large language models has reinvigorated debate over whether human cognitive capacities might emerge in such generic models given sufficient training data. Of particular interest is the ability of these models to reason about novel problems \textit{zero-shot}, without any direct training. In human cognition, this capacity is closely tied to an ability to reason by analogy. Here, we performed a direct comparison between human reasoners and a large language model (the text-davinci-003 variant of GPT-3) on a range of analogical tasks, including a non-visual matrix reasoning task based on the rule structure of Raven's Standard Progressive Matrices. We found that GPT-3 displayed a surprisingly strong capacity for abstract pattern induction, matching or even surpassing human capabilities in most settings; preliminary tests of GPT-4 indicated even better performance. Our results indicate that large language models such as GPT-3 have acquired an emergent ability to find zero-shot solutions to a broad range of analogy problems. 

\end{abstract}

\section{Introduction}

Analogical reasoning is at the heart of human intelligence and creativity. When confronted with an unfamiliar problem, human reasoners can often identify a reasonable solution through a process of structured comparison to a more familiar situation~\cite{holyoak2012analogy}. This process is an essential part of human reasoning in domains ranging from everyday problem-solving~\cite{bassok2012problem} to creative thought and scientific innovation~\cite{dunbar2012scientific}. Indeed, tests of analogical reasoning ability are uniquely effective as measures of fluid intelligence: the capacity to reason about novel problems~\cite{cattell1971abilities,snow1984topography}.

Recently, there has been considerable debate about whether and how a capacity for analogical thought might be captured in deep learning systems~\cite{mitchell2021abstraction}. Much of this recent work has focused on training neural networks on very large datasets (sometimes containing millions of problems)~\cite{barrett2018measuring,zhang2019raven}. Though this is a challenging task that has spurred the development of some interesting approaches~\cite{hill2019learning,wu2020scattering,hersche2023neuro,subhramondal2023objects}, it does not address the issue of whether analogical reasoning can emerge \textit{zero-shot} (i.e., without direct training), the capacity most central to human thought.

An alternative approach, also based on deep learning, involves  large language models (LLMs)~\cite{brown2020language}. LLMs have recently sparked great interest (and controversy) for their potential to perform few-shot, and even zero-shot, reasoning. These models employ relatively generic neural network architectures with up to billions of parameters, and are trained using a simple predictive objective (predicting the next token in a sequence of text) with massive web-based text corpora consisting of billions of tokens. Though there is significant debate about the capabilities of these models~\cite{mahowald2023dissociating}, a potential advantage is their ability to solve problems with little direct training, sometimes requiring only a few examples, or even a simple task instruction (typically without any updating of model parameters). This feature raises the question of whether LLMs might be capable of human-like, zero-shot analogical reasoning.

\captionsetup{labelfont=bf,font=small}
\begin{figure}[h!]
\centering
\begin{subfigure}[t]{.53\linewidth}\vskip 0pt
    \includegraphics[width=\linewidth]{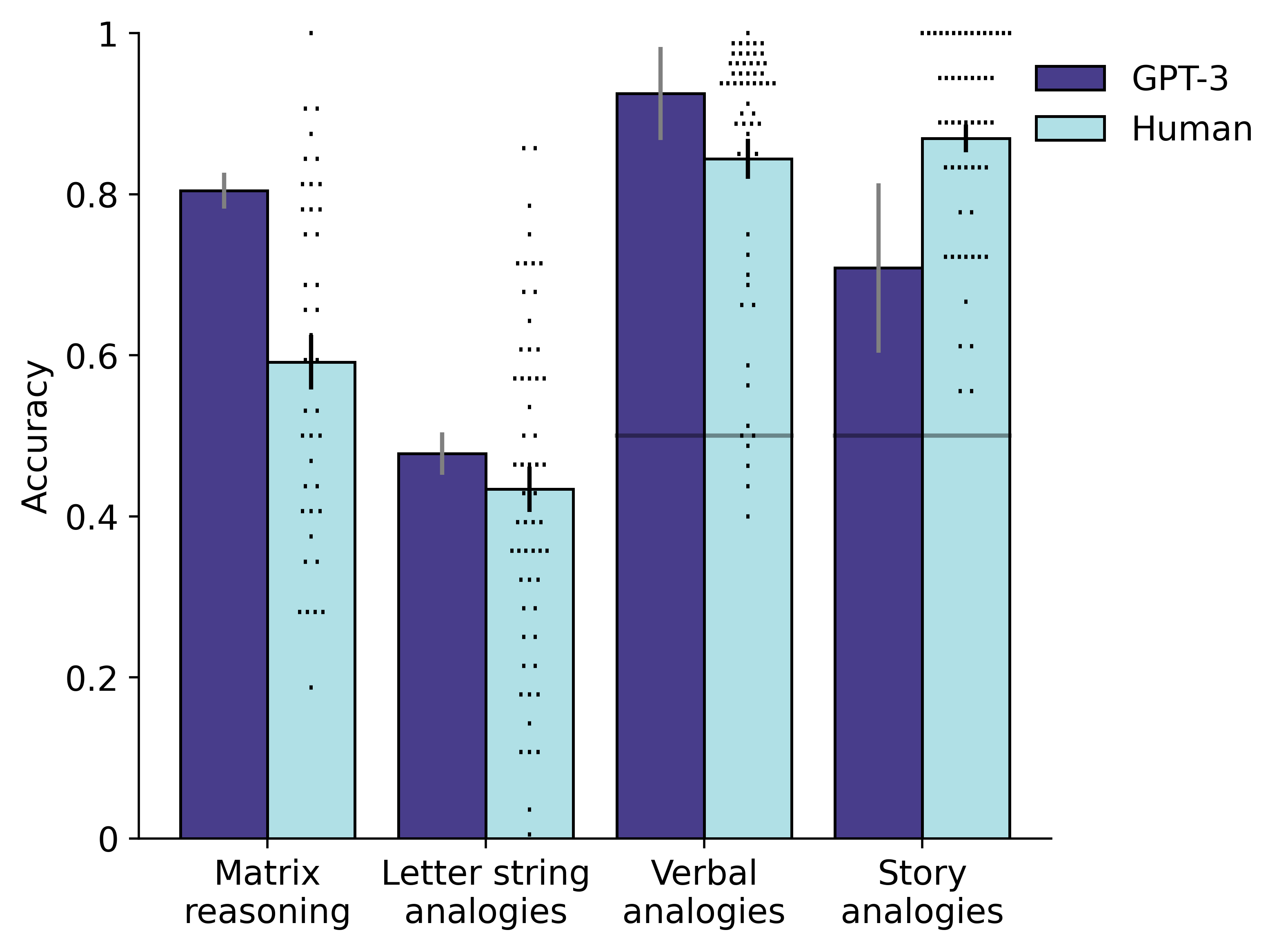} 
    \subcaption{}
\end{subfigure}
\caption{\textbf{Summary of results.} Matrix reasoning results show average accuracy on all problems in Digit Matrices problem set, a novel text-based matrix reasoning task designed to emulate Raven’s Standard Progressive Matrices (SPM) problems~\cite{raven1938raven}. Note that the Digit Matrices were purely text-based, and therefore do not test for the ability to perform abstract reasoning directly over visual inputs, as in the original SPM. Letter string results show average performance for novel letter string analogy problem set, based on problems from Hofstadter and Mitchell~\cite{hofstadterMitchell94}. Both matrix reasoning and letter string results reflect performance on generative task. Verbal analogy results show average performance on UCLA Verbal Analogy Test~\cite{lu2019emergence}. Story analogy problems involved identification of analogous stories based on higher-order relations, using materials from Gentner et al.~\cite{gentner1993roles}. Both verbal and story analogy results reflect multiple-choice accuracy, with chance performance indicated by gray horizontal line. Chance performance for the two generative tasks (matrix reasoning and letter string analogies) is close to zero, due to the very large space of possible generative responses. Black error bars represent standard error of the mean for average performance across participants. Each dot represents accuracy for a single participant (matrix reasoning, N=40; letter string analogies, N=57; verbal analogies, N=57; story analogies, N=54). Gray error bars represent 95\% binomial confidence intervals for average performance across multiple problems.}  
\label{results_all_tasks}
\end{figure}

To answer this question, we evaluated the language model GPT-3~\cite{brown2020language} on a range of zero-shot analogy tasks, and performed direct comparisons with human behavior. These tasks included a novel text-based matrix reasoning task based on the rule structure of Raven's Standard Progressive Matrices (SPM)~\cite{raven1938raven}, a visual analogy problem set commonly viewed as one of the best measures of fluid intelligence~\cite{snow1984topography}. Unlike the original visual SPM problems, our Digit Matrices task was purely text-based so that it could be used to evaluate GPT-3's ability to induce abstract rules (though not the ability to do so directly from visual inputs). Strikingly, we found that GPT-3 performed as well or better than college students in most conditions, despite receiving no direct training on this task. GPT-3 also displayed strong zero-shot performance on letter string analogies~\cite{hofstadterMitchell94}, four-term verbal analogies~\cite{sternberg1980developmental,turney2003combining,lu2019emergence,jones2022differential}, and identification of analogies between stories~\cite{gick1980analogical,holyoak1984development, gentner1993roles}. These results add to the growing body of work characterizing the emergent capabilities of LLMs~\cite{dasgupta2022language,srivastava2022beyond,wei2022emergent,chan2022data,binz2023using}, and suggest that the most sophisticated LLMs may already possess an emergent capacity to reason by analogy.

\section{Results}

We evaluated the language model GPT-3 on a set of analogy tasks, and compared its performance to human behavior. GPT-3 is a large-scale (175B parameters), transformer-based~\cite{vaswani2017attention} language model developed by OpenAI~\cite{brown2020language}. The original base model was trained on a web-based corpus of natural language consisting of over 400 billion tokens, using a training objective based on next-token prediction (given a string of text, the model is trained to predict the token most likely appear next). A number of variants on this base model have since been developed by fine-tuning it in various ways. These include training the model to generate code~\cite{chen2021evaluating}, and training it to respond appropriately to human prompts, using either supervised learning or reinforcement learning from human feedback (RLHF)~\cite{ouyang2022training}. Our evaluation focused on the most recent model variant, text-davinci-003 (here referred to simply as `GPT-3'), which was the first to incorporate RLHF (along with the concurrently released, but distinct, ChatGPT model). We found that text-davinci-003 displayed particularly strong performance on our analogy tasks, but earlier model variants also performed well in some task settings, suggesting that multiple factors contributed to text-davinci-003's analogical capabilities (Supplementary Figures~\ref{digit_mat_all_GPT}-~\ref{UCLA_VAT_all_GPT}). See Section~\ref{GPT3_variant_discussion} for further discussion.

Our evaluation featured four separate task domains, each designed to probe different aspects of analogical reasoning: 1) text-based matrix reasoning problems, 2) letter-string analogies, 3) four-term verbal analogies, and 4) story analogies. For each task domain, we performed a direct comparison with human behavior, assessing both overall performance and error patterns across a range of conditions relevant to human analogical reasoning. Figure~\ref{results_all_tasks} shows a summary of these results. We also performed a qualitative analysis of GPT-3's ability to use analogical reasoning to solve problems.

\captionsetup{labelfont=bf,font=small}
\begin{figure}[!tb]
\centering
\begin{subfigure}[t]{0.02\textwidth}
    \textbf{\fontfamily{phv}\selectfont{a}}
\end{subfigure}
\begin{subfigure}[t]{.3\linewidth}\vskip 0pt
    \includegraphics[width=\linewidth]{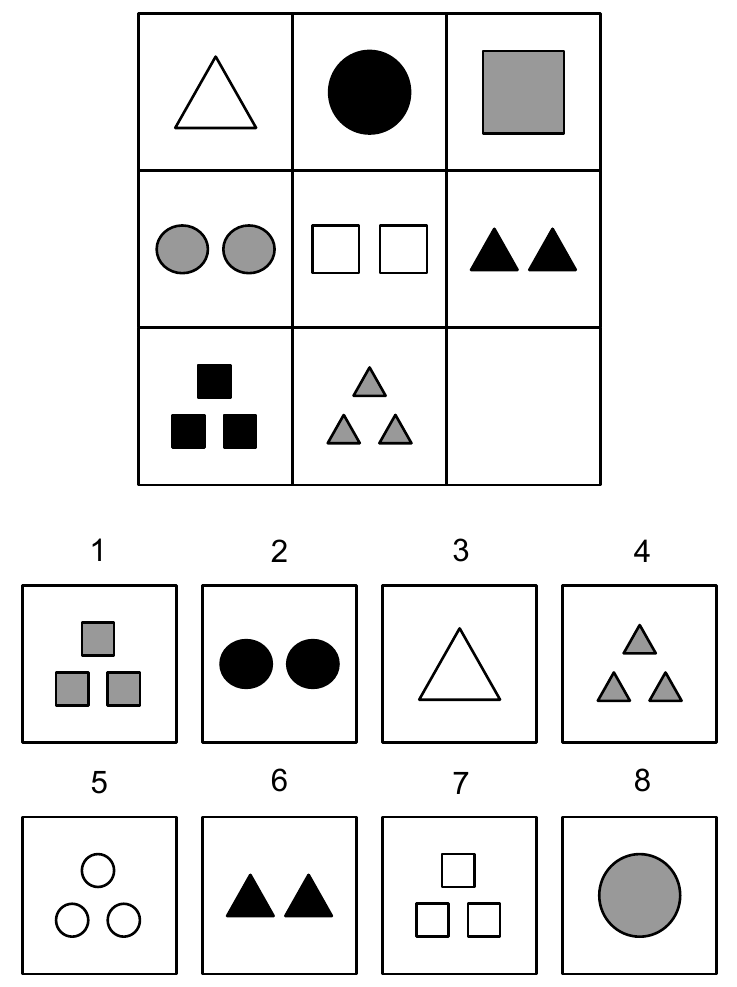} 
    \subcaption{}
    \label{RPM_example}
\end{subfigure}
\begin{subfigure}[t]{0.02\textwidth}
    \textbf{\fontfamily{phv}\selectfont{b}}
\end{subfigure}
\begin{subfigure}[t]{.45\linewidth}\vskip 0pt
    \includegraphics[width=\linewidth]{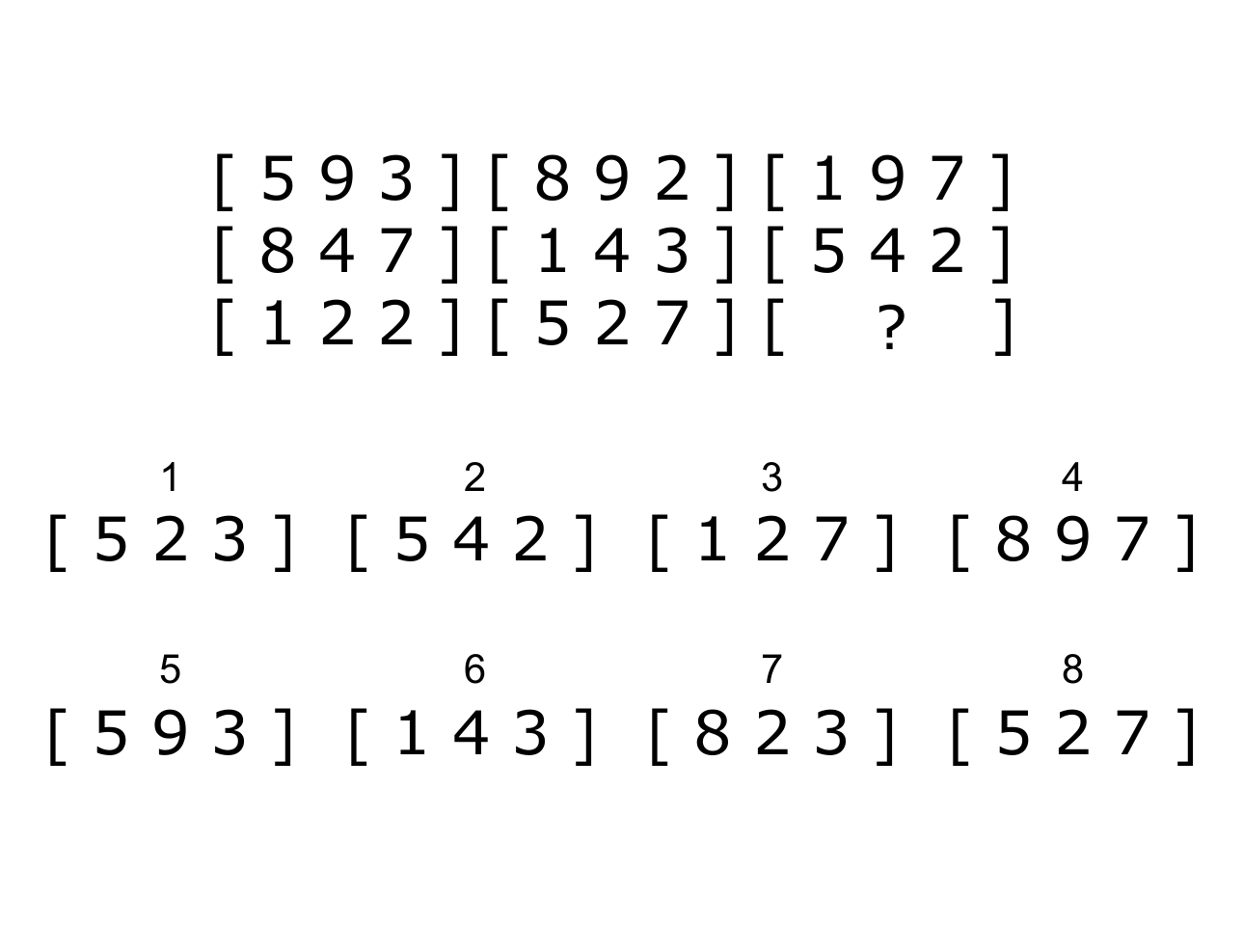} 
    \subcaption{}
    \label{digit_mat_example}
\end{subfigure}
\\
\begin{subfigure}[t]{0.02\textwidth}
    \textbf{\fontfamily{phv}\selectfont{c}}
\end{subfigure}
\begin{subfigure}[t]{.2\linewidth}\vskip 0pt
    \includegraphics[width=\linewidth]{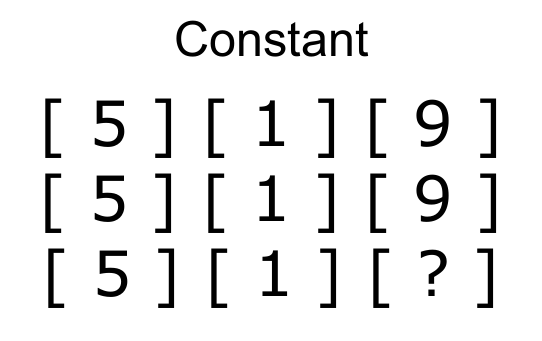} 
    \subcaption{}
    \label{constant}
\end{subfigure}
\begin{subfigure}[t]{0.02\textwidth}
    \textbf{\fontfamily{phv}\selectfont{d}}
\end{subfigure}
\begin{subfigure}[t]{.2\linewidth}\vskip 0pt
    \includegraphics[width=\linewidth]{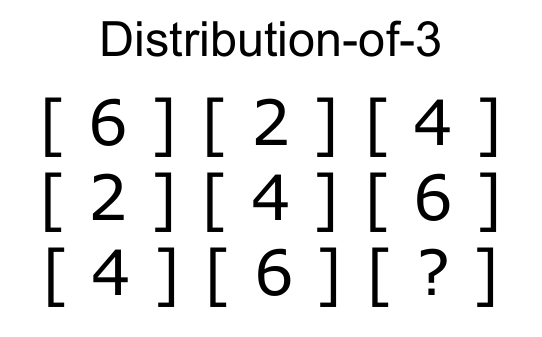} 
    \subcaption{}
    \label{distribution}
\end{subfigure}
\begin{subfigure}[t]{0.02\textwidth}
    \textbf{\fontfamily{phv}\selectfont{e}}
\end{subfigure}
\begin{subfigure}[t]{.2\linewidth}\vskip 0pt
    \includegraphics[width=\linewidth]{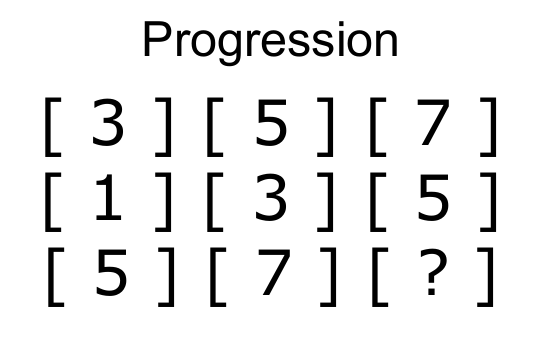} 
    \subcaption{}
    \label{progression}
\end{subfigure}
\\
\begin{subfigure}[t]{0.02\textwidth}
    \textbf{\fontfamily{phv}\selectfont{f}}
\end{subfigure}
\begin{subfigure}[t]{.325\linewidth}\vskip 0pt
    \includegraphics[width=\linewidth]{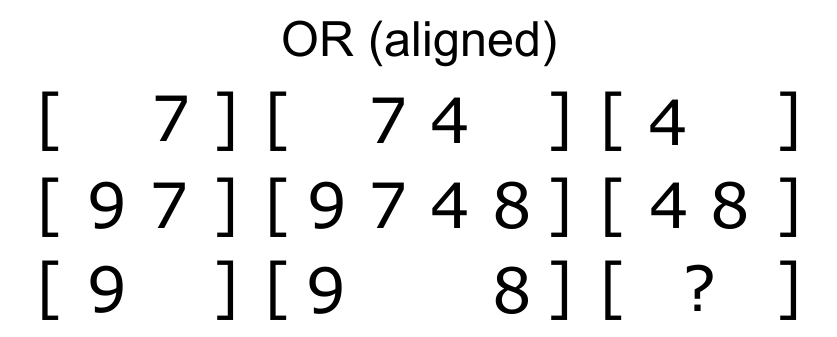} 
    \subcaption{}
    \label{OR_aligned}
\end{subfigure}
\begin{subfigure}[t]{0.02\textwidth}
    \textbf{\fontfamily{phv}\selectfont{g}}
\end{subfigure}
\begin{subfigure}[t]{.325\linewidth}\vskip 0pt
    \includegraphics[width=\linewidth]{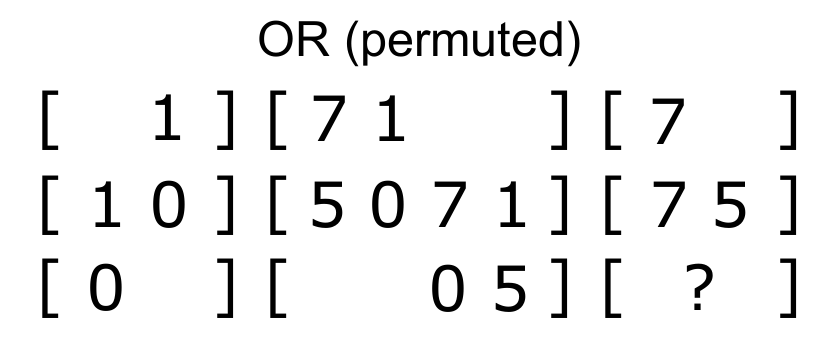} 
    \subcaption{}
    \label{OR_permuted}
\end{subfigure}
\caption{\textbf{Matrix reasoning problems.} \textbf{(a)} Example problem depicting structure of Raven's Progressive Matrices~\cite{raven1938raven}. Problems consist of a $3\times3$ matrix populated with geometric forms, in which each row or column is governed by the same set of abstract rules. Problem solvers must identify these rules, and use them to infer the missing cell in the lower right, by selecting from the set of 8 choices below. \textbf{(b)} Example problem illustrating the novel Digit Matrices problem set. Problems consist of a $3\times3$ matrix, in which each cell is demarcated by brackets, and populated by digits. The problems are governed by the same rule structure as Raven's Standard Progressive Matrices. The example problems in (a) and (b) are structurally isomorphic (i.e., governed by the same set of rules). The reader is encouraged to derive the solution to each problem. The solutions to both problems are given in Supplementary Section~\ref{example_mat_solutions}. Problems were governed either by one or more \textit{transformation} rules (b-e), or by a single \textit{logic} rule (f,g). \textbf{(c)} \textit{Constant} rule: same digit appears across either rows or columns. \textbf{(d)} \textit{Distribution-of-3} rule: same set of 3 digits appears in each row or column, but with order varied. \textbf{(e)} \textit{Progression} rule: digits either increase or decrease, by values of 1 or 2, across rows or columns. In the example shown here, digits increase by 2 across rows. \textbf{(f)} \textit{OR} rule: the set of digits present in a particular row or column are defined as the union of the sets present in the other rows or columns. In the illustrated example, the digits in the second column are formed from the union of the sets in the first and third columns. This example illustrates how the spatial alignment of the corresponding elements can make it easier to intuitively grasp the underlying rule. \textbf{(g)} More challenging logic problem governed by same rule (OR), but in which the corresponding elements are spatially permuted. Other logic problems were governed either by an \textit{AND} rule or an \textit{XOR} rule (not pictured).} 
\label{RPM_vs_digit_mat}
\end{figure}

\captionsetup{labelfont=bf,font=small}
\begin{figure}[!tb]
\centering
\begin{subfigure}[t]{0.02\textwidth}
    \textbf{\fontfamily{phv}\selectfont{a}}
\end{subfigure}
\begin{subfigure}[t]{.4\linewidth}\vskip 0pt
    \includegraphics[width=\linewidth]{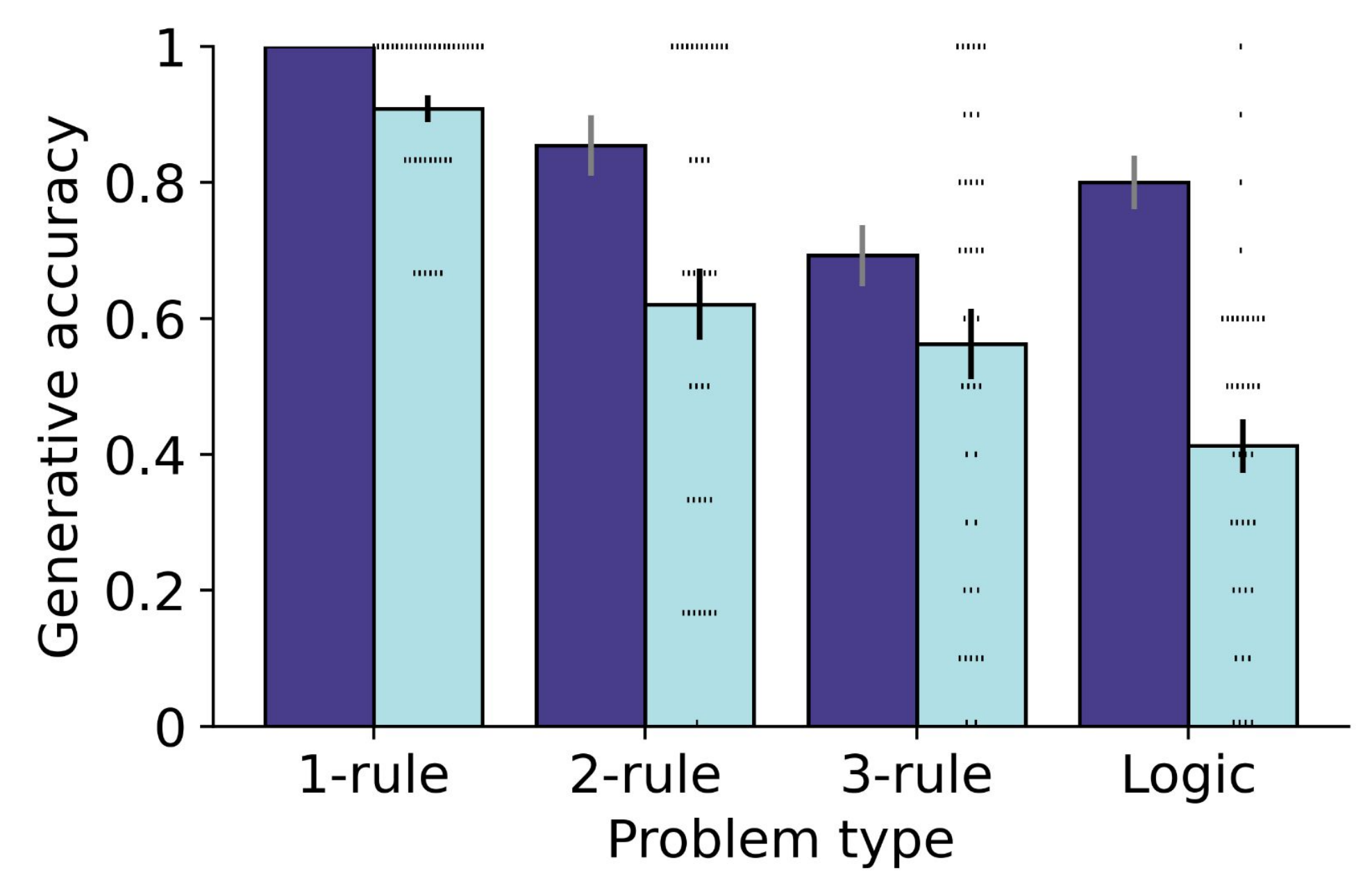} 
    \subcaption{}
    \label{exp1_gen_overall}
\end{subfigure}
\begin{subfigure}[t]{0.02\textwidth}
    \textbf{\fontfamily{phv}\selectfont{b}}
\end{subfigure}
\begin{subfigure}[t]{.5\linewidth}\vskip 0pt
    \includegraphics[width=\linewidth]{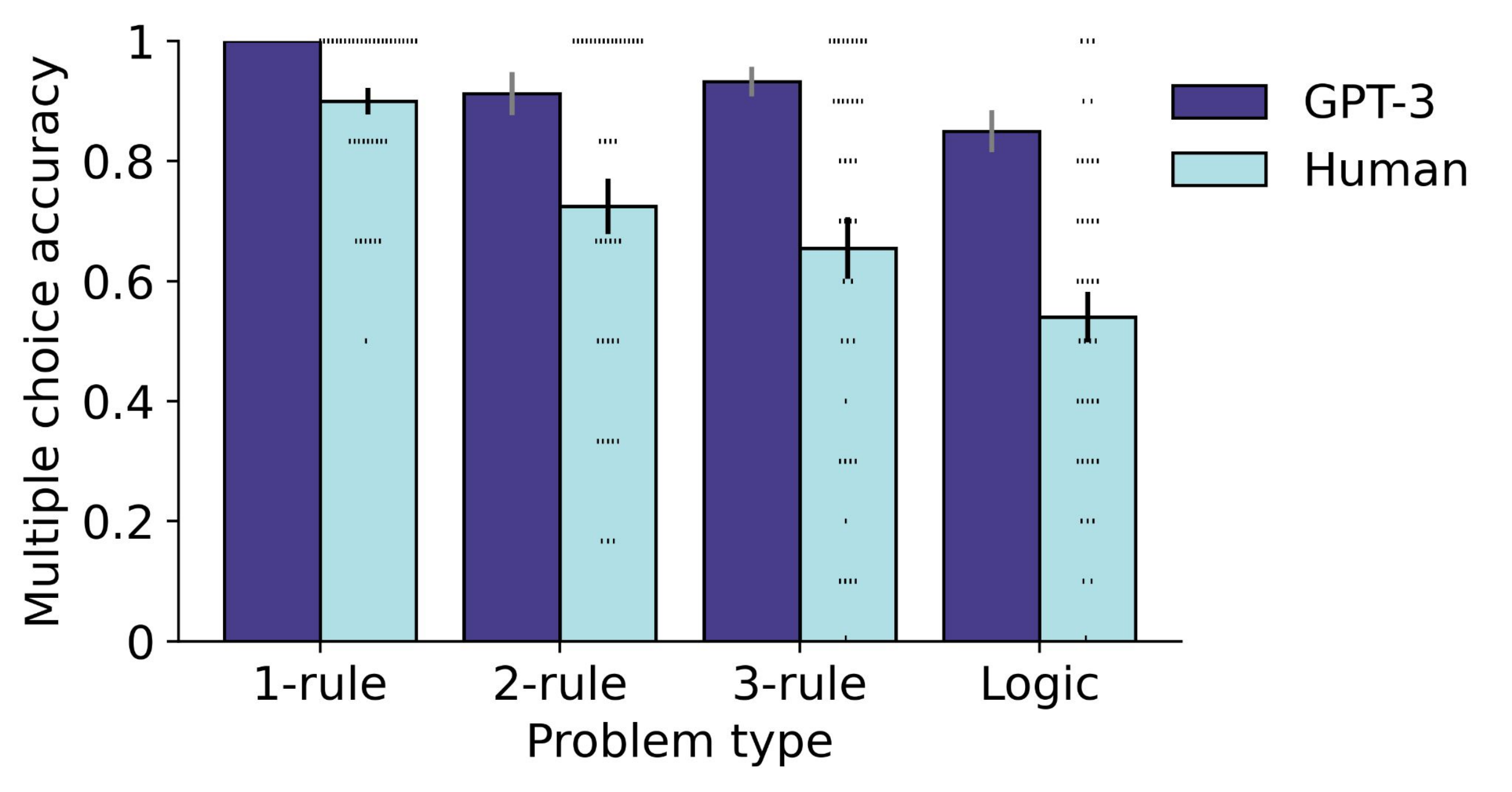} 
    \subcaption{}
    \label{exp1_MC_overall}
\end{subfigure}
\\
\begin{subfigure}[t]{0.02\textwidth}
    \textbf{\fontfamily{phv}\selectfont{c}}
\end{subfigure}
\begin{subfigure}[t]{.25\linewidth}\vskip 0pt
    \includegraphics[width=\linewidth]{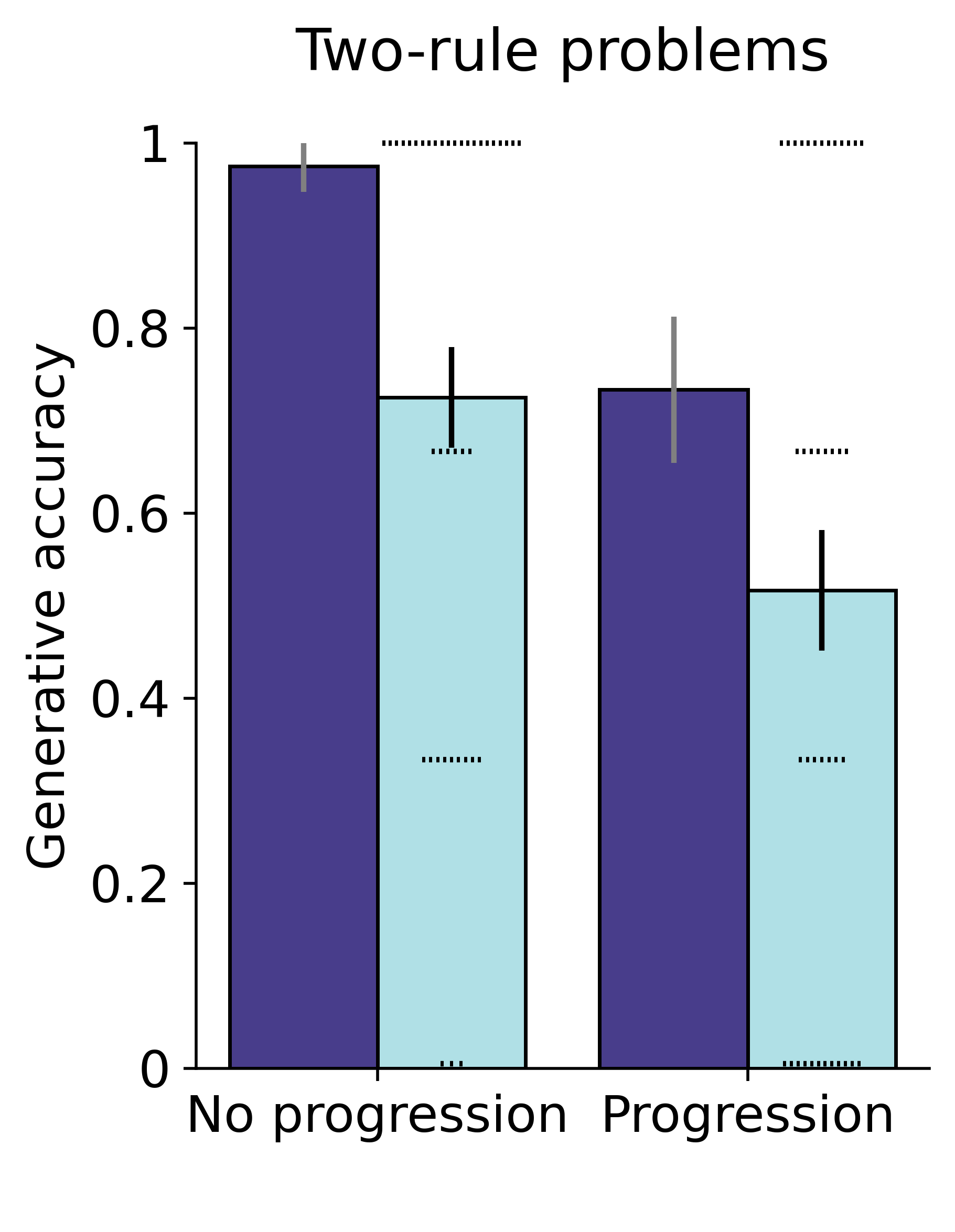} 
    \subcaption{}
    \label{exp1_gen_tworule}
\end{subfigure}
\begin{subfigure}[t]{0.02\textwidth}
    \textbf{\fontfamily{phv}\selectfont{d}}
\end{subfigure}
\begin{subfigure}[t]{.357\linewidth}\vskip 0pt
    \includegraphics[width=\linewidth]{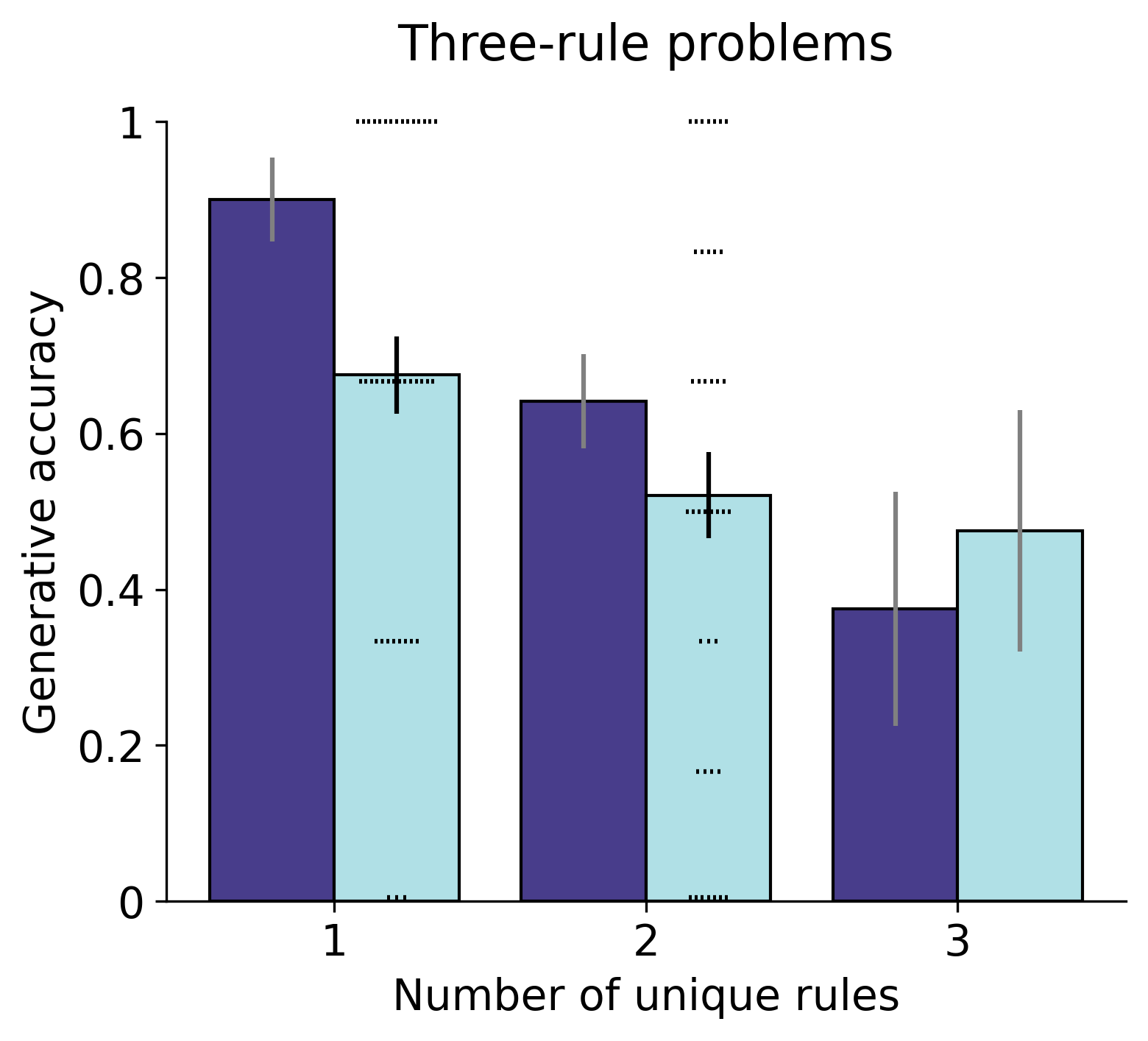} 
    \subcaption{}
    \label{exp1_gen_threerule}
\end{subfigure}
\begin{subfigure}[t]{0.02\textwidth}
    \textbf{\fontfamily{phv}\selectfont{e}}
\end{subfigure}
\begin{subfigure}[t]{.25\linewidth}\vskip 0pt
    \includegraphics[width=\linewidth]{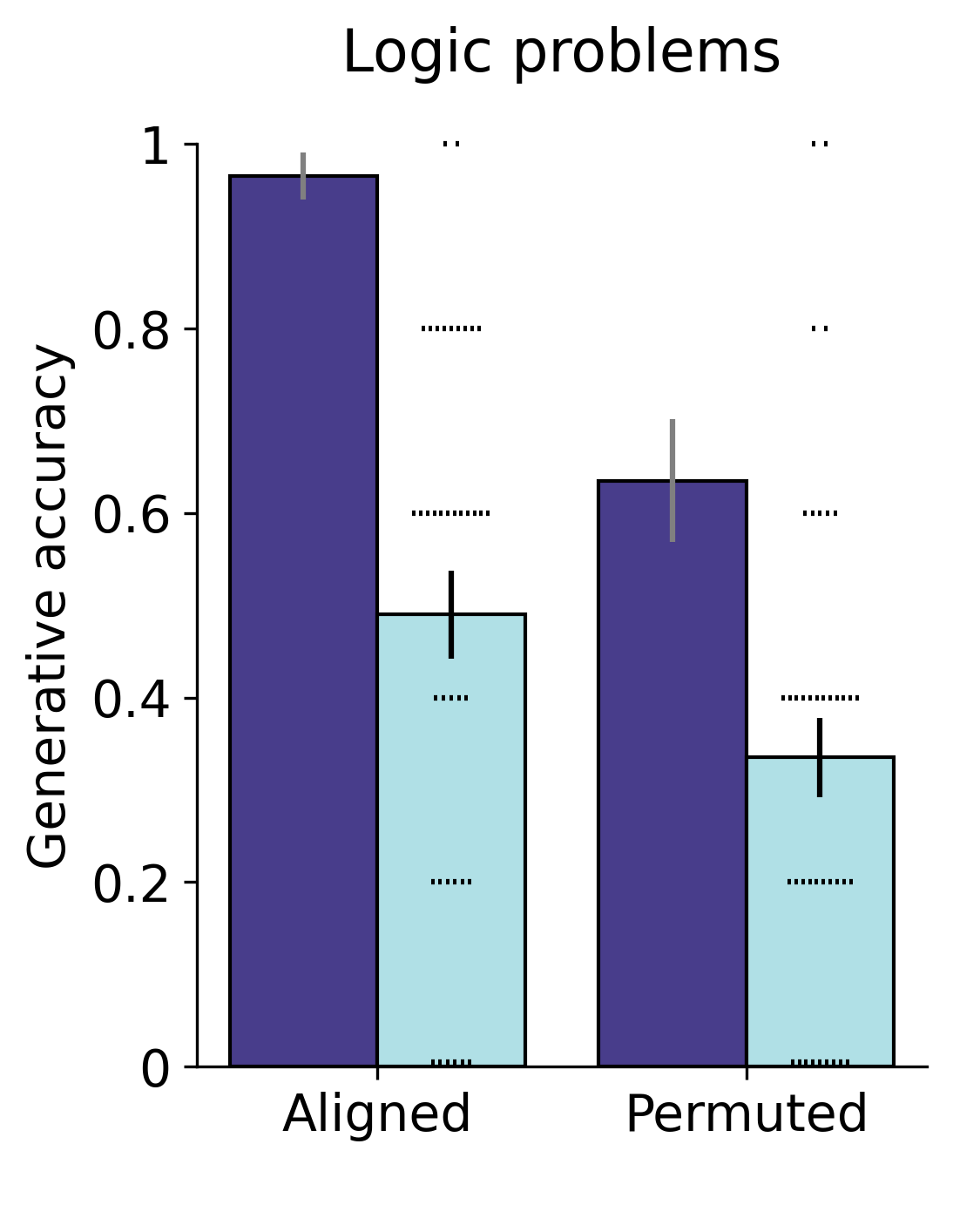} 
    \subcaption{}
    \label{exp1_gen_logic}
\end{subfigure}
\caption{\textbf{Matrix reasoning results.} GPT-3 matched or exceeded human performance for zero-shot Digit Matrices. \textbf{(a)} Generative accuracy for major problem types, including transformation problems with between one and three rules, and logic problems. \textbf{(b)} Multiple-choice accuracy for major problem types. \textbf{(c)} Two-rule problems with at least one progression rule were more difficult than those without. \textbf{(d)} For three-rule problems, performance was a function of the number of unique rules. \textbf{(e)} Spatially permuted logic problems were more difficult than spatially aligned problems. Human results reflect average performance for N=40 participants (UCLA undergraduates). Black error bars represent standard error of the mean across participants. Each dot represents accuracy for a single participant. Gray error bars represent 95\% binomial confidence intervals for average performance across multiple problems. Note that the rightmost bar in (d) does not show individual scores because each participant only completed a single problem with three unique rules.} 
\label{exp1_results}
\end{figure}

\subsection{Matrix reasoning problems}
\label{digit_mat_section}

We designed a text-based matrix reasoning task, the Digit Matrices, to emulate the structure of Raven's Standard Progressive Matrices (SPM)~\cite{raven1938raven}. The task is illustrated in Figure~\ref{RPM_vs_digit_mat}. The dataset was structured similarly to the work of Matzen et al.~\cite{matzen2010recreating}, who created, and behaviorally validated, a visual matrix reasoning dataset with the same rule structure as the original SPM. The Digit Matrices dataset thus has a similar rule structure to SPM, but is guaranteed to be novel for both humans and LLMs. 

\captionsetup{labelfont=bf,font=small}
\begin{figure}[!tb]
\centering
\begin{subfigure}[t]{.65\linewidth}\vskip 0pt
    \includegraphics[width=\linewidth]{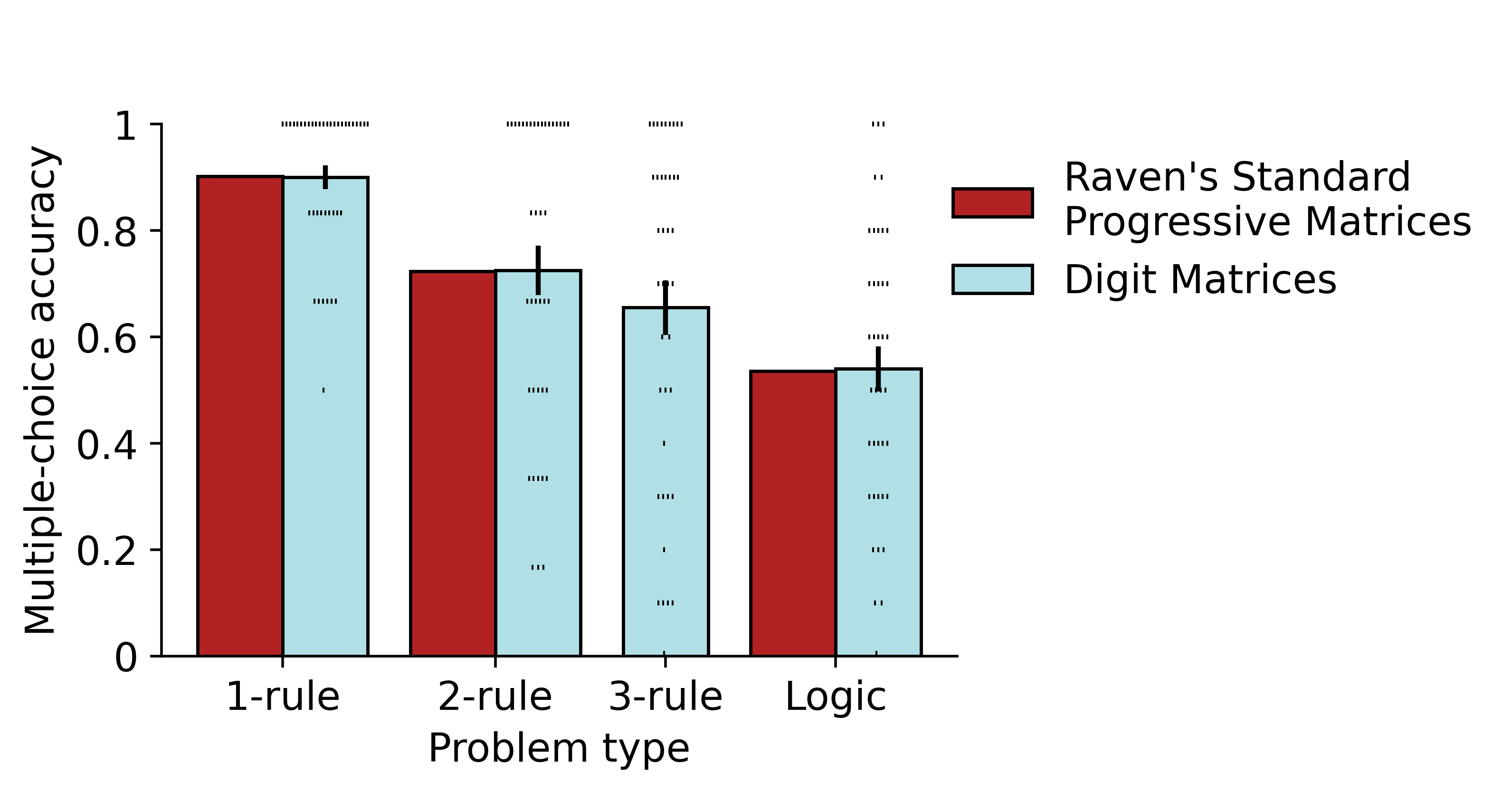} 
    \subcaption{}
\end{subfigure}
\caption{\textbf{Human performance for Digit Matrices vs. Raven's Standard Progressive Matrices (SPM).} SPM~\cite{raven1938raven} does not contain three-rule problems, but performance was very similar across one-rule, two-rule, and logic problems. SPM results reflect average performance for N=80 participants (data from~\cite{matzen2010recreating}). Digit Matrices results reflect average performance for N=40 participants. Error bars represent standard error of the mean. Each dot represents accuracy for a single participant.} 
\label{SPM_vs_digit_mat}
\end{figure}

Digit Matrix problems consisted of either digit transformations (Figures~\ref{digit_mat_example}-~\ref{progression}) or logic problems (Figures~\ref{OR_aligned}-~\ref{OR_permuted}). Transformation problems were defined based on a set of three rule types -- \textit{constant} (Figure~\ref{constant}), \textit{distribution-of-3} (Figure~\ref{distribution}), and \textit{progression} (Figure~\ref{progression}) -- and consisted of one or more rules per problem. When multiple rules were present (Figure~\ref{digit_mat_example}), each rule was bound to a different spatial location within each cell (e.g., one rule was bound to the left digit in each cell, and another rule was bound to the right digit). Logic problems were defined based on set relations -- \textit{OR}, \textit{AND}, and \textit{XOR} -- and involved only a single rule per problem. In some logic problems, the corresponding elements were spatially aligned (Figure~\ref{OR_aligned}), whereas in others they were permuted (Figure~\ref{OR_permuted}). We hypothesized that spatial alignment would be beneficial when solving the problems via analogical mapping, as it should highlight the isomorphism ~\cite{matlen2020spatial}. Digit Matrices problems were presented to GPT-3 without any prompt or in-context task examples.

Figure~\ref{exp1_results} shows zero-shot performance on the Digit Matrices problems for GPT-3 and human participants (N=40, UCLA undergraduates). GPT-3 surpassed the average level of human performance on all problem types, both when generating answers directly (Figure~\ref{exp1_gen_overall}; logistic regression, main effect of GPT-3 vs. human participants: odds ratio (OR) $=1.88$, $p = 0.005$, 95\% confidence intervals (CI) $=[1.21,2.91]$), and when selecting from a set of answer choices (Figure~\ref{exp1_MC_overall}; main effect of GPT-3 vs. human participants: OR $=6.27$, $p = 2.3\times10^{-8}$, CI $=[3.28,11.99]$). It is worth emphasizing, however, that participants displayed a range of performance levels on this task, with some participants outperforming GPT-3 (indeed, the best participant answered every problem correctly). 

In addition to showing strong overall performance, GPT-3's pattern of performance across problem subtypes was similar to that observed in human participants (correlation analysis: $r(30)=0.39$, $p=0.027$). This correlation was driven both by the pattern of performance across major problem types (one-, two-, three-rule, and logic problems; main effect of problem type on generative accuracy: OR $=0.5$, $p = 2\times10^{-16}$, CI $=[0.44,0.56]$,; main effect of problem type on multiple-choice accuracy: OR $=0.56$, $p = 2\times10^{-16}$, CI $=[0.5,0.64]$), and by differences within each problem type. Problems with progression rules were more difficult than those without them (Figure~\ref{exp1_gen_tworule}; main effect of progression vs. no progression, human participants: OR $=0.41$, $p=0.0001$, CI $=[0.24,0.69]$; GPT-3: OR $=0.07$, $p= 1.9\times10^{-5}$, CI $=[0.02,0.24]$); for multi-rule problems, performance was negatively correlated with the number of unique rules in each problem, even when holding constant the number of total rules (Figure~\ref{exp1_gen_threerule}; main effect of number of unique rules, human participants: OR $=0.61$, $p=0.0047$, CI $=[0.44,0.86]$; GPT-3: OR $=0.25$, $p = 3\times10^{-10}$, CI $=[0.17,0.39]$); and logic problems were more difficult when the corresponding elements were spatially permuted vs. aligned (Figure~\ref{exp1_gen_logic}; main effect of spatial alignment, human participants: OR $=0.52$, $p = 0.0017$, CI $=[0.35,0.79]$; GPT-3: OR $=0.06$, $p = 2\times10^{-11}$, CI $=[0.03,0.14]$). These effects replicate well-known characteristics of human analogical reasoning: problems defined by relations (e.g., progression) are typically more difficult than problems defined by the features of individual entities (e.g., constant or distribution-of-3)~\cite{kroger2004varieties,matzen2010recreating}; problem difficulty is typically driven by the degree of relational complexity, as defined by the number of unique relations~\cite{halford1998processing}; and analogical mapping is easier when a greater number of constraints supports the correct mapping (as is the case in the spatially aligned logic problems)~\cite{matlen2020spatial}. GPT-3's pattern of performance thus displayed many of the characteristics of a human-like analogical mapping process. We also found that GPT-3 was sensitive to contextual information in ways that both improved and impaired its performance, similar to human reasoners (Supplementary Figure~\ref{prog_5rule}).

\captionsetup{labelfont=bf,font=small}
\begin{figure}[h!]
\centering
\begin{subfigure}[t]{0.02\textwidth}
    \textbf{\fontfamily{phv}\selectfont{a}}
\end{subfigure}
\begin{subfigure}[t]{.198\linewidth}\vskip 0pt
    \includegraphics[width=\linewidth]{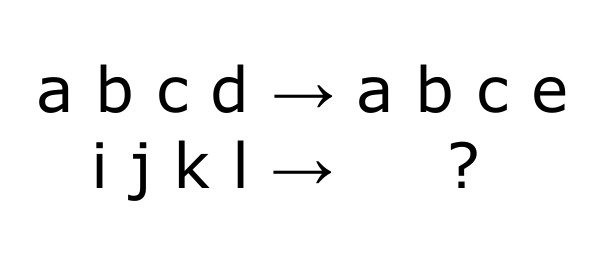} 
    \subcaption{}
    \label{letterstring_zerogen_prob}
\end{subfigure}
\begin{subfigure}[t]{0.02\textwidth}
    \textbf{\fontfamily{phv}\selectfont{b}}
\end{subfigure}
\begin{subfigure}[t]{.333\linewidth}\vskip 0pt
    \includegraphics[width=\linewidth]{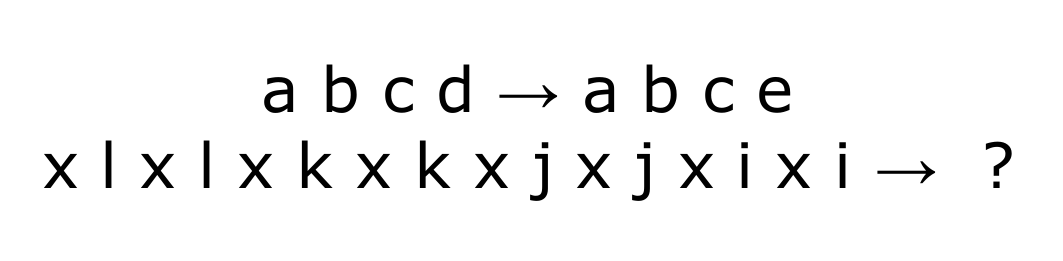} 
    \subcaption{}
    \label{letterstring_threegen_prob}
\end{subfigure}
\begin{subfigure}[t]{0.02\textwidth}
    \textbf{\fontfamily{phv}\selectfont{c}}
\end{subfigure}
\begin{subfigure}[t]{.216\linewidth}\vskip 0pt
    \includegraphics[width=\linewidth]{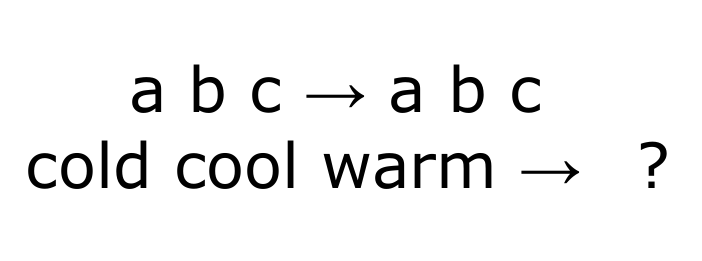} 
    \subcaption{}
    \label{letterstring_realworld_prob}
\end{subfigure}
\\
\begin{subfigure}[t]{0.02\textwidth}
    \textbf{\fontfamily{phv}\selectfont{d}}
\end{subfigure}
\begin{subfigure}[t]{.765\linewidth}\vskip 0pt
    \includegraphics[width=\linewidth]{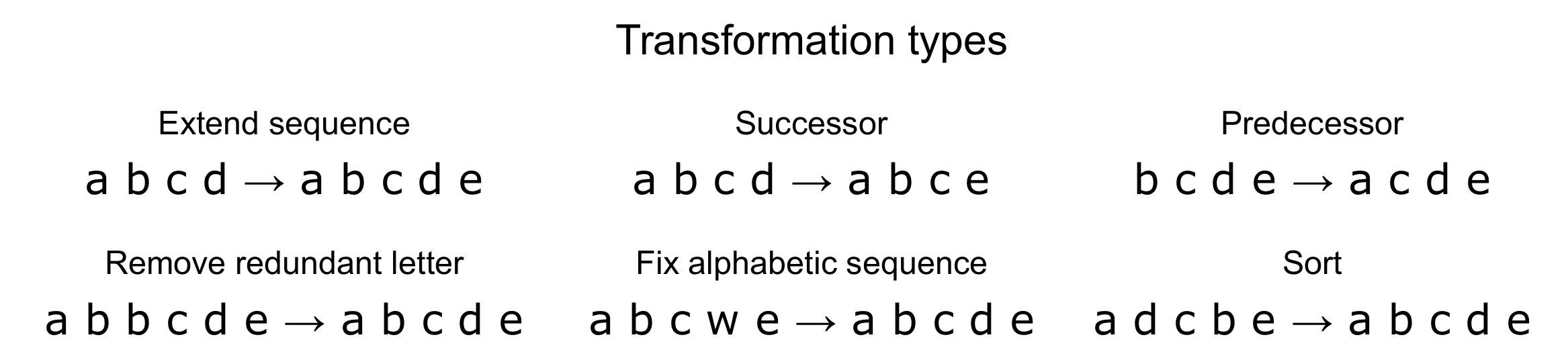}
    \subcaption{}
    \label{letterstring_transformations}
\end{subfigure}
\\
\begin{subfigure}[t]{0.02\textwidth}
    \textbf{\fontfamily{phv}\selectfont{e}}
\end{subfigure}
\begin{subfigure}[t]{.765\linewidth}\vskip 0pt
    \includegraphics[width=\linewidth]{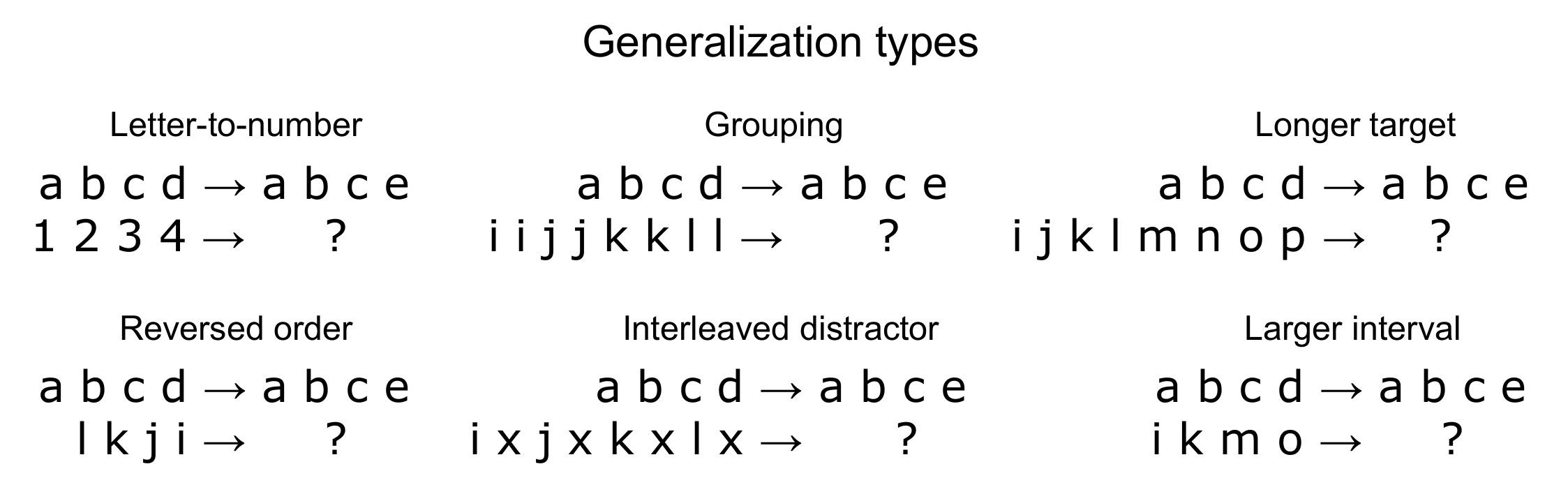}
    \subcaption{}
    \label{letterstring_generalizations}
\end{subfigure}
\caption{\textbf{Letter string analogy problems.} Transformation between source strings must be identified and applied to target string. Mapping between source and target may involve one or more generalizations. \textbf{(a)} Easy problem involving zero generalizations. \textbf{(b)} Difficult problem involving three generalizations (grouping, reversed order, and interleaved distractors). \textbf{(c)} Problem involving generalization from letters to real-world concepts. \textbf{(d)} Transformations were sampled from set of six possible types: sequence extension, successor transformation (applied to the last letter in the string), predecessor transformation (applied to the first letter in the string), removal of a redundant letter, `fixing' an alphabetic sequence (replacing an out-of-place letter), and sorting. \textbf{(e)} Generalizations were sampled from set of six possible types: letter-to-number, grouping, longer target string, reversed order, interleaved distractors, and larger interval.} 
\label{letterstring_problems}
\end{figure}

It is important to highlight the differences between the Digit Matrices and traditional visual matrix reasoning problems. In order to solve visual matrix reasoning problems, pixel-level inputs must be parsed into objects, and visual attributes (shape, size, etc.) must be disentangled. In the Digit Matrices, the text-based inputs are already parsed and disentangled, essentially providing GPT-3 (which is not capable of visual processing) with pseudo-symbolic inputs. Interestingly, despite these significant differences, we found that overall error rates for human participants were very similar for the Digit Matrices vs. the original image-based SPM problem set, and showed a similar pattern across problem types (Figure~\ref{SPM_vs_digit_mat}). These results suggest that, while the Digit Matrices do not engage the visual processes involved in traditional SPM problems (i.e., deriving disentangled representations from pixel-level inputs), they likely engage a similar set of core reasoning processes (i.e., inducing abstract rules from those representations). More generally, performance on verbal, visuospatial, and mathematical analogy problems are known to be highly correlated for people~\cite{snow1984topography}. Accordingly, GPT-3's success on the Digit Matrices can be taken as evidence that it has acquired core capabilities underlying analogy, though it will be important in future work to investigate how these reasoning processes might be integrated with visual processing.

\subsection{Letter string analogies}
\label{letterstring_section}

A central feature of human analogical reasoning is its flexibility. Human reasoners are capable of identifying abstract similarities between situations even when these situations are superficially quite different. Often this involves a process of \textit{re-representation}, in which an initial problem representation is revised so as to facilitate the discovery of an analogy~\cite{chalmers1992high,hofstadter1995fluid,lovett2017modeling}.

\captionsetup{labelfont=bf,font=small}
\begin{figure}[!tb]
\centering
\begin{subfigure}[t]{0.02\textwidth}
    \textbf{\fontfamily{phv}\selectfont{a}}
\end{subfigure}
\begin{subfigure}[t]{.37\linewidth}\vskip 0pt
    \includegraphics[width=\linewidth]{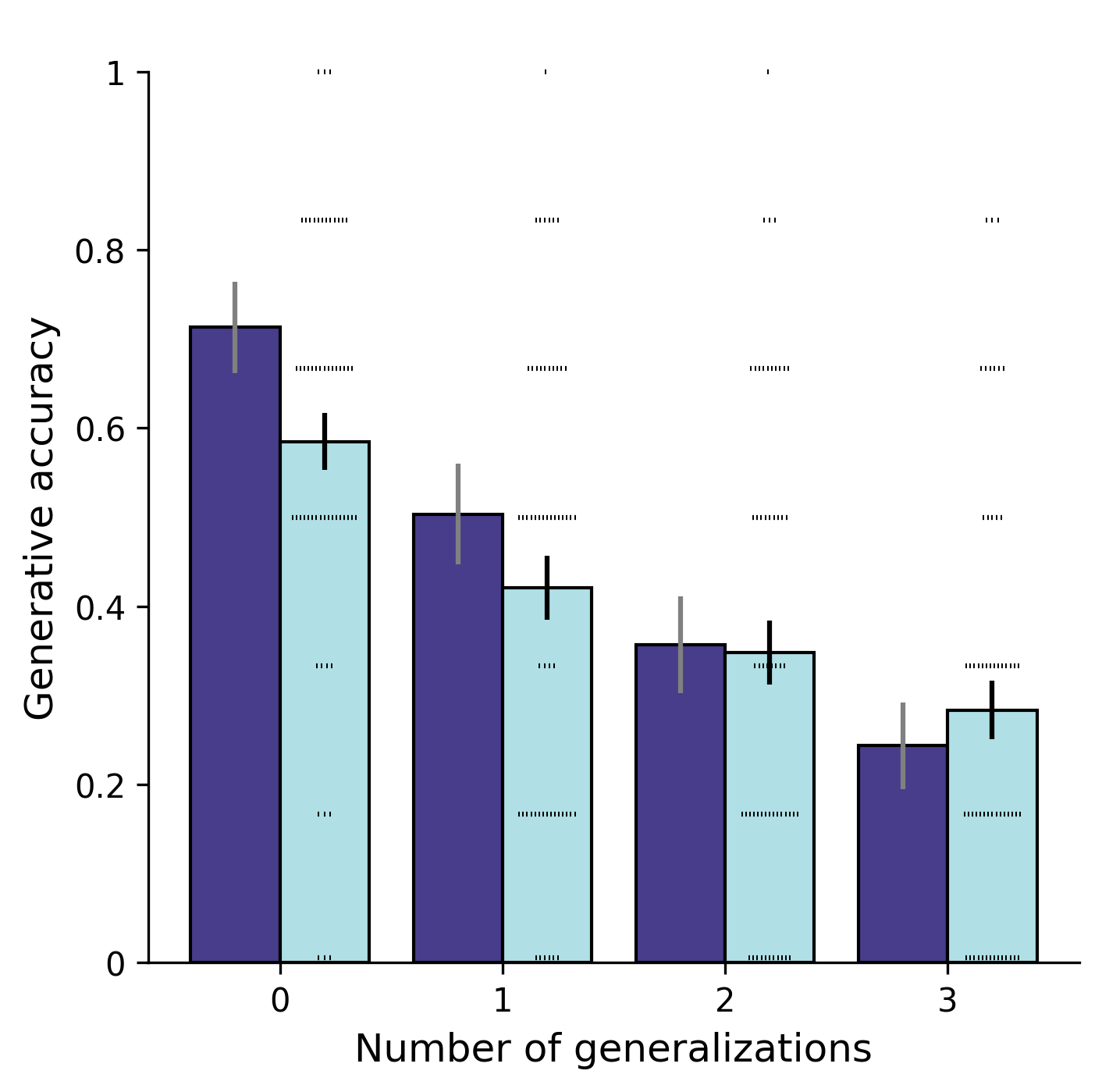} 
    \subcaption{}
    \label{letterstring_all_gen}
\end{subfigure}
\begin{subfigure}[t]{0.02\textwidth}
    \textbf{\fontfamily{phv}\selectfont{b}}
\end{subfigure}
\begin{subfigure}[t]{.535\linewidth}\vskip 0pt
    \includegraphics[width=\linewidth]{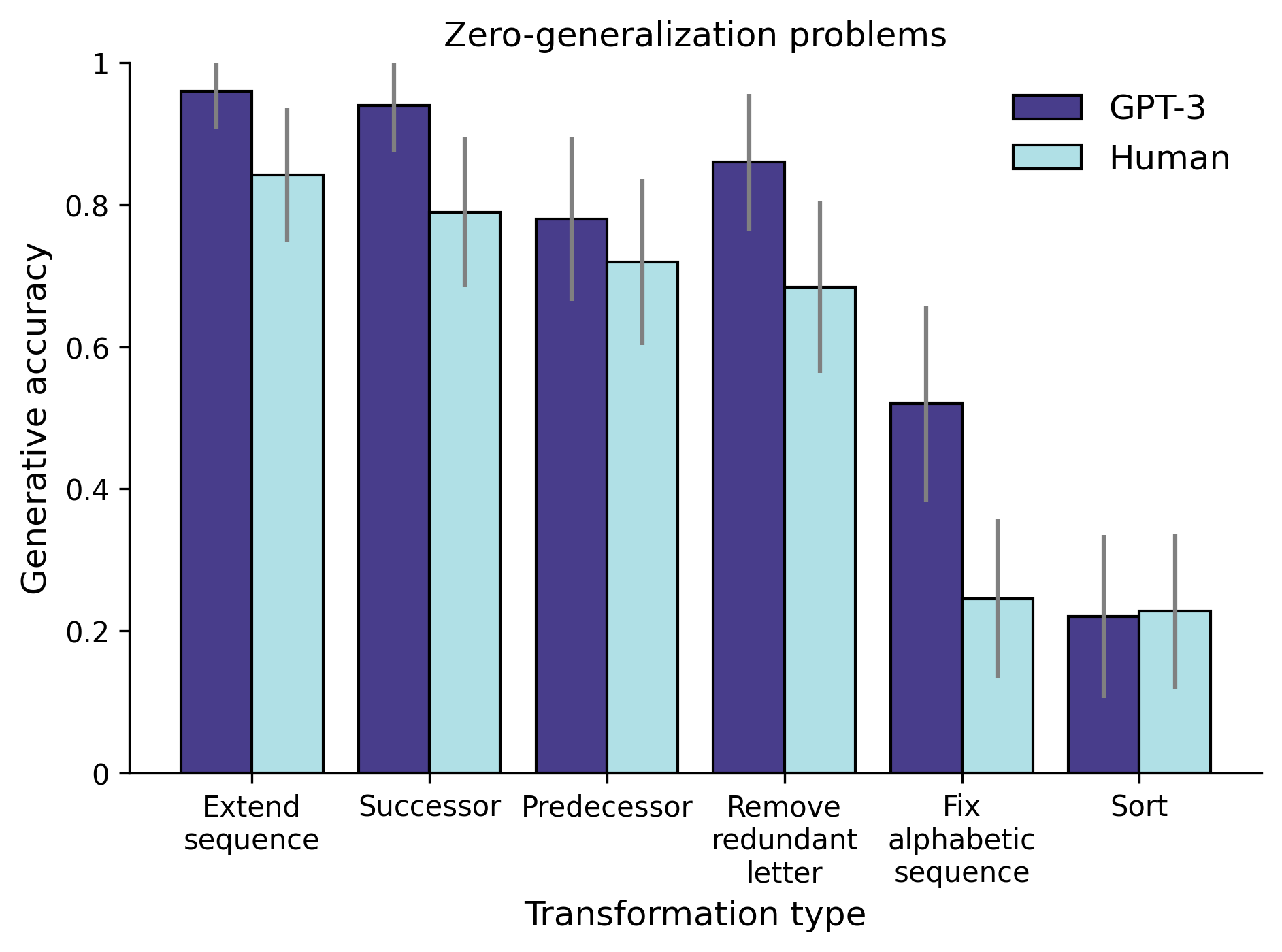} 
    \subcaption{}
    \label{letterstring_zero_gen}
\end{subfigure}
\\
\begin{subfigure}[t]{0.02\textwidth}
    \textbf{\fontfamily{phv}\selectfont{c}}
\end{subfigure}
\begin{subfigure}[t]{.535\linewidth}\vskip 0pt
    \includegraphics[width=\linewidth]{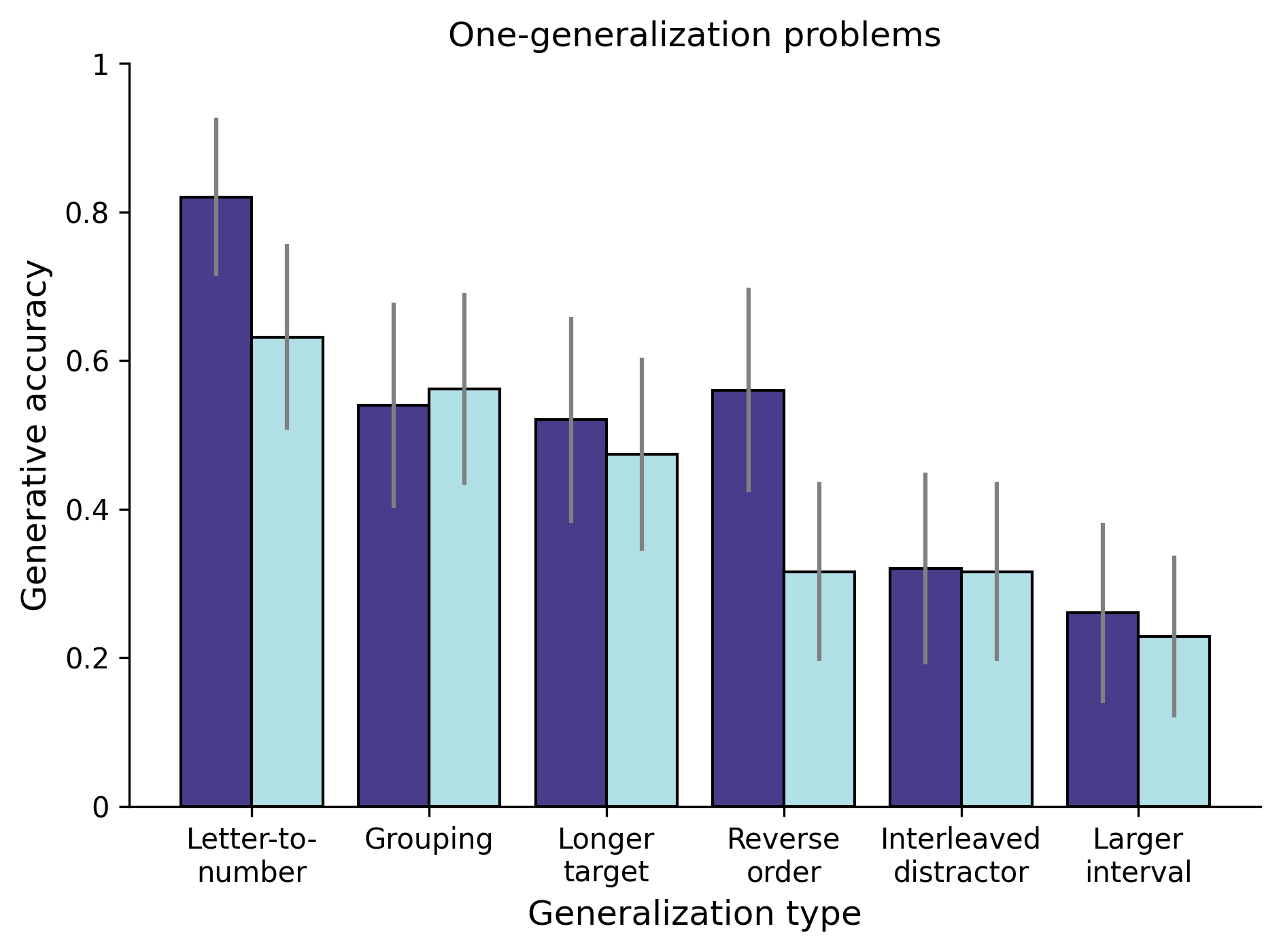} 
    \subcaption{}
    \label{letterstring_one_gen}
\end{subfigure}
\begin{subfigure}[t]{0.02\textwidth}
    \textbf{\fontfamily{phv}\selectfont{d}}
\end{subfigure}
\begin{subfigure}[t]{.39\linewidth}\vskip 0pt
    \includegraphics[width=\linewidth]{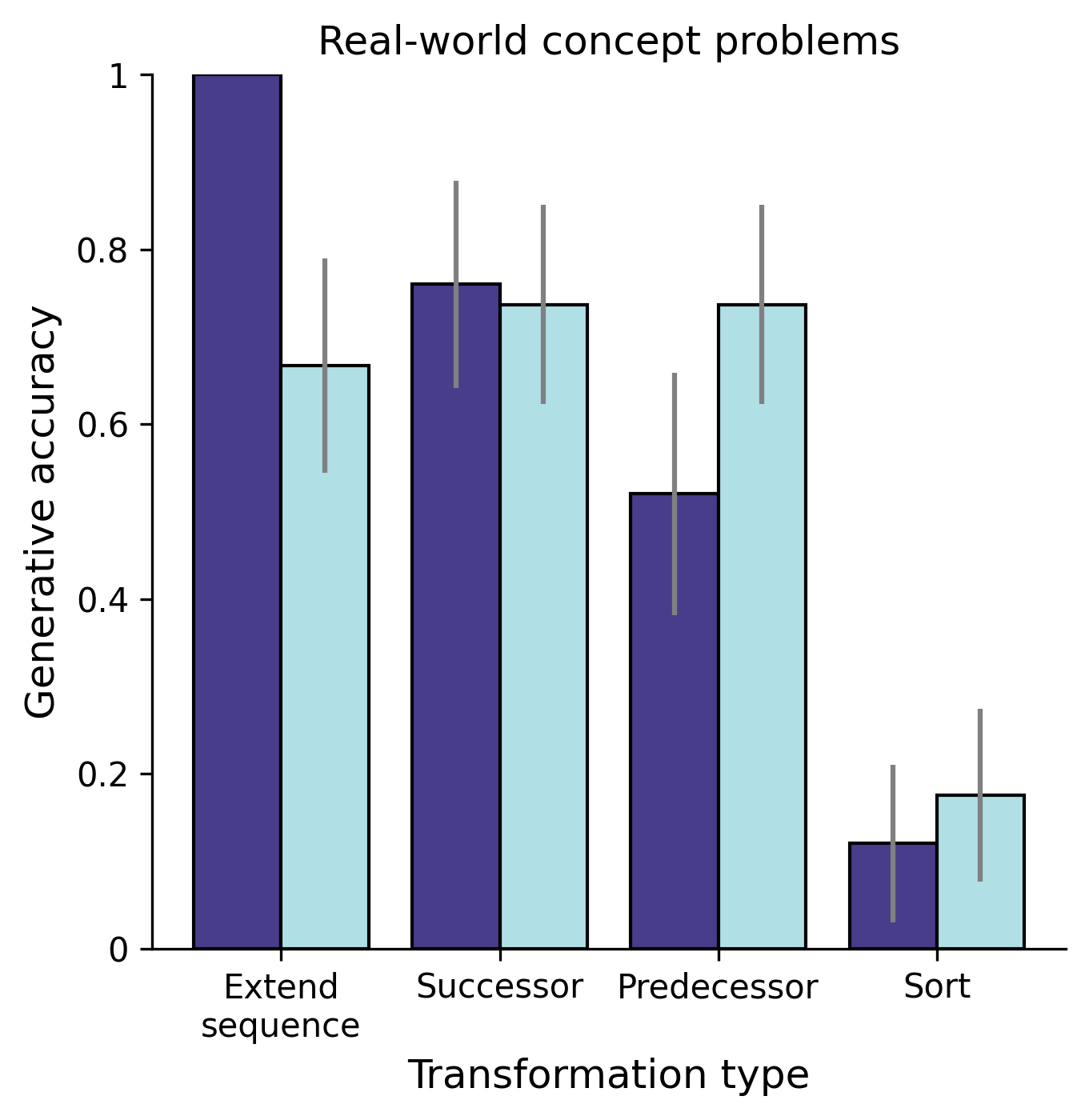} 
    \subcaption{}
    \label{letterstring_realworld_results}
\end{subfigure}
\caption{\textbf{Letter string analogy results.} GPT-3 displayed strong performance on letter string problems, and showed a similar pattern to human participants across conditions. \textbf{(a)} GPT-3 and human performance as a function of the number of generalizations between source and target. \textbf{(b)} Performance on zero-generalization problems as a function of transformation type. \textbf{(c)} Performance on one-generalization problems as a function of generalization type. \textbf{(d)} Performance on problems requiring generalization from letters to real-world concepts. Human results reflect average performance for N=57 participants (UCLA undergraduates). Black error bars represent standard error of the mean across participants. Each dot represents accuracy for a single participant. Note that (b-d) do not show individual participant results because each participant only completed one problem in each condition. Gray error bars represent 95\% binomial confidence intervals for average performance across multiple problems.} 
\label{letterstring_results}
\end{figure}

Hofstadter and Mitchell~\cite{hofstadterMitchell94,mitchell1993analogy} introduced the letter string analogy domain to evaluate computational models of analogical reasoning, with a particular emphasis on the process of re-representation. The basic problem structure is illustrated in Figure~\ref{letterstring_zerogen_prob}. In this example, the source string `a b c d' has been transformed by converting the final letter to its successor, resulting in the string `a b c e'. This transformation must be identified, and then applied to the target string `i j k l', yielding the answer `i j k m'.

Though this example is simple, letter string problems can be made quite complex by introducing various generalizations between the source and target strings. For instance, the target may involve groups of letters rather than individual letters (e.g., `i i j j k k l l'), or may involve a sequence with a reversed order relative to the source (e.g., `l k j i'). In these cases, the transformation identified in the source (e.g., a successor transformation applied to the final letter in the sequence) must be generalized to an analogous transformation (e.g., a successor transformation applied to the final \textit{group of letters}, or a \textit{predecessor} transformation applied to the first letter). This feature makes letter string analogy problems well-suited to test the capacity for re-representation.

To evaluate GPT-3, we created a novel letter string problem set (Figure~\ref{letterstring_problems}), and carried out a systematic comparison with human participants (N=57, UCLA undergraduates). The problem set involved a range of different transformation (Figure~\ref{letterstring_transformations}) and generalization types (Figure~\ref{letterstring_generalizations}). Each transformation type could be combined with any generalization type, and multiple generalization types could be combined together to yield more challenging problems (Figure~\ref{letterstring_threegen_prob}). Problems were presented to GPT-3 along with a prompt (`Let's try to complete the pattern:'), using a format similar to the Digit Matrices.

Figure~\ref{letterstring_results} shows the results of this evaluation. GPT-3 showed stronger overall performance than human participants (Figure~\ref{letterstring_all_gen}; logistic regression, main effect of GPT-3 vs. human participants: OR $=1.76$, $p = 6.3\times10^{-5}$, CI $=[1.34,2.31]$), an effect that was driven primarily by stronger performance on zero-generalization problems (main effect of GPT-3 vs. human participants for zero-generalization problems: OR $=1.76$, $p = 0.0007$, CI $=[1.27,2.46]$). Performance was strongly affected by the number of generalizations in both GPT-3 and human participants (main effect of number of generalizations, GPT-3: OR $=0.51$, $p = 2\times10^{-16}$, CI $=[0.45,0.57]$; human participants: OR $=0.66$, $p = 5.9\times10^{-16}$, CI $=[0.6,0.73]$). GPT-3 and human participants also showed similar error patterns across transformation types (Figure~\ref{letterstring_zero_gen}) and generalization types (Figure~\ref{letterstring_one_gen}), as quantified by a correlation analysis for accuracy across different problem subtypes ($r(39)=0.7$, $p = 3.6\times10^{-7}$).

We also investigated a novel variant on letter string problems involving generalization from letters to real-world concepts (Figure~\ref{letterstring_realworld_prob}). GPT-3 showed strong performance on these problems, though with some discrepancies for different transformation types (Figure~\ref{letterstring_realworld_results}). These results suggest that GPT-3 has developed an abstract notion of successorship that can be flexibly generalized between different domains (e.g., alphabetic successorship vs. temperature successorship).

One important caveat is that GPT-3's performance on this task was somewhat sensitive to the way in which problems were formatted. For instance, performance suffered when no prompt was provided (Supplementary Figure~\ref{letterstring_all_gen_noprompt}), or when problems were presented in the form of a complete sentence (Supplementary Figure~\ref{letterstring_all_gen_sentence}). However, even in these cases, GPT-3's zero-shot performance was both within the range of human participants (within one standard deviation), and closely matched the pattern of human performance across problem types (correlation analysis, no prompt: $r(39)=0.6$, $p = 5.3\times10^{-5}$, sentence format: $r(39)=0.76$, $p = 4.2\times10^{-6}$).

\captionsetup{labelfont=bf,font=small}
\begin{figure}[h!]
\centering
\begin{subfigure}[t]{0.02\textwidth}
    \textbf{\fontfamily{phv}\selectfont{a}}
\end{subfigure}
\begin{subfigure}[t]{.328\linewidth}\vskip 0pt
    \includegraphics[width=\linewidth]{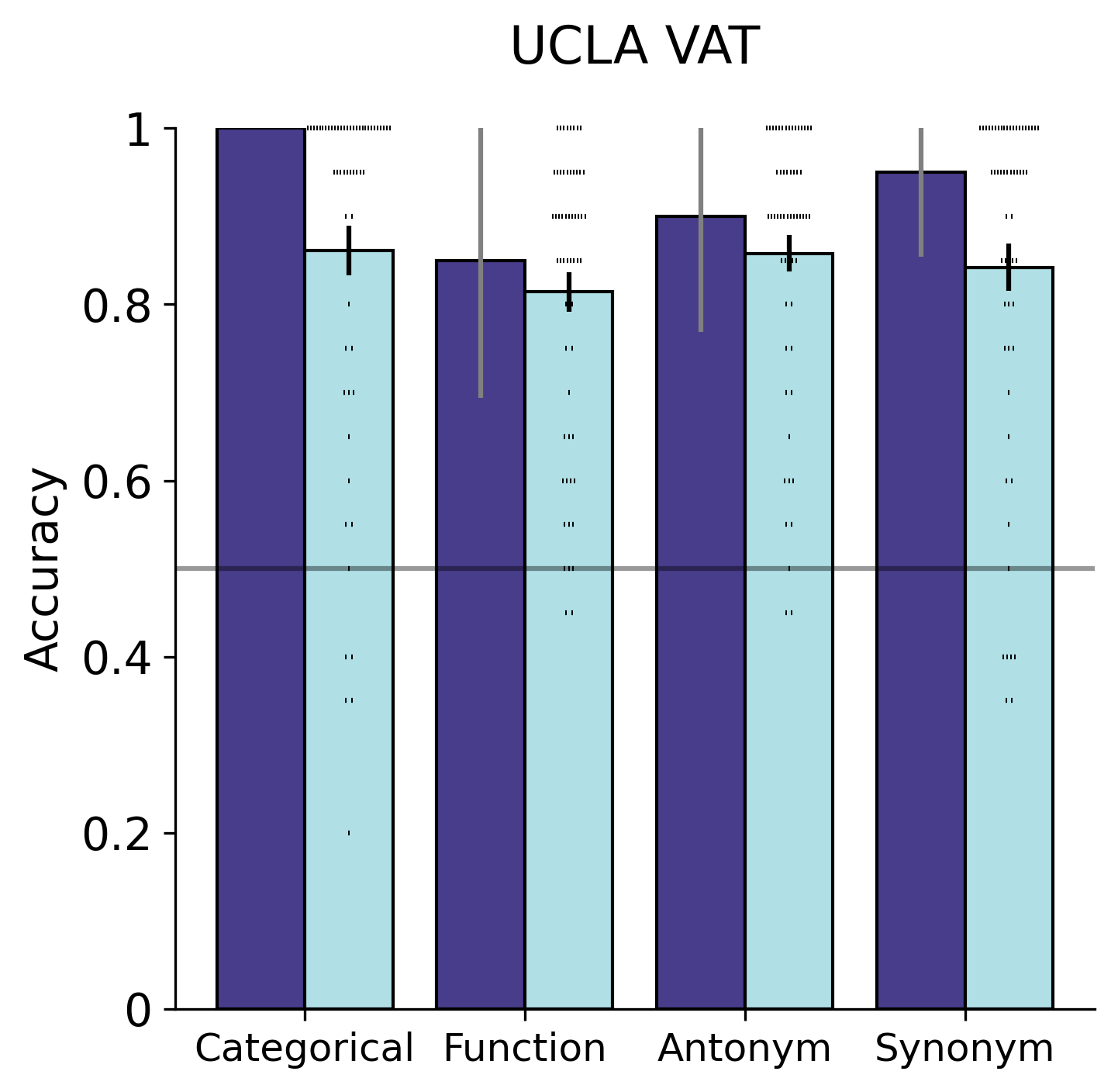} 
    \subcaption{}
    \label{UCLA VAT_results}
\end{subfigure}
\begin{subfigure}[t]{0.02\textwidth}
    \textbf{\fontfamily{phv}\selectfont{b}}
\end{subfigure}
\begin{subfigure}[t]{.43\linewidth}\vskip 0pt
    \includegraphics[width=\linewidth]{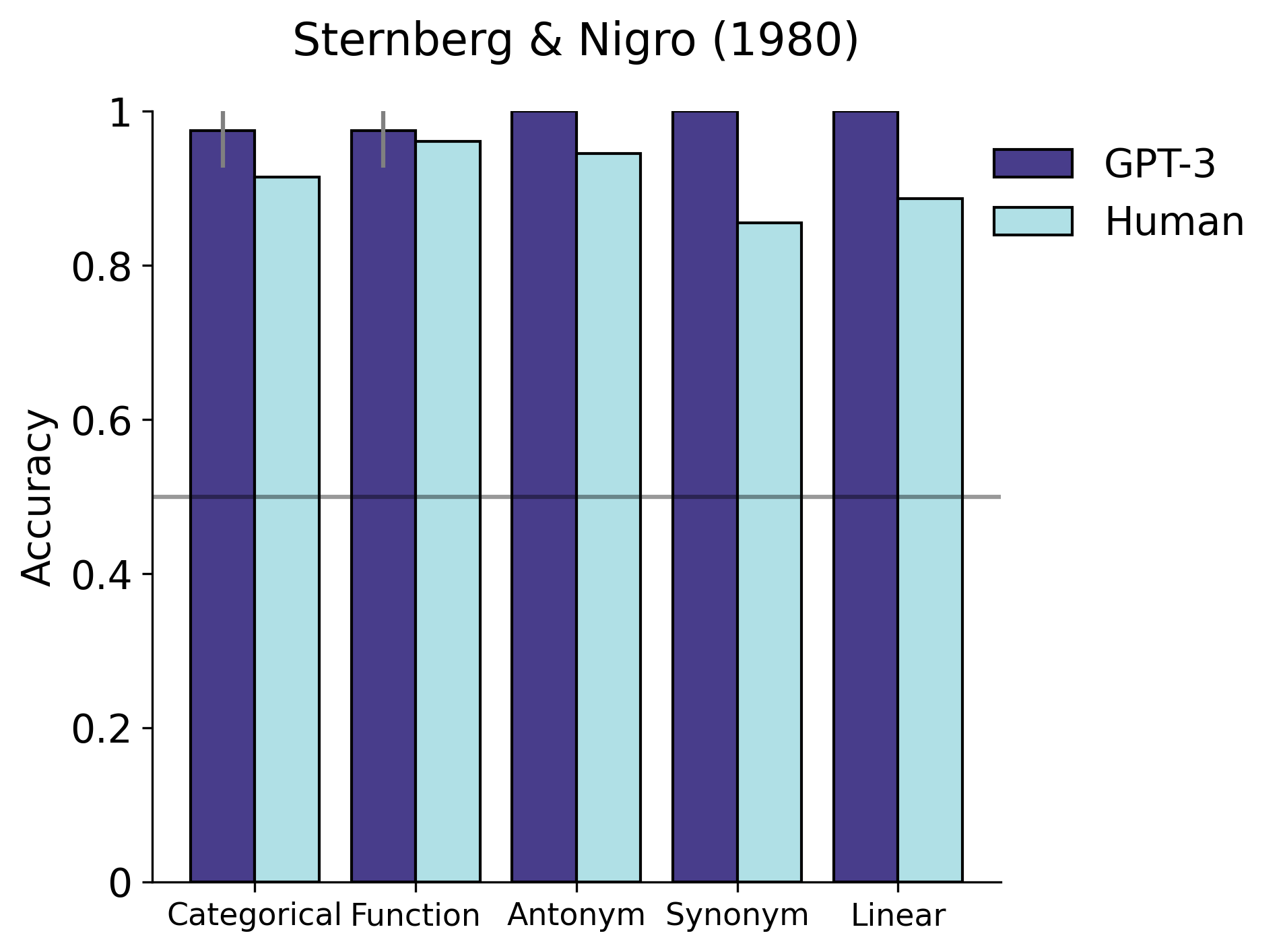} 
    \subcaption{}
    \label{sternberg_results}
\end{subfigure}
\\
\begin{subfigure}[t]{0.02\textwidth}
    \textbf{\fontfamily{phv}\selectfont{c}}
\end{subfigure}
\begin{subfigure}[t]{.129\linewidth}\vskip 0pt
    \includegraphics[width=\linewidth]{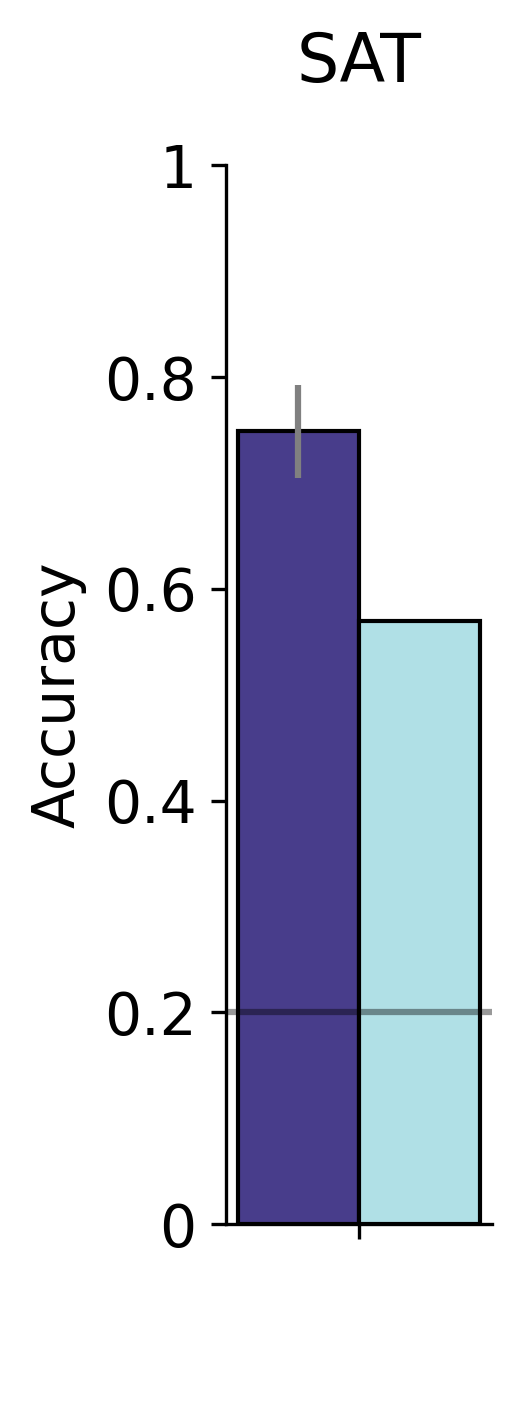} 
    \subcaption{}
    \label{SAT_results}
\end{subfigure}
\begin{subfigure}[t]{0.02\textwidth}
    \textbf{\fontfamily{phv}\selectfont{d}}
\end{subfigure}
\begin{subfigure}[t]{.51\linewidth}\vskip 0pt
    \includegraphics[width=\linewidth]{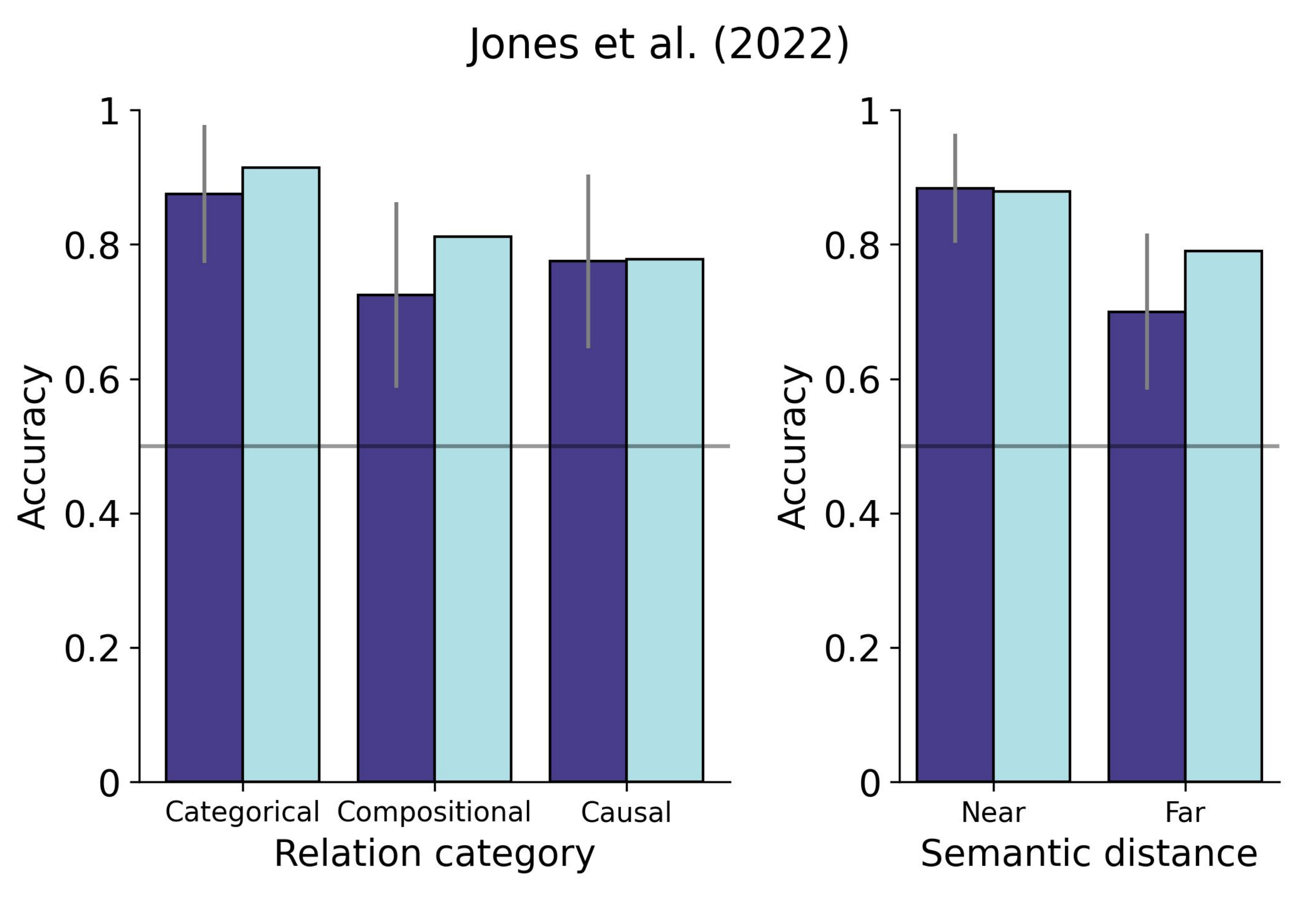} 
    \subcaption{}
    \label{jones_results}
\end{subfigure}
\caption{\textbf{Verbal analogy results.} \textbf{(a)} Results for UCLA Verbal Analogy Test (VAT)~\cite{lu2019emergence}. Human results reflect average performance for N=57 participants. Black error bars represent standard error of the mean. Each dot represents accuracy for a single participant. \textbf{(b)} Results for dataset from Sternberg and Nigro~\cite{sternberg1980developmental}. Human results reflect average performance for N=20 participants. \textbf{(c)} Results for SAT analogy problems from Turney et al.~\cite{turney2003combining}. These problems involve five answer choices, and thus chance performance is $20\%$. Human results reflect an estimate of the average performance for high school students taking the SAT (see~\cite{turney2005corpus} for details). \textbf{(d)} Results for dataset from Jones et al.~\cite{jones2022differential}. Human results reflect average performance for N=241 participants. Gray error bars represent 95\% binomial confidence intervals for average performance across multiple problems. Gray horizontal lines represent chance performance.} 
\label{verbal_analogy_figure}
\end{figure}

\subsection{Four-term verbal analogies}
\label{verbal_analogy_section}

Though matrix reasoning and letter string analogies involve a high degree of relational complexity, one limitation is that they consist of highly constrained, synthetic relations, such as alphabetic or numerical successorship. GPT-3's ability to solve problems involving more real-world concepts (e.g., `a b c $\rightarrow$ a b d, cold cool warm $\rightarrow$ ?') suggests that its analogical capabilities may not be limited to such artificial settings. To further evaluate GPT-3's capacity to reason about real-world relational concepts, we tested it on four-term verbal analogy problems involving a broader range of semantic relations.

We evaluated GPT-3 on four separate datasets~\cite{sternberg1980developmental,turney2003combining,lu2019emergence,jones2022differential}. To the best of our knowledge, these constitute an exhaustive set of four-term verbal analogy problems for which human behavioral data is available~\cite{ichien2020verbal}. Each dataset contains a series of four-term analogy problems in the form `A:B::C:?', together with a set of answer choices (i.e., potential choices of D). For each problem, GPT-3 was evaluated by presenting the problem together with each potential answer choice, and selecting the option for which GPT-3 assigned a higher log probability. The problem and GPT-3's choice were then appended to the context window for the next problem, thereby simulating any contextual effects that might arise when solving multiple problems in a row, as human participants typically do.

Figure~\ref{verbal_analogy_figure} shows the results for all datasets. GPT-3 performed as well or better than human participants (minimum education level of high-school graduation, located in the United States and recruited using Amazon Mechanical Turk) on the UCLA Verbal Analogy Test (VAT)~\cite{lu2019emergence}, involving categorical, functional, antonym, and synonym relations (Figure~\ref{UCLA VAT_results}), and on a dataset from Sternberg and Nigro~\cite{sternberg1980developmental} involving these same four relation types and linear order relations (Figure~\ref{sternberg_results}). On a dataset of SAT analogy problems from Turney et al.~\cite{turney2003combining}, GPT-3 surpassed the estimated average level of performance for high school students taking the SAT (Figure~\ref{SAT_results}). GPT-3 also showed performance in the same range as human participants (though numerically weaker) on a problem set from Jones et al.~\cite{jones2022differential} involving categorical, compositional, and causal relations (Figure~\ref{jones_results}).

In addition to displaying generally strong performance on these problem sets, GPT-3 also displayed sensitivity to semantic content similar to that observed in human participants. In the dataset from Jones et al.~\cite{jones2022differential} (Figure~\ref{jones_results}), participants performed worse on problems in which the analogs were semantically distant (i.e., the A and B terms had low semantic similarity to C and D), an effect that was also displayed by GPT-3 (logistic regression, effect of semantic distance for GPT-3: OR $=3.24$, $p = 0.0165$, CI $=[1.24,8.5]$). These results align with a more general phenomenon in which human reasoning is facilitated by semantically meaningful or coherent content~\cite{wason1968reasoning,dasgupta2022language}.

\captionsetup{labelfont=bf,font=small}
\begin{figure}[h!]
\centering
\begin{subfigure}[t]{.4\linewidth}\vskip 0pt
    \includegraphics[width=\linewidth]{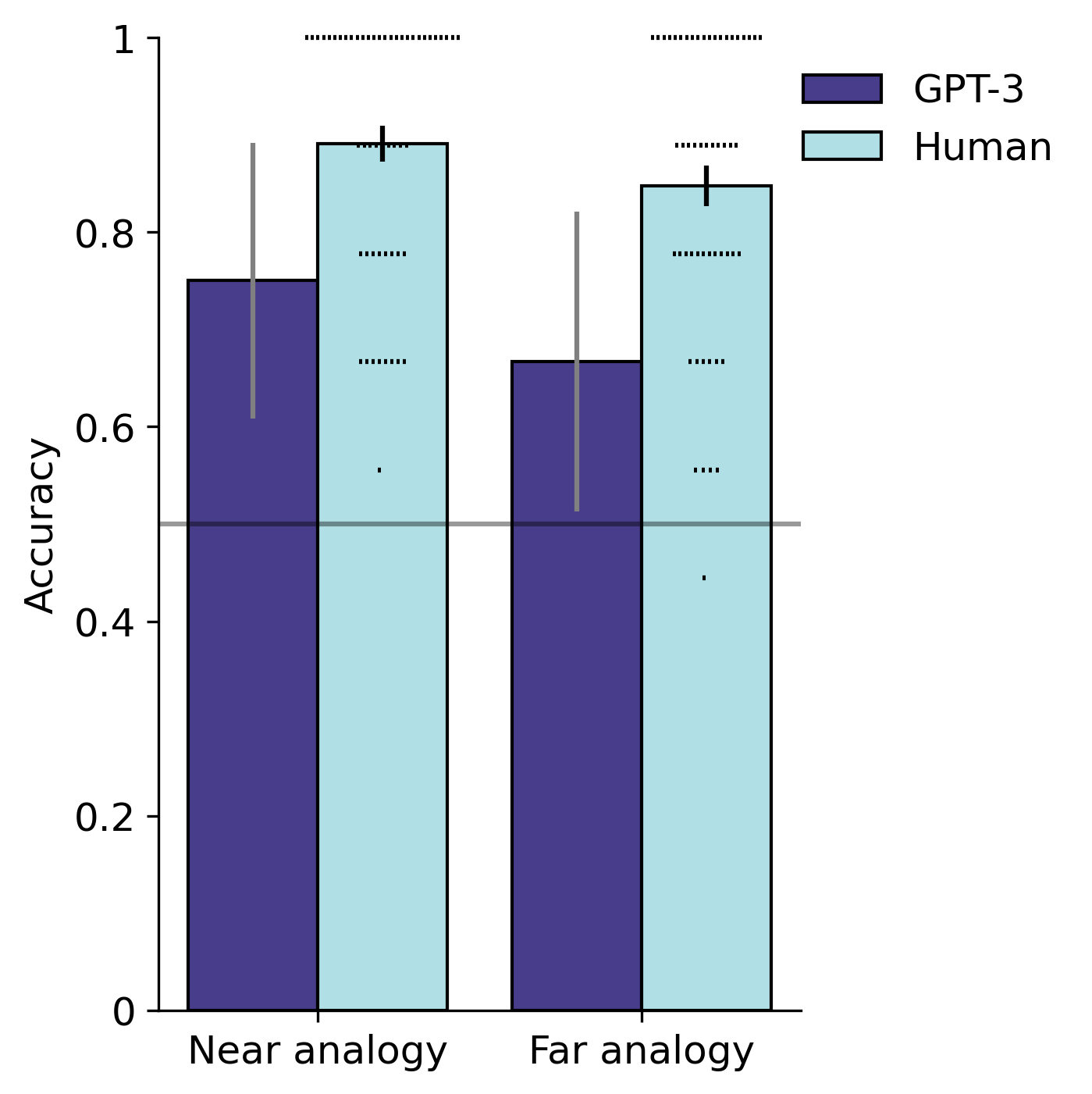} 
    \subcaption{}
\end{subfigure}
\caption{\textbf{Story analogy results.} Results for identification of analogies between stories, using materials from Gentner et al.~\cite{gentner1993roles}. When presented with a source story and two target stories, both GPT-3 and human participants showed a preference for target stories that shared higher-order relations with the source vs. those that only shared first-order relations. Near analogy condition involves within-domain comparison between stories with similar entities. Far analogy condition involves cross-domain comparison between stories with different entities. Human results reflect average performance for N=54 participants (UCLA undergraduates). Black error bars represent standard error of the mean across participants. Each dot represents accuracy for a single participant. Gray error bars represent 95\% binomial confidence intervals for average performance across multiple problems.  Gray horizontal line represents chance performance.} 
\label{story_analogy_results}
\end{figure}

\subsection{Story analogies}
\label{story_analogy_section}

Human reasoners are able not only to form analogies between individual concepts, but can also identify correspondences between complex real-world events, involving many entities and relations. When making such comparisons, human reasoning is especially sensitive to \textit{higher-order} relations -- relations between relations -- notably causal relations between events. Such higher-order relations play a central role in some cognitive theories of analogy~\cite{gentner1983structure}, and it is thus important to establish whether GPT-3 displays a similar sensitivity to them.

To address this question, we tested GPT-3 on a set of story analogies from Gentner et al.~\cite{gentner1993roles}. In each set, a source story is compared to two potential target stories, each of which is matched with the source story in terms of first-order relations, but only one of which shares the same causal relations as the source (see Methods Section~\ref{example_story_materials} for examples). Gentner et al. found that human participants rated the target stories as more similar when they shared the same causal relations as the source story. These problems are further defined by two different comparison conditions. In the \textit{near analogy} condition (referred to as `literal similarity' vs. `mere appearance' by Gentner et al.), the target stories also share the same basic entities as the source story, making for a less abstract, within-domain comparison. In the \textit{far analogy} condition (referred to as `true analogy' vs. `false analogy' by Gentner et al.), the target stories involve different entities from the source story, but share first-order relations, resulting in a more challenging, cross-domain comparison.

To facilitate a direct comparison with GPT-3, we performed a new behavioral study with these materials. For each source story, participants indicated which of two target stories was more analogous. Both GPT-3 and human participants (N=54, UCLA undergraduates) showed a sensitivity to higher-order relations (Figure~\ref{story_analogy_results}), most often selecting the target story that shared causal relations with the source (combined near and far analogy; GPT-3, binomial test: $p = 0.0005$; human participants, one-sample t-test: $t(53)=21.3$, $p = 1.1\times10^{-27}$; null hypotheses for both tests is chance-level performance of 0.5). This effect was significant for both GPT-3 and human participants in the near analogy condition (GPT-3, binomial test: $p = 0.0039$; human participants, one-sample t-test: $t(53)=21.5$, $p = 8.5\times10^{-28}$), but only human participants showed a significant effect in the far analogy condition (GPT-3, binomial test: $p = 0.065$; human participants, one-sample t-test: $t(53)=16.7$, $p = 9.3\times10^{-23}$).

Unlike the other task domains considered in the present work, this was a case in which college students clearly outperformed GPT-3 (logistic regression, main effect of GPT-3 vs. human participants: OR $=0.37$, $p = 0.0003$, CI $=[0.21,0.63]$). Indeed, a significant number of participants (15/54) selected the analogous story on every trial. However, in an initial investigation of GPT-4~\cite{gpt4_tech_report}, we found that it displays stronger performance on this task, more robustly picking the analogous story even in the far analogy condition, and displaying nearly perfect performance in the near analogy condition (Supplementary Figure~\ref{story_analogy_results_gpt4}, Section~\ref{gpt4_story_example}). It therefore seems likely that further scaling of large language models will enhance their sensitivity to causal relations. 

\subsection{Analogical problem-solving}
\label{analogical_problem_solving_section}

In everyday thinking and reasoning, analogical comparisons are often made for the purpose of achieving some goal, or solving a novel problem. Thus far, our tests of GPT-3 have assessed its capacity for identifying analogies in text-based inputs with varying formats, but can GPT-3 also use these analogies to derive solutions to novel problems, as human reasoners do?

As a preliminary investigation of this issue, we performed a qualitative evaluation using a paradigm developed by Gick and Holyoak~\cite{gick1980analogical}. In that paradigm, participants are presented with a target problem in the form of a story. In the original study, Duncker's radiation problem was used~\cite{duncker1945problem}. In that problem, a doctor wants to use radiation to destroy a malignant tumor, but destroying the tumor with a single high-intensity ray will also damage the surrounding healthy tissue. The solution -- to use several low-intensity rays that converge at the site of the tumor -- is rarely identified spontaneously, but participants are more likely to discover this solution when they are first presented with an analogous source story. In the original study, the source story involved a general who wants to capture a fortress ruled by an evil dictator, but cannot do so by sending his entire army along a single road, which would trigger landmines. The general instead breaks his army up into small groups that approach the fortress from multiple directions, thus avoiding triggering the mines. 

We first presented GPT-3 with the target problem in isolation. GPT-3 proposed a solution that involved injecting a radiation source directly into the tumor, rather than identifying the intended solution based on the convergence of multiple low-intensity radiation sources (Supplementary Section~\ref{radiation_prob_only}). However, when first presented with the general story, followed by the target problem, GPT-3 correctly identified the convergence solution (Supplementary Section~\ref{radiation_prob_general_story}). GPT-3 was further able to correctly explain the analogy, and to identify the specific correspondences between the source story and target problem when prompted (e.g., general $\leftrightarrow$ doctor, dictator $\leftrightarrow$ tumor, army $\leftrightarrow$ rays). We also found similar results when using distinct source analogs taken from another study~\cite{holyoak1987surface} (Supplementary Section~\ref{radiation_prob_other_source}).

In a more challenging version of this paradigm, participants were first presented with both the general story, and two other non-analogous stories intended to serve as distractors. In this context, human participants were much less likely to identify the convergence solution. However, when given a prompt to explicitly consider the previously presented stories when trying to solve the radiation problem, participants were often able to correctly identify the analogous general story, and use this analogy to devise the convergence solution. Remarkably, we found that GPT-3 displayed these same effects. When presented with these same distracting, non-analogous stories, GPT-3 no longer identified the convergence solution, instead proposing the same solution that it proposed in response to the radiation problem alone (Supplementary Section~\ref{radiation_prob_distracting_stories}). But when prompted to consider the previous stories, GPT-3 both correctly identified the general story as most relevant, and proposed the convergence solution (Supplementary Section~\ref{radiation_prob_distracting_stories_prompt}).

We also evaluated GPT-3 using materials from a developmental study that employed a similar paradigm~\cite{holyoak1984development}. In that study, children were tasked with transferring gumballs from one bowl to another bowl that was out of reach, and provided with a number of materials for doing so (e.g., a posterboard, an aluminum walking cane, a cardboard tube), permitting multiple possible solutions. The key result was that when children were first presented with an analogous source story (about a magical genie trying to transfer jewels between two bottles), they were more likely to identify a solution to the target problem that was analogous to the events described in the source story. 

When presented with this target problem, GPT-3 mostly proposed elaborate, but mechanically nonsensical solutions, with many extraneous steps, and no clear mechanism by which the gumballs would be transferred between the two bowls (Supplementary Sections~\ref{gumball_prob_only}-\ref{gumball_prob_magic_carpet_story}). However, when asked to explicitly identify an analogy between the source story and target problem, GPT-3 \textit{was} able to identify all of the major correspondences, even though it could not use this analogy to discover an appropriate solution. This finding suggests that GPT-3's difficulty with this problem likely stems from its lack of physical reasoning skills, rather than being due to a difficulty with analogical mapping per se. It is also worth noting that in the original study, this task was presented to children with real physical objects, which likely aided the physical reasoning process relative to the purely text-based input provided to GPT-3. Overall, these results provide some evidence that GPT-3 is capable of using analogies for the purposes of problem-solving, but its ability to do so is constrained by the content about which it can reason, with particular difficulty in the domain of physical reasoning.

\section{Discussion}

We have presented an extensive evaluation of analogical reasoning in a state-of-the-art large language model. We found that GPT-3 appears to display an emergent ability to reason by analogy, matching or surpassing human performance across a wide range of text-based problem types. These included a novel problem set (Digit Matrices) modeled closely on Raven's Progressive Matrices, where GPT-3 both outperformed human participants, and captured a number of specific signatures of human behavior across problem types. Because we developed the Digit Matrix task specifically for this evaluation, we can be sure GPT-3 had never been exposed to problems of this type, and therefore was performing zero-shot reasoning. GPT-3 also displayed an ability to solve analogies based on more meaningful relations, including four-term verbal analogies and analogies between stories describing complex real-world events.

It is certainly not the case that GPT-3 mimics human analogical reasoning in all respects. Our tests were limited to processes that can be carried out within a local temporal context, but humans are also capable of retrieving potential source analogs from long-term memory, and ultimately of developing new concepts based on the comparison of multiple analogs. Unlike humans, GPT-3 does not have long-term memory for specific episodes. It is therefore unable to search for previously-encountered situations that might create useful analogies with a current problem. For example, GPT-3 can use the general story to guide its solution to the radiation problem, but as soon as its context buffer is emptied, it reverts to giving its non-analogical solution to the problem -- the system has learned nothing from processing the analogy. GPT-3's reasoning ability is also limited by its lack of physical understanding of the world, as evidenced by its failure (in comparison with human children) to use an analogy to solve a transfer problem involving construction and use of simple tools. GPT-3's difficulty with this task is likely due at least in part to its purely text-based input, lacking the multimodal experience necessary to build a more integrated world model~\cite{mcclelland2020placing}. Finally, we found GPT-3 was limited in its ability to evaluate analogies based on causal relations, particularly in cross-domain comparisons between stories (far analogy). 

But despite these major caveats, our evaluation reveals that GPT-3 exhibits a very general capacity to identify and generalize -- in zero-shot fashion -- relational patterns to be found within both formal problems and meaningful texts. These results are extremely surprising. It is commonly held that although neural networks can achieve a high level of performance within a narrowly-defined task domain, they cannot robustly generalize what they learn to new problems in the way that human learners do~\cite{marcus2001algebraic,lake2017building,webb2020learning,mitchell2021abstraction}. Analogical reasoning is typically viewed as a quintessential example of this human capacity for abstraction and generalization, allowing human reasoners to intelligently approach novel problems zero-shot. Our results indicate that GPT-3 -- unlike any other neural network previously tested on analogy problems -- displays a capacity for such zero-shot analogical reasoning across a broad range of tasks.

The deep question that now arises is how GPT-3 achieves the analogical capacity that is often considered the core of human intelligence. One possibility is that, perhaps as a result of the sheer size and diversity of GPT-3's training data, it has been forced to develop mechanisms similar to those thought to underlie human analogical reasoning -- despite not being explicitly trained to do so. The consensus among cognitive scientists working on analogy is that this human ability depends on systematic comparison of knowledge based on explicit relational representations. It is unclear whether and how GPT-3 would implement these processes. Does GPT-3 possess some form of emergent relational representations, and if so, how are they computed? Does it perform a mapping process similar to the type that plays a central role in cognitive theories of analogy~\cite{gentner1983structure}? 

A few properties of the transformer architecture~\cite{vaswani2017attention}, on which GPT-3 and other large language models are based, are worth considering here. The first is the central role played by \textit{similarity}. Transformers are built on a self-attention operation, which involves explicitly computing the similarity between each pair of vectors in the inputs to each layer. This pairwise evaluation of similarity is also a key feature of cognitive models of analogy, where it provides the primary constraint guiding the process of analogical mapping. In traditional symbolic models~\cite{falkenhainer1989structure}, this takes the form of literal identicality between symbols, but in more recent models~\cite{lu2022probabilistic,webb2022zero}, a graded similarity function that operates over vector-based inputs is used, much like the self-attention operation in transformers. Second, transformer self-attention employs a form of \textit{indirection}, in which one set of embeddings is used to reference another set of embeddings (i.e., keys vs. values) -- arguably a form of variable binding. Cognitive scientists have long hypothesized that variable binding plays a central role in analogical reasoning, and abstract reasoning more broadly, as it potentially allows generalization of abstract roles across different contexts~\cite{smolensky1990tensor,holyoak2000proper,marcus2001algebraic,kriete2013indirection,webb2020emergent,greff2020binding}. It may be that these features of the transformer make it better equipped to perform zero-shot reasoning than other neural architectures. This possibility aligns with recent evidence that the transformer architecture is an important factor contributing toward the emergence of few-shot learning~\cite{chan2022data}.

But although the mechanisms incorporated into large language models such as GPT-3 may have some important links to building blocks of human reasoning, we must also entertain the possibility that this type of machine intelligence is fundamentally different from the human variety. Humans have evolved to reason within bounds imposed by limited computational power and biological constraints ~\cite{griffiths2020understanding}. Thus, we tend to approach complex problems by breaking them into a set of simpler problems that can be solved separately~\cite{newell1958elements}, an approach that plays a particularly important role in solving challenging analogy problems such as Raven's Matrices~\cite{carpenter1990one}. It is possible that GPT-3, through sheer computational scale, is able to solve such complex problems in a holistic and massively parallel manner, without the need to segment them into more manageable components.

It must also be noted that, regardless of the extent to which GPT-3 employs human-like mechanisms to perform analogical reasoning, we can be certain that it did not \textit{acquire} these mechanisms in a human-like manner. LLMs receive orders of magnitude more training data than do individual human beings (at least if we consider linguistic inputs alone)~\cite{griffiths2020understanding}, and so they cannot be considered as models of the acquisition of analogical reasoning over the course of human development. Nor can they be considered good models of the evolution of analogical reasoning, as their analogical abilities are derived entirely from being trained to predict human-generated text. Human natural language is replete with analogies; accurately predicting natural language therefore likely requires an ability to appreciate analogies. But there is no reason to suppose that the same system, absent human-generated inputs, would spontaneously develop a disposition to think analogically, as apparently happened at some point in human evolution~\cite{penn2008darwin}. Thus, to the extent that large language models capture the analogical abilities of adult human reasoners, their capacity to do so is fundamentally parasitic on natural human intelligence. Nevertheless, the present results indicate that this approach may be sufficient to approximate human-like reasoning abilities, albeit through a radically different route than that taken by biological intelligence.

\section{Methods}

The present research complied with all relevant ethical regulations, and human behavioral experiments were approved by the UCLA Institutional Review Board (IRB protocol \#22-000841, approved May 17, 2022).

\subsection{Code}

Most code was written in Python v3.9.6, using the following packages: NumPy v1.24.3~\cite{harris2020array}, SciPy v1.10.1~\cite{2020SciPy-NMeth}, statsmodels v0.13.5~\cite{seabold2010statsmodels}, Matplotlib v3.7.1~\cite{Hunter:2007}, and pandas v2.0.1~\cite{reback2020pandas}. Logistic regression analyses were carried out in R v4.2.2~\cite{rcite}. Experimental stimuli for human behavioral experiments were written in JavaScript using jsPsych v7.2.1~\cite{de2015jspsych}.

\subsection{GPT-3}

We queried GPT-3 in an automated fashion through the OpenAI API. All simulations reported in the main text employed the text-davinci-003 model variant. Additional simulations, reported in the Supplementary Results, also employed the davinci, code-davinci-002, and text-davinci-002 variants.  The temperature was set to 0 in all simulations. We set max\_tokens (the parameter controlling the maximum number of generated tokens for a given prompt) to 10 for Digit Matrices, 40 for letter string analogies, 10 for four-term verbal analogies, and 256 for story analogies and analogical problem-solving. All other parameters were set to their default values.

For each prompt, GPT-3 generates a proposed completion (a string of tokens), and assigns log probabilities to each token in the prompt and the completion. We used these log probabilities to evaluate GPT-3 on multiple-choice problems. For each choice in a given problem, we concatenated the problem with the choice, and treated the average log probability assigned to the choice tokens as a score, selecting the answer choice with the highest score. This approach was used for Digit Matrices and four-term verbal analogies.

\subsection{Digit Matrices}

\subsubsection{Dataset}
\label{digit_mat_methods}

The digit matrix problems consisted of two major problem categories: transformation and logic problems. Transformation problems contained anywhere from one to five rules, whereas logic problems each contained only a single rule. Transformation problems were defined using a combination of three rule types: constant, distribution-of-3, and progression. The constant rule was defined by the same digit appearing across either rows or columns. The following example shows an instance of a column-wise constant rule (correct answer: `9'):

\begin{center}
[5] [1] [9]
\linebreak[4][5] [1] [9]
\linebreak[4][5] [1] [?]
\end{center}

\noindent The distribution-of-3 rule was defined by the same set of three digits appearing in each row or column, but with the order permuted. In the following example, the digits 6, 2, and 4 appear in each row (correct answer: `2'): 

\begin{center}
[6] [2] [4]
\linebreak[4][2] [4] [6]
\linebreak[4][4] [6] [?]
\end{center}

\noindent The progression rule was defined by a progressive increase or decrease in value, in units of either 1 or 2, across either rows or columns. In the following example, digits increase by units of 2 across rows (correct answer: `9'):

\begin{center}
[3] [5] [7]
\linebreak[4][1] [3] [5]
\linebreak[4][5] [7] [?]
\end{center}

\noindent Transformation rules could be combined to form multi-rule problems, by assigning each rule to a particular spatial location within each cell. The following example shows a two-rule problem, in which the left digit in each cell is governed by a progression rule (digits decrease by units of 1 across columns), and the right digit in each cell is governed by a distribution-of-3 rule (correct answer: `4 9'):

\begin{center}
[7 1] [8 9] [6 3]
\linebreak[4][6 9] [7 3] [5 1]
\linebreak[4][5 3] [6 1] [ ? ]
\end{center}

\noindent Logic problems were defined by one of three rules: OR, XOR, and AND. In the OR rule, a particular row or column contained all entities that appeared in either of the other rows or columns. In the following example, the middle column contains all entities that appear either in the left or right columns (correct answer: `8'):

\begin{center}
[~ 7] [~ 7 4 ~] [4 ~]
\linebreak[4][9 7] [9 7 4 8] [4 8]
\linebreak[4][9 ~] [9 ~ ~ 8] [ ? ]
\end{center}

\noindent The XOR rule was the same, except that entities appearing in both of the other rows or columns were excluded. In the following example, only items that appear in either the left or middle columns, but not both, will appear in the right column (correct answer: `4 3'):

\begin{center}
[6 4] [6 1] [4 1]
\linebreak[4][6 1] [3 6] [1 3]
\linebreak[4][4 1] [1 3] [ ? ]
\end{center}

\noindent In the AND rule, a particular row or column contained only entities that appeared in both of the other rows or columns. In the following example, the left column contains only digits that appear in both the left and middle columns (correct answer: `9'): 

\begin{center}
[2 9 7] [1 9 7] [~ 9 7]
\linebreak[4][2 9 5] [1 9 5] [~ 9 5]
\linebreak[4][2 9 ~] [1 9 ~] [~ ? ~]
\end{center}

\noindent For some logic problems, the within-cell spatial position of corresponding elements was aligned, as in the previously presented OR and AND problems. In other logic problems, corresponding elements were spatially permuted. The following example (involving an OR rule) illustrates how this makes it more difficult to intuitively grasp the underlying rule (correct answer: `0'):

\begin{center}
[~ 1] [7 1 ~ ~] [7 ~]
\linebreak[4][1 0] [5 0 7 1] [7 5]
\linebreak[4][0 ~] [~ ~ 0 5] [ ? ]
\end{center}

Within each problem type (one- through five-rule and logic problems), there were a number of specific problem subtypes. There were 6 one-rule subtypes, 6 two-rule subtypes, and 10 subtypes for three-rule, four-rule, five-rule, and logic problems. We generated 100 instances of each subtype (except in the case of progression problems, for which there were fewer possible problem instances). The one-rule problem subtypes consisted of a row-wise constant problem, a column-wise constant problem, two distribution-of-3 problems, and two progression problems (one with an increment of 1 and one with an increment of 2). The two- and three-rule problem subtypes consisted of all possible combinations of two or three rules (allowing for the same rule to be used multiple times within each problem). The four- and five-rule problem subtypes were sampled from the set of all possible combinations of four or five rules. There were five spatially aligned logic problem subtypes, and five spatially permuted logic problem subtypes. Three out of each of these five subtypes were OR problems (defined by the row or column in which the set union appeared), and the other two were AND and XOR problems.

For each problem, we also procedurally generated a set of 7 distractor choices, making for a set of 8 total answer choices. Distractors were generated using different methods for the transformation and logic problems. These methods were chosen based on the approach of Matzen et al.~\cite{matzen2010recreating}, who performed an analysis of the answer choices in the original SPM. For transformation problems, the following methods were used to generate distractors:

\begin{enumerate}
    \item Sample a random cell from the problem.
    \item Sample a random cell from the problem, sample a random digit within that cell, and apply an increment or decrement of either 1 or 2.
    \item Start with the correct answer, apply an increment or decrement of either 1 or 2 to a randomly sampled digit.
    \item Randomly sample a previously generated distractor for this problem, apply an increment or decrement of either 1 or 2 to a randomly sampled digit.
    \item Randomly generate a new answer choice (with the appropriate number of digits given the problem type).
\end{enumerate}

\noindent For multi-rule transformation problems, the following additional methods were also used:

\begin{enumerate}
    \item Start with the correct answer, randomly permute the digits.
    \item Sample a random cell from the problem, randomly permute the digits.
    \item Randomly sample a previously generated distractor for this problem, randomly permute the digits.
    \item Randomly sample digits from multiple cells within the problem and combine.
    \item Randomly sample digits from previously generated distractors for this problem and combine.
\end{enumerate}

\noindent For logic problems, distractors were generated by sampling from the set of all possible subsets of elements that appeared within the problem, including the empty set (the correct answer was an empty set on some logic problems), but excluding the correct answer. For spatially permuted logic problems, the spatial position of the elements within each distractor was randomly permuted. For spatially aligned logic problems, the order of the elements within each distractor was chosen so as to be consistent with the order that they appeared in the problem.

\subsubsection{Human behavioral experiments}

Human behavioral data was collected in two online experiments. All experiments were approved by the UCLA Institutional Review Board (IRB protocol \#22-000841, approved May 17, 2022), and all participants provided informed consent. All participants were UCLA undergraduates. Forty-three participants completed the first experiment, but three participants were excluded from analysis due to the fact that they got nearly every answer incorrect, and produced an apparently random pattern of responses (e.g. random permutations of the same three digits for all problems). The remaining 40 participants (31 female, 18-35 years old, average age = 21.3 years old) were included in our analysis. Forty-seven participants (37 female, 18-42 years old, average age = 21.2 years old) completed the second experiment. No statistical methods were used to pre-determine sample sizes. There was no overlap between the participants in the first and second experiments. Participants received course credit for their participation.

In both experiments, participants were first presented with a set of instructions, and a single one-rule example problem involving a constant rule. For each problem, participants first generated a free-response answer, and then selected from the set of answer choices. Problems were presented in a spatially arranged matrix format, as they appear in Figure~\ref{RPM_vs_digit_mat}. Problems remained on the screen until participants made a response.

In the first experiment (Figure~\ref{exp1_results}), participants were presented with one-, two-, three-rule, and logic problems. There were 6 problem subtypes each for the one- and two-rule problems, and 10 problem subtypes each for the three-rule and logic problems, making for 32 problem subtypes in total. Participants received these problem subtypes in random order. Each participant received randomly sampled instances of each problem subtype. 

In the second experiment (Supplementary Figure~\ref{prog_5rule}), participants were presented with one- through five-rule problems. There were 6 problem subtypes each for the one- and two-rule problems, and 10 problem subtypes each for the three- through five-rule problems, making for 42 problem subtypes in total. Problems were presented in order of increasing complexity, with all one-rule problem subtypes first, followed by all two-rule problem subtypes, and so on. For one-rule problems, the two constant problems were presented first, followed by the two distribution-of-3 problems, followed by the two progression problems.

\subsubsection{Evaluating GPT-3}

GPT-3 was evaluated on the Digit Matrices by presenting each complete problem as a prompt, including brackets and line breaks, followed by an open bracket at the start of the final cell. For example, the three-rule problem in Figure~\ref{digit_mat_example} would be presented to GPT-3 in the following format:

\begin{center}
    [5 9 3] [8 9 2] [1 9 7]\textbackslash n[8 4 7] [1 4 3] [5 4 2]\textbackslash n[1 2 2] [5 2 7] [
\end{center}

\noindent GPT-3's generated responses were truncated at the point where a closing bracket was generated. For logic problems, generated answers were counted as correct if they contained the correct set of digits, regardless of their order. For transformation problems, generated answers were only counted as correct if they contained the correct digits in the correct order. The same criteria were applied when evaluating human responses.

To evaluate GPT-3's multiple-choice performance, for each answer choice, the choice was appended to the problem followed by a closing bracket, and presented to GPT-3 as a prompt. The average log probability of the tokens corresponding to the answer choice (not counting the brackets) was computed. The answer choice with the highest average log probability was treated as GPT-3's selection.

In our primary evaluation (Figure~\ref{exp1_results}), GPT-3 was presented with 40 problem instances from each of the 32 problem subtypes used in the first human behavioral experiment. GPT-3 solved each one zero-shot (without any fine-tuning or in-context learning). 

We also evaluated how GPT-3 performed when presented with problems in order of increasing complexity (Supplementary Figure~\ref{prog_5rule}).  GPT-3 performed 20 runs on this task. For each run, GPT-3 was presented with a series of the same 42 problem subtypes used in the second human behavioral experiment (with different instances of these subtypes in each run). After GPT-3 answered each problem, the selected multiple-choice answer was appended to the problem, and the combined problem and answer choice were recursively appended to the prompt for the next problem. This meant that the size of the prompt grew with each problem. For some of the final five-rule problems, the prompt exceeded the size of GPT-3's context window (4096 tokens). When this occurred, problems from the beginning of the context window were deleted until the entire prompt fit within the window. This resulted in the deletion of a few one-rule problems from the beginning of the prompt. For one-rule problems, the two constant problems were presented first, followed by the two distribution-of-3 rules, followed by the two progression problems.

\subsubsection{Statistical analyses}

Results were analyzed using both regression and correlation analyses. Logistic regression analyses were carried out at the individual trial level, with each data point corresponding to a particular trial from a particular participant (or GPT-3). The dependent variable in all regression analyses was a binary variable coding for whether a particular response was correct or incorrect.

For the first digit matrix experiment, we fit separate regression models for generative vs. multiple-choice responses. Two predictors were used: problem type (one-, two-, three-rule, and logic problems), and a binary predictor coding for GPT-3 vs. human participants. We also performed more fine-grained analyses for generative responses within each problem type. These analyses were performed separately for GPT-3 vs. human responses. For two-rule problems, a single binary predictor coded for whether a problem contained a progression rule. For three-rule problems, a single predictor coded for the number of unique rules present in a given problem. For logic problems, a binary predictor coding for whether a problem was spatially aligned vs. permuted.

We also fit regression models comparing the results of the first and second experiments. These analyses  were performed separately for GPT-3 vs. human responses, and only included responses for one- to three-rule problems (since these were the only problem types in common between the two experiments). Two predictors were used: problem type (one-, two-, and three-rule problems), and experiment (experiment 1 vs. 2).

Correlation analyses were carried out by correlating the accuracy for GPT-3 vs. human participants across all 32 problem subtypes.

\subsection{Letter string analogies}

\subsubsection{Problem set}

Each letter string analogy problem involved one of six transformation types: sequence extension, successor, predecessor, removing a redundant letter, fixing an alphabetic sequence, and sorting. In the sequence extension transformation, the source involved an alphabetically ordered sequence of four letters followed by an extension of this sequence involving five letters, as in the following example:

\begin{center}
[a b c d] [a b c d e]
\end{center}

\noindent In the successor transformation, the source involved an alphabetically ordered sequence of four letters, followed by that same sequence, but with the final letter replaced by its successor, as in the following example:

\begin{center}
[a b c d] [a b c e]
\end{center}

\noindent In the predecessor transformation, the source involved an alphabetically ordered sequence of four letters, followed by that same sequence, but with the first letter replaced by its predecessor, as in the following example:

\begin{center}
[b c d e] [a c d e]
\end{center}

\noindent In the transformation involving removal of a redundant letter, the source involved an alphabetically ordered sequence of five letters with one letter repeated, followed by that same sequence with the redundant letter removed, as in the following example:

\begin{center}
[a b b c d e] [a b c d e]
\end{center}

\noindent In the transformation involving fixing an alphabetic sequence, the source involved an alphabetically ordered sequence of five letters with one out-of-place letter (not part of the alphabetic sequence), followed by that same sequence with the out-of-place letter replaced, as in the following example:

\begin{center}
[a b c w e] [a b c d e]
\end{center}

\noindent In the sorting transformation, the source involved an alphabetically ordered sequence of five letters with the position of two letters swapped, followed by a sorted version of the same sequence, as in the following example:

\begin{center}
[a d c b e] [a b c d e]
\end{center}

Problems involved varying degrees of generalization between the source and target. In the zero-generalization problems, the target involved a different instance of the source transformation (instantiated with different letters). Transformation parameters (e.g., the location of the redundant letter) were independently sampled for source and target.

Generalization problems involved generalizations sampled from the following set of generalization types: generalization from letters to numbers, grouping, generalization to a longer target, reversed order, interleaved distractors, and generalization to a larger interval. In the letter-to-number generalization, target letters were replaced by numbers corresponding to their alphabetic indices, as in the following example:

\begin{center}
[a b c d] [a b c d e]
\linebreak[4][7 8 9 10] [ ~ ? ~ ]
\end{center}

\noindent In the grouping generalization, target letters were replaced by groups with two instances of each letter, as in the following example:

\begin{center}
[a b c d] [a b c d e]
\linebreak[4][i i j j k k l l] [ ~ ? ~ ]
\end{center}

\noindent In the longer target generalization, the target sequence was replaced with a sequence that was twice as long as the source, as in the following example:

\begin{center}
[a b c d] [a b c d e]
\linebreak[4][i j k l m n o p] [ ~ ? ~ ]
\end{center}

\noindent In the reversed order generalization, the order of the target letters was reversed relative to the source, as in the following example:

\begin{center}
[a b c d] [a b c d e]
\linebreak[4][l k j i] [ ~ ? ~ ]
\end{center}

\noindent In the interleaved distractor generalization, the letter `x' was interleaved between each letter in the target sequence, as in the following example:

\begin{center}
[a b c d] [a b c d e]
\linebreak[4][i x j x k x l x] [ ~ ? ~ ]
\end{center}

\noindent In the larger interval generalization, the sequence of target letters was replaced with a sequence involving an interval of size 2, as in the following example:

\begin{center}
[a b c d] [a b c d e]
\linebreak[4][i k m o] [ ~ ? ~ ]
\end{center}

Each transformation type could be combined with any generalization type. Multiple generalizations could also be combined together. Generalization problems contained between one and three generalizations. We generated a set of 600 zero-generalization problems (involving 100 problems with each transformation type), 600 one-generalization problems (involving 100 problems with each generalization type, with randomly sampled transformation type), and 600 problems each with two and three generalizations (with randomly sampled combinations of transformation and generalization type).

We also generated a separate set of problems involving generalization from letters to real-world concepts. In these problems, the source instantiated a transformation using letters, and the target instantiated that same transformation using real-world instances of successorship. These problems involved shorter sequences (maximum length of four), due to the difficulty of identifying real-world instances of successorship with more than four points. The following sequences were used:

\begin{center}
cold cool warm hot
\linebreak[4]love like dislike hate
\linebreak[4]jack queen king ace
\linebreak[4]penny nickel dime quarter
\linebreak[4]second minute hour day
\end{center}

\noindent The transformation types included sequence extension, successor, predecessor, and sorting. No other generalizations were applied to these problems. We generated 100 problems with each transformation type.

\subsubsection{Evaluating GPT-3}

We presented letter string analogies to GPT-3 using the prompt `Let's try to complete the pattern:', similar to~\cite{kojima2022large}. We also formatted each analogy problem using brackets and line breaks, similar to the presentation format of the Digit Matrices. The presentation format is illustrated in the following example:

\begin{center}
    Let's try to complete the pattern:\textbackslash n\textbackslash n[a b c d] [a b c e]\textbackslash n[i j k l] [
\end{center}

GPT-3's generated responses were truncated at the point where a closing bracket was generated. We also evaluated GPT-3 with two alternative problem formats: 1) no prompt, and 2) a sentence format, as in the following example:

\begin{center}
    If a b c d changes to a b c e, then i j k l should change to 
\end{center}

For this format, GPT-3's generated responses were truncated at the point where a period was generated. We evaluated GPT-3 on 300 zero-generalization problems (50 problems for each transformation type), 300 one-generalization problems (50 problems for each generalization type), and 300 problems each with two and three generalizations. We also evaluated GPT-3 on 50 real-world concept generalization problems for each transformation type.

\subsubsection{Human behavioral experiment}

Human behavioral data was collected in an online experiment. The experiment was approved by the UCLA Institutional Review Board (IRB protocol \#22-000841, approved May 17, 2022), and all participants provided informed consent. All participants were UCLA undergraduates. Fifty-seven participants (50 female, 18-35 years old, average age = 21.1 years old) completed the experiment. No statistical methods were used to pre-determine sample sizes. Participants received course credit for their participation.

Participants were first presented with a set of instructions, and the following example problem (not involving any of the transformations or generalizations employed in the actual experiment):

\begin{center}
[a a a] [b b b]
\linebreak[4][c c c] [~ ? ~]
\end{center}

Each participant completed 28 problems, including 6 zero-generalization problems (1 problem for each transformation type), 6 one-generalization problems (1 problem for each generalization type), 6 problems each with two and three generalizations, and 4 real-world concept generalization problems (1 for each transformation type). The specific problem instances were randomly sampled for each participant, and participants received these problems in a random order. Participants generated a free response for each problem. 

\subsubsection{Statistical analyses}

Results were analyzed using both regression and correlation analyses. Logistic regression analyses were carried out at the individual trial level, with each data point corresponding to a particular trial from a particular participant (or GPT-3). The dependent variable in all regression analyses was a binary variable coding for whether a particular response was correct or incorrect. 

Separate analyses were performed for problems that only involved alphanumeric characters vs. those that involved real-world concepts. For problems involving alphanumeric characters, a regression model was fit with two predictors: number of generalizations (zero to three), and a binary predictor coding for GPT-3 vs. human participants. We also fit regression models at each generalization level with a single binary predictor coding for GPT-3 vs. human participants. For real-world concept problems, a regression model was fit with a predictor coding for GPT-3 vs. human participants.

For correlation analyses, problem subtypes were defined based on each combination of transformation type and generalization type. The accuracy for each subtype was computed for GPT-3 vs. human participants, and these values were subjected to correlation analysis. There were only a few examples of some problem subtypes (across all participants), especially for problems with more generalizations (the space of possible subtypes grows exponentially with the number of generalizations). We only included subtypes for which there were at least five trials from human participants (across all participants) and five trials from GPT-3. Out of the 252 possible problem subtypes, 41 subtypes met this criterion and were included in the analysis.

\subsection{Four-term verbal analogies}

We evaluated GPT-3 on four separate four-term analogy datasets~\cite{sternberg1980developmental,turney2003combining,lu2019emergence,jones2022differential}. The UCLA-VAT dataset contains 80 problems, with four relation types: categorical (B/D is a member of the category A/C), functional (A/C is the function of B/D), antonym, and synonym. There are 20 problems for each relation type. Each problem contains two answer choices for the final term (D and D'). We evaluated GPT-3 by presenting the problem along with each possible answer choice (A:B::C:D or A:B::C:D'), using the standard colon notation, and selected the answer choice for which GPT-3 assigned a higher log probability to the final term. The problem and GPT-3's selected answer were then recursively appended to the prompt for the next problem. The problems were presented in a shuffled order. We compared against human behavioral data from~\cite{lu2019emergence} (N=57, UCLA undergraduates). Example problems from each of the four relation categories are shown below:

\begin{center}
    \textbf{Categorical}\\
    vegetable : cabbage :: insect : ?\\
    1. beetle   2. frog 
\end{center}

\begin{center}
    \textbf{Function}\\
    drive : car :: burn : ?\\
    1. wood   2. fire 
\end{center}

\begin{center}
    \textbf{Antonym}\\
    love : hate :: rich : ?\\
    1. poor   2. wealthy 
\end{center}

\begin{center}
    \textbf{Synonym}\\
    rob : steal :: cry : ?\\
    1. weep   2. laugh
\end{center}

The dataset of Sternberg and Nigro~\cite{sternberg1980developmental} contains 200 problems, including 40 problems for each of five relation types: categorical, functional, antonym, synonym, and linear order. We evaluated GPT-3 in the same way that we did for UCLA-VAT, and compared against human behavioral data from ~\cite{sternberg1980developmental} (N=20, Yale undergraduates). An example problem illustrating the linear order relation type is shown below (the categorical, functional, antonym, and synonym problems were similar to those from the UCLA VAT):

\begin{center}
    \textbf{Linear order}\\
    month : year :: inch : ?\\
    1. foot   2. length
\end{center}

The dataset of SAT problems from Turney et al.~\cite{turney2003combining} contains 374 problems, covering a range of different relation types. Each problem contains five answer choices for both C and D terms (including the correct answer). We evaluated GPT-3 by presenting each of the five possible analogies for each problem, and selecting the choice for which the C and D terms were assigned the highest log probability. The problem, and GPT-3's choice, were then appended to the prompt for the next problem. We compared against an estimate of the average performance level for high school students taking the SAT (see~\cite{turney2005corpus}).

The dataset of Jones et al.~\cite{jones2022differential} contains 120 problems, including 40 problems for each of three relation types: categorical, causal, and compositional. Half of these problems are categorized as semantically near (A and B are similar to C and D), and half are categorized as semantically far (A and B are dissimilar to C and D). Each problem contains two answer choices. We evaluated GPT-3 in the same way that we did for UCLA-VAT, and compared against human behavioral data from ~\cite{jones2022differential} (N=241, Wayne State University undergraduates). Example problems for each of the three relation categories are shown below:

\begin{center}
    \textbf{Categorical}\\
    diesel : fuel :: bed : ?\\
    1. furniture   2. pillow
\end{center}

\begin{center}
    \textbf{Causal}\\
    motion : sickness :: drought : ?\\
    1. famine   2. rain
\end{center}

\begin{center}
    \textbf{Compositional}\\
    steel : scissors :: apple : ?\\
    1. cider   2. tree
\end{center}

\subsection{Story analogies}

\subsubsection{Materials}
\label{example_story_materials}

All story analogy materials were taken from a problem set created by Gentner et al.~\cite{gentner1993roles} (from their Experiment 2), and included in a verbal analogy inventory~\cite{ichien2020verbal}. These materials involve 18 source stories. Each source story is accompanied by four potential target stories, forming four conditions: correct and incorrect near analogies (respectively termed `literal similarity' and `mere appearance' by Gentner et al.), both involving similar entities and first-order relations as the source, while differing from each other in higher-order causal relations; and correct and incorrect far analogies (respectively termed `true analogy' and `false analogy' by Gentner et al.), both involving similar first-order relations as the source but distinct entities, while differing from each other in causal relations. An example source story, along with target stories from each condition, is presented below:

\begin{quote}

\noindent Source story: Karla, an old hawk, lived at the top of a tall oak tree. One afternoon, she saw a hunter on the ground with a bow and some crude arrows that had no feathers. The hunter took aim and shot at the hawk but missed. Karla knew the hunter wanted her feathers so she glided down to the hunter and offered to give him a few. The hunter was so grateful that he pledged never to shoot at a hawk again. He went off and shot deer instead. 

\vspace{3mm}

\noindent Near analogy -- correct target story: Once there was an eagle named Zerdia who nested on a rocky cliff. One day she saw a sportsman coming with a crossbow and some bolts that had no feathers. The sportsman attacked but the bolts missed. Zerdia realized that the sportsman wanted her tailfeathers so she flew down and donated a few of her tailfeathers to the sportsman. The sportsman was pleased. He promised never to attack eagles again. 

\vspace{3mm}

\noindent Near analogy -- incorrect target story: Once there was an eagle named Zerdia who donated a few of her tailfeathers to a sportsman so he would promise never to attack eagles. One day Zerdia was nesting high on a rocky cliff when she saw the sportsman coming with a crossbow. Zerdia flew down to meet the man, but he attacked and felled her with a single bolt. As she fluttered to the ground Zerdia realized that the bolt had her own tailfeathers on it. 

\vspace{3mm}

\noindent Far analogy -- correct target story: Once there was a small country called Zerdia that learned to make the world's smartest computer. One day Zerdia was attacked by its warlike neighbor, Gagrach. But the missiles were badly aimed and the attack failed. The Zerdian government realized that Gagrach wanted Zerdian computers so it offered to sell some of its computers to the country. The government of Gagrach was very pleased. It promised never to attack Zerdia again. 

\vspace{3mm}

\noindent Far analogy -- incorrect target story: Once there was a small country called Zerdia that learned to make the world's smartest computer. Zerdia sold one of itssupercomputers to its neighbor, Gagrach, so Gagrach would promise never to attack Zerdia. But one day Zerdia was overwhelmed by a surprise attack from Gagrach. As it capitulated the crippled government of Zerdia realized that the attacker's missiles had been guided by Zerdian supercomputers. 

\end{quote}

\subsubsection{Human behavioral experiment}

Human behavioral data was collected in an online experiment. The experiment was approved by the UCLA Institutional Review Board (IRB protocol \#22-000841, approved May 17, 2022), and all participants provided informed consent. All participants were UCLA undergraduates. Fifty-four participants (47 female, 18-44 years old, average age = 20.7 years old) completed the experiment. No statistical methods were used to pre-determine sample sizes. Participants received course credit for their participation.

After receiving instructions, participants were presented with 18 trials, each involving a different source story. On each trial, participants were presented with a source story (referred to as `Story 1'), followed by two target stories (referred to as `Story A' and `Story B'), and asked `Which of Story A and Story B is a better analogy to Story 1?'. Participants could select either story A or story B, or could indicate that they were both equally analogous. Accuracy was computed as the proportion of trials for which participants selected the correct target story.

On half of the trials, the target stories were from the near analogy conditions. On the other half of the trials, the target stories were from the far analogy conditions. The order of the two target stories was randomly shuffled on all trials.

\subsubsection{Evaluating GPT-3}

GPT-3 was evaluated by entering stories directly into the \href{https://beta.openai.com/playground}{OpenAI playground}. For each source story, GPT-3 was evaluated on both the near analogy comparison, and the far analogy comparison, and was also evaluated on both possible orderings for each pair of target stories, resulting in $18\times2\times2=72$ total comparisons. For each comparison, the stories were presented in the following format:
\newline
\newline
\indent Consider the following story:
\newline
\newline
\indent Story 1: $\ll$ source story text $\gg$
\newline
\newline
\indent Now consider two more stories:
\newline
\newline
\indent Story A: $\ll$ target story A text $\gg$
\newline
\newline
\indent Story B: $\ll$ target story B text $\gg$
\newline
\newline
\indent Which of Story A and Story B is a better analogy to Story 1? 

\indent Is the best answer Story A, Story B, or both are equally analogous?
\newline
\newline
\noindent where $\ll$ source story text $\gg$, $\ll$ target story A text $\gg$, and $\ll$ target story B text $\gg$ were replaced by the text for the corresponding stories. In addition to answering the forced-choice question, GPT-3 sometimes spontaneously produced explanations, but only the forced-choice response was used in our analysis. GPT-3's context window was cleared after obtaining the results of each comparison.

\subsubsection{Evaluating GPT-4}

GPT-4 was evaluated by entering stories directly into the \href{https://chat.openai.com/chat}{ChatGPT web interface}. GPT-4 was evaluated on the same 72 problems, using the same format as was used for GPT-3. GPT-4's context window was cleared after obtaining the results of each comparison.

\subsubsection{Statistical analyses}

The task performed by both GPT-3 and human participants involved a three-choice discrimination (Story A is more analogous, Story B is more analogous, both are equally analogous). Statistical analyses were carried out to determine whether this discrimination was made at a level greater than expected from chance alone. To be conservative, we assumed a chance performance level of 50\% accuracy. For GPT-3, a binomial test was performed (using data at the individual trial level). For human participants, a one-sample t-test was performed (using data averaged at the individual subject level). These analyses were carried out separately for the near analogy and far analogy conditions.

To compare GPT-3 with human performance, a logistic regression analysis was carried out at the individual trial level. The dependent variable was a binary variable coding for whether a particular response was correct or incorrect. A single binary predictor coded for GPT-3 vs. human responses.

\subsection{Analogical problem-solving}

Problems were entered directly into the \href{https://beta.openai.com/playground}{OpenAI playground}. Materials were taken from~\cite{gick1980analogical} and~\cite{holyoak1984development}. All prompts and responses are shown in Supplementary Section~\ref{problem_solving_results}. Each subsection shows the results for a single continuous session, with GPT-3's responses presented in bold text. Responses were not truncated or curated in any way.

\section*{Data Availability}

Data for all human behavioral experiments, along with the Digit Matrices, letter string analogy, and UCLA VAT problem sets, can be downloaded from:

\begin{center}
    \href{https://github.com/taylorwwebb/emergent_analogies_LLM}{https://github.com/taylorwwebb/emergent\_analogies\_LLM}
\end{center}

\noindent The four-term verbal analogy problem sets from Sternberg and Nigro~\cite{sternberg1980developmental} and Jones et al.~\cite{jones2022differential}, and the story analogy materials from Gentner et al.~\cite{gentner1993roles} can be downloaded from:

\begin{center}
    \href{http://cvl.psych.ucla.edu/resources/AnalogyInventory.zip}{http://cvl.psych.ucla.edu/resources/AnalogyInventory.zip}
\end{center}

\noindent Information about the problem set of SAT four-term verbal analogies from Turney et al.~\cite{turney2003combining} can be found at:

\begin{center}
    \href{https://aclweb.org/aclwiki/SAT\_Analogy\_Questions\_(State\_of\_the\_art)}{https://aclweb.org/aclwiki/SAT\_Analogy\_Questions\_(State\_of\_the\_art)}
\end{center}

\section*{Code Availability}

Code for all simulations can be downloaded from:

\begin{center}
    \href{https://github.com/taylorwwebb/emergent_analogies_LLM}{https://github.com/taylorwwebb/emergent\_analogies\_LLM}
\end{center}

\section*{Acknowledgements}

We would like to thank Bryor Snefjella and Peter Turney for helpful feedback and discussions. Preparation of this paper was supported by NSF grant IIS-1956441 and AFOSR MURI grant FA9550-22-1-0380 to H.L. 

\section*{Author Contributions Statement}

T.W., K.J.H., and H.L. conceived project and planned experiments. T.W. implemented experiments and analyzed results. T.W., K.J.H., and H.L. drafted manuscript.

\section*{Competing Interests Statement}

The authors declare no competing interests.

\bibliographystyle{unsrt}
\bibliography{main.bib}

\begin{thebibliography}{10}

\bibitem{holyoak2012analogy}
Keith~J Holyoak.
\newblock Analogy and relational reasoning.
\newblock In Keith~J Holyoak and Robert~G Morrison, editors, {\em Oxford
  handbook of thinking and reasoning}, pages 234--259. Oxford University Press,
  2012.

\bibitem{bassok2012problem}
Miriam Bassok and Laura~R Novick.
\newblock Problem solving.
\newblock In Keith~J Holyoak and Robert~G Morrison, editors, {\em Oxford
  handbook of thinking and reasoning}, pages 413--432. Oxford University Press,
  2012.

\bibitem{dunbar2012scientific}
Kevin~N Dunbar and David Klahr.
\newblock Scientific thinking and reasoning.
\newblock In Keith~J Holyoak and Robert~G Morrison, editors, {\em Oxford
  handbook of thinking and reasoning}, pages 701--718. Oxford University Press,
  2012.

\bibitem{cattell1971abilities}
Raymond~B Cattell.
\newblock {\em Abilities: Their structure, growth, and action}.
\newblock Houghton Mifflin, 1971.

\bibitem{snow1984topography}
Richard~E Snow, Patrick~C Kyllonen, Brachia Marshalek, et~al.
\newblock The topography of ability and learning correlations.
\newblock {\em Advances in the psychology of human intelligence}, 2(S 47):103,
  1984.

\bibitem{mitchell2021abstraction}
Melanie Mitchell.
\newblock Abstraction and analogy-making in artificial intelligence.
\newblock {\em Annals of the New York Academy of Sciences}, 1505(1):79--101,
  2021.

\bibitem{barrett2018measuring}
David Barrett, Felix Hill, Adam Santoro, Ari Morcos, and Timothy Lillicrap.
\newblock Measuring abstract reasoning in neural networks.
\newblock In {\em International conference on machine learning}, pages
  511--520. PMLR, 2018.

\bibitem{zhang2019raven}
Chi Zhang, Feng Gao, Baoxiong Jia, Yixin Zhu, and Song-Chun Zhu.
\newblock Raven: A dataset for relational and analogical visual reasoning.
\newblock In {\em Proceedings of the IEEE/CVF Conference on Computer Vision and
  Pattern Recognition}, pages 5317--5327, 2019.

\bibitem{hill2019learning}
Felix Hill, Adam Santoro, David G.~T. Barrett, Ari~S. Morcos, and Timothy~P.
  Lillicrap.
\newblock Learning to make analogies by contrasting abstract relational
  structure.
\newblock In {\em 7th International Conference on Learning Representations,
  {ICLR}}, 2019.

\bibitem{wu2020scattering}
Yuhuai Wu, Honghua Dong, Roger Grosse, and Jimmy Ba.
\newblock The scattering compositional learner: Discovering objects,
  attributes, relationships in analogical reasoning.
\newblock {\em arXiv preprint arXiv:2007.04212}, 2020.

\bibitem{hersche2023neuro}
Michael Hersche, Mustafa Zeqiri, Luca Benini, Abu Sebastian, and Abbas Rahimi.
\newblock A neuro-vector-symbolic architecture for solving raven’s
  progressive matrices.
\newblock {\em Nature Machine Intelligence}, 2023.

\bibitem{subhramondal2023objects}
Shanka Subhra~Mondal, Taylor~W Webb, and Jonathan~D Cohen.
\newblock Learning to reason over visual objects.
\newblock In {\em 11th International Conference on Learning Representations,
  {ICLR}}, 2023.

\bibitem{brown2020language}
Tom Brown, Benjamin Mann, Nick Ryder, Melanie Subbiah, Jared~D Kaplan, Prafulla
  Dhariwal, Arvind Neelakantan, Pranav Shyam, Girish Sastry, Amanda Askell,
  et~al.
\newblock Language models are few-shot learners.
\newblock {\em Advances in neural information processing systems},
  33:1877--1901, 2020.

\bibitem{mahowald2023dissociating}
Kyle Mahowald, Anna~A Ivanova, Idan~A Blank, Nancy Kanwisher, Joshua~B
  Tenenbaum, and Evelina Fedorenko.
\newblock Dissociating language and thought in large language models: a
  cognitive perspective.
\newblock {\em arXiv preprint arXiv:2301.06627}, 2023.

\bibitem{raven1938raven}
John~C Raven.
\newblock {\em Progressive matrices: A perceptual test of intelligence,
  individual form}.
\newblock London: Lewis, 1938.

\bibitem{hofstadterMitchell94}
Douglas~R Hofstadter and Melanie Mitchell.
\newblock The copycat project: A model of mental fluidity and analogy-making.
\newblock In Keith~J Holyoak and J~A Barnden, editors, {\em Advances in
  connectionist and neural computation theory}, volume~2, page 31–112. Ablex,
  Norwood, NJ, 1994.

\bibitem{lu2019emergence}
Hongjing Lu, Ying~Nian Wu, and Keith~J Holyoak.
\newblock Emergence of analogy from relation learning.
\newblock {\em Proceedings of the National Academy of Sciences},
  116(10):4176--4181, 2019.

\bibitem{gentner1993roles}
Dedre Gentner, Mary~Jo Rattermann, and Kenneth~D Forbus.
\newblock The roles of similarity in transfer: Separating retrievability from
  inferential soundness.
\newblock {\em Cognitive psychology}, 25(4):524--575, 1993.

\bibitem{sternberg1980developmental}
Robert~J Sternberg and Georgia Nigro.
\newblock Developmental patterns in the solution of verbal analogies.
\newblock {\em Child Development}, pages 27--38, 1980.

\bibitem{turney2003combining}
Peter~D Turney, Michael~L Littman, Jeffrey Bigham, and Victor Shnayder.
\newblock Combining independent modules to solve multiple-choice synonym and
  analogy problems.
\newblock In {\em Proceedings of the International Conference on Recent
  Advances in Natural Language Processing (RANLP-03)}, pages 482--489, 2003.

\bibitem{jones2022differential}
Lara~L Jones, Matthew~J Kmiecik, Jessica~L Irwin, and Robert~G Morrison.
\newblock Differential effects of semantic distance, distractor salience, and
  relations in verbal analogy.
\newblock {\em Psychonomic bulletin \& review}, 29(4):1480--1491, 2022.

\bibitem{gick1980analogical}
Mary~L Gick and Keith~J Holyoak.
\newblock Analogical problem solving.
\newblock {\em Cognitive psychology}, 12(3):306--355, 1980.

\bibitem{holyoak1984development}
Keith~J Holyoak, Ellen~N Junn, and Dorrit~O Billman.
\newblock Development of analogical problem-solving skill.
\newblock {\em Child development}, pages 2042--2055, 1984.

\bibitem{dasgupta2022language}
Ishita Dasgupta, Andrew~K Lampinen, Stephanie~CY Chan, Antonia Creswell,
  Dharshan Kumaran, James~L McClelland, and Felix Hill.
\newblock Language models show human-like content effects on reasoning.
\newblock {\em arXiv preprint arXiv:2207.07051}, 2022.

\bibitem{srivastava2022beyond}
Aarohi Srivastava, Abhinav Rastogi, Abhishek Rao, Abu Awal~Md Shoeb, Abubakar
  Abid, Adam Fisch, Adam~R Brown, Adam Santoro, Aditya Gupta, Adri{\`a}
  Garriga-Alonso, et~al.
\newblock Beyond the imitation game: Quantifying and extrapolating the
  capabilities of language models.
\newblock {\em arXiv preprint arXiv:2206.04615}, 34:1877–--1901, 2022.

\bibitem{wei2022emergent}
Jason Wei, Yi~Tay, Rishi Bommasani, Colin Raffel, Barret Zoph, Sebastian
  Borgeaud, Dani Yogatama, Maarten Bosma, Denny Zhou, Donald Metzler, et~al.
\newblock Emergent abilities of large language models.
\newblock {\em arXiv preprint arXiv:2206.07682}, 2022.

\bibitem{chan2022data}
Stephanie~CY Chan, Adam Santoro, Andrew~Kyle Lampinen, Jane~X Wang, Aaditya~K
  Singh, Pierre~Harvey Richemond, James McClelland, and Felix Hill.
\newblock Data distributional properties drive emergent in-context learning in
  transformers.
\newblock In {\em Advances in Neural Information Processing Systems}, 2022.

\bibitem{binz2023using}
Marcel Binz and Eric Schulz.
\newblock Using cognitive psychology to understand gpt-3.
\newblock {\em Proceedings of the National Academy of Sciences},
  120(6):e2218523120, 2023.

\bibitem{vaswani2017attention}
Ashish Vaswani, Noam Shazeer, Niki Parmar, Jakob Uszkoreit, Llion Jones,
  Aidan~N Gomez, {\L}ukasz Kaiser, and Illia Polosukhin.
\newblock Attention is all you need.
\newblock {\em Advances in neural information processing systems},
  31:5998--6008, 2017.

\bibitem{chen2021evaluating}
Mark Chen, Jerry Tworek, Heewoo Jun, Qiming Yuan, Henrique Ponde de~Oliveira
  Pinto, Jared Kaplan, Harri Edwards, Yuri Burda, Nicholas Joseph, Greg
  Brockman, et~al.
\newblock Evaluating large language models trained on code.
\newblock {\em arXiv preprint arXiv:2107.03374}, 2021.

\bibitem{ouyang2022training}
Long Ouyang, Jeff Wu, Xu~Jiang, Diogo Almeida, Carroll~L Wainwright, Pamela
  Mishkin, Chong Zhang, Sandhini Agarwal, Katarina Slama, Alex Ray, et~al.
\newblock Training language models to follow instructions with human feedback.
\newblock {\em Advances in neural information processing systems},
  36:4299--4307, 2022.

\bibitem{matzen2010recreating}
Laura~E Matzen, Zachary~O Benz, Kevin~R Dixon, Jamie Posey, James~K Kroger, and
  Ann~E Speed.
\newblock Recreating {R}aven’s: Software for systematically generating large
  numbers of {R}aven-like matrix problems with normed properties.
\newblock {\em Behavior research methods}, 42(2):525--541, 2010.

\bibitem{matlen2020spatial}
Bryan~J Matlen, Dedre Gentner, and Steven~L Franconeri.
\newblock Spatial alignment facilitates visual comparison.
\newblock {\em Journal of Experimental Psychology: Human Perception and
  Performance}, 46(5):443, 2020.

\bibitem{kroger2004varieties}
James~K Kroger, Keith~J Holyoak, and John~E Hummel.
\newblock Varieties of sameness: The impact of relational complexity on
  perceptual comparisons.
\newblock {\em Cognitive Science}, 28(3):335--358, 2004.

\bibitem{halford1998processing}
Graeme~S Halford, William~H Wilson, and Steven Phillips.
\newblock Processing capacity defined by relational complexity: Implications
  for comparative, developmental, and cognitive psychology.
\newblock {\em Behavioral and brain sciences}, 21(6):803--831, 1998.

\bibitem{chalmers1992high}
David~J Chalmers, Robert~M French, and Douglas~R Hofstadter.
\newblock High-level perception, representation, and analogy: A critique of
  artificial intelligence methodology.
\newblock {\em Journal of Experimental \& Theoretical Artificial Intelligence},
  4(3):185--211, 1992.

\bibitem{hofstadter1995fluid}
Douglas~R Hofstadter.
\newblock {\em Fluid concepts and creative analogies: Computer models of the
  fundamental mechanisms of thought.}
\newblock Basic books, 1995.

\bibitem{lovett2017modeling}
Andrew Lovett and Kenneth Forbus.
\newblock Modeling visual problem solving as analogical reasoning.
\newblock {\em Psychological review}, 124(1):60, 2017.

\bibitem{mitchell1993analogy}
Melanie Mitchell.
\newblock {\em Analogy-making as perception: A computer model}.
\newblock MIT Press, 1993.

\bibitem{turney2005corpus}
Peter~D Turney and Michael~L Littman.
\newblock Corpus-based learning of analogies and semantic relations.
\newblock {\em Machine Learning}, 60:251--278, 2005.

\bibitem{ichien2020verbal}
Nicholas Ichien, Hongjing Lu, and Keith~J Holyoak.
\newblock Verbal analogy problem sets: An inventory of testing materials.
\newblock {\em Behavior research methods}, 52:1803--1816, 2020.

\bibitem{wason1968reasoning}
Peter~C Wason.
\newblock Reasoning about a rule.
\newblock {\em Quarterly journal of experimental psychology}, 20(3):273--281,
  1968.

\bibitem{gentner1983structure}
Dedre Gentner.
\newblock Structure-mapping: A theoretical framework for analogy.
\newblock {\em Cognitive science}, 7(2):155--170, 1983.

\bibitem{gpt4_tech_report}
OpenAI.
\newblock Gpt-4 technical report.
\newblock {\em arXiv preprint arXiv:2303.08774}, 2023.

\bibitem{duncker1945problem}
Karl Duncker.
\newblock On problem-solving.
\newblock {\em Psychological monographs}, 58(5):i, 1945.

\bibitem{holyoak1987surface}
Keith~J Holyoak and Kyunghee Koh.
\newblock Surface and structural similarity in analogical transfer.
\newblock {\em Memory \& cognition}, 15(4):332--340, 1987.

\bibitem{mcclelland2020placing}
James~L McClelland, Felix Hill, Maja Rudolph, Jason Baldridge, and Hinrich
  Sch{\"u}tze.
\newblock Placing language in an integrated understanding system: Next steps
  toward human-level performance in neural language models.
\newblock {\em Proceedings of the National Academy of Sciences},
  117(42):25966--25974, 2020.

\bibitem{marcus2001algebraic}
Gary~F Marcus.
\newblock {\em The algebraic mind: Integrating connectionism and cognitive
  science}.
\newblock MIT press, 2001.

\bibitem{lake2017building}
Brenden~M Lake, Tomer~D Ullman, Joshua~B Tenenbaum, and Samuel~J Gershman.
\newblock Building machines that learn and think like people.
\newblock {\em Behavioral and brain sciences}, 40, 2017.

\bibitem{webb2020learning}
Taylor~W Webb, Zachary Dulberg, Steven Frankland, Alexander Petrov, Randall
  O’Reilly, and Jonathan Cohen.
\newblock Learning representations that support extrapolation.
\newblock In {\em International conference on machine learning}, pages
  10136--10146. PMLR, 2020.

\bibitem{falkenhainer1989structure}
Brian Falkenhainer, Kenneth~D Forbus, and Dedre Gentner.
\newblock The structure-mapping engine: Algorithm and examples.
\newblock {\em Artificial intelligence}, 41(1):1--63, 1989.

\bibitem{lu2022probabilistic}
Hongjing Lu, Nicholas Ichien, and Keith~J Holyoak.
\newblock Probabilistic analogical mapping with semantic relation networks.
\newblock {\em Psychological Review}, 2022.

\bibitem{webb2022zero}
Taylor~W Webb, Shuhao Fu, Trevor Bihl, Keith~J Holyoak, and Hongjing Lu.
\newblock Zero-shot visual reasoning through probabilistic analogical mapping.
\newblock {\em arXiv preprint arXiv:2209.15087}, 2022.

\bibitem{smolensky1990tensor}
Paul Smolensky.
\newblock Tensor product variable binding and the representation of symbolic
  structures in connectionist systems.
\newblock {\em Artificial intelligence}, 46(1-2):159--216, 1990.

\bibitem{holyoak2000proper}
Keith~J Holyoak and John~E Hummel.
\newblock The proper treatment of symbols in a connectionist architecture.
\newblock {\em Cognitive dynamics: Conceptual change in humans and machines},
  229:263, 2000.

\bibitem{kriete2013indirection}
Trenton Kriete, David~C Noelle, Jonathan~D Cohen, and Randall~C O’Reilly.
\newblock Indirection and symbol-like processing in the prefrontal cortex and
  basal ganglia.
\newblock {\em Proceedings of the National Academy of Sciences},
  110(41):16390--16395, 2013.

\bibitem{webb2020emergent}
Taylor~W Webb, Ishan Sinha, and Jonathan~D. Cohen.
\newblock Emergent symbols through binding in external memory.
\newblock In {\em 9th International Conference on Learning Representations,
  {ICLR}}, 2021.

\bibitem{greff2020binding}
Klaus Greff, Sjoerd Van~Steenkiste, and J{\"u}rgen Schmidhuber.
\newblock On the binding problem in artificial neural networks.
\newblock {\em arXiv preprint arXiv:2012.05208}, 2020.

\bibitem{griffiths2020understanding}
Thomas~L Griffiths.
\newblock Understanding human intelligence through human limitations.
\newblock {\em Trends in Cognitive Sciences}, 24(11):873--883, 2020.

\bibitem{newell1958elements}
Allen Newell, John~Calman Shaw, and Herbert~A Simon.
\newblock Elements of a theory of human problem solving.
\newblock {\em Psychological review}, 65(3):151, 1958.

\bibitem{carpenter1990one}
Patricia~A Carpenter, Marcel~A Just, and Peter Shell.
\newblock What one intelligence test measures: a theoretical account of the
  processing in the raven progressive matrices test.
\newblock {\em Psychological review}, 97(3):404, 1990.

\bibitem{penn2008darwin}
Derek~C Penn, Keith~J Holyoak, and Daniel~J Povinelli.
\newblock Darwin's mistake: Explaining the discontinuity between human and
  nonhuman minds.
\newblock {\em Behavioral and brain sciences}, 31(2):109--130, 2008.

\bibitem{harris2020array}
Charles~R. Harris, K.~Jarrod Millman, St{\'{e}}fan~J. van~der Walt, Ralf
  Gommers, Pauli Virtanen, David Cournapeau, Eric Wieser, Julian Taylor,
  Sebastian Berg, Nathaniel~J. Smith, Robert Kern, Matti Picus, Stephan Hoyer,
  Marten~H. van Kerkwijk, Matthew Brett, Allan Haldane, Jaime~Fern{\'{a}}ndez
  del R{\'{i}}o, Mark Wiebe, Pearu Peterson, Pierre G{\'{e}}rard-Marchant,
  Kevin Sheppard, Tyler Reddy, Warren Weckesser, Hameer Abbasi, Christoph
  Gohlke, and Travis~E. Oliphant.
\newblock Array programming with {NumPy}.
\newblock {\em Nature}, 585(7825):357--362, September 2020.

\bibitem{2020SciPy-NMeth}
Pauli Virtanen, Ralf Gommers, Travis~E. Oliphant, Matt Haberland, Tyler Reddy,
  David Cournapeau, Evgeni Burovski, Pearu Peterson, Warren Weckesser, Jonathan
  Bright, St{\'e}fan~J. {van der Walt}, Matthew Brett, Joshua Wilson, K.~Jarrod
  Millman, Nikolay Mayorov, Andrew R.~J. Nelson, Eric Jones, Robert Kern, Eric
  Larson, C~J Carey, {\.I}lhan Polat, Yu~Feng, Eric~W. Moore, Jake
  {VanderPlas}, Denis Laxalde, Josef Perktold, Robert Cimrman, Ian Henriksen,
  E.~A. Quintero, Charles~R. Harris, Anne~M. Archibald, Ant{\^o}nio~H. Ribeiro,
  Fabian Pedregosa, Paul {van Mulbregt}, and {SciPy 1.0 Contributors}.
\newblock {{SciPy} 1.0: Fundamental Algorithms for Scientific Computing in
  Python}.
\newblock {\em Nature Methods}, 17:261--272, 2020.

\bibitem{seabold2010statsmodels}
Skipper Seabold and Josef Perktold.
\newblock statsmodels: Econometric and statistical modeling with python.
\newblock In {\em 9th Python in Science Conference}, 2010.

\bibitem{Hunter:2007}
J.~D. Hunter.
\newblock Matplotlib: A 2d graphics environment.
\newblock {\em Computing in Science \& Engineering}, 9(3):90--95, 2007.

\bibitem{reback2020pandas}
The pandas~development team.
\newblock pandas-dev/pandas: Pandas, February 2020.

\bibitem{rcite}
{R Core Team}.
\newblock {\em R: A Language and Environment for Statistical Computing}.
\newblock R Foundation for Statistical Computing, Vienna, Austria, 2021.

\bibitem{de2015jspsych}
Joshua~R De~Leeuw.
\newblock jspsych: A javascript library for creating behavioral experiments in
  a web browser.
\newblock {\em Behavior research methods}, 47(1):1--12, 2015.

\bibitem{kojima2022large}
Takeshi Kojima, Shixiang~Shane Gu, Machel Reid, Yutaka Matsuo, and Yusuke
  Iwasawa.
\newblock Large language models are zero-shot reasoners.
\newblock {\em arXiv preprint arXiv:2205.11916}, 2022.

\end{thebibliography}

\pagebreak
\begin{center}
\textbf{\LARGE Supplementary Results}
\end{center}
\setcounter{figure}{0}
\setcounter{table}{0}
\setcounter{page}{1}
\setcounter{section}{0}
\makeatletter
\renewcommand{\figurename}{Supplementary Figure}
\renewcommand{\thesection}{\large S\arabic{section}} 
\renewcommand{\thepage}{S\arabic{page}}

\captionsetup{labelfont=bf,font=small}
\begin{figure}[h!]
\centering
\begin{subfigure}[t]{1\linewidth}\vskip 0pt
    \centering
    \includegraphics[width=\linewidth]{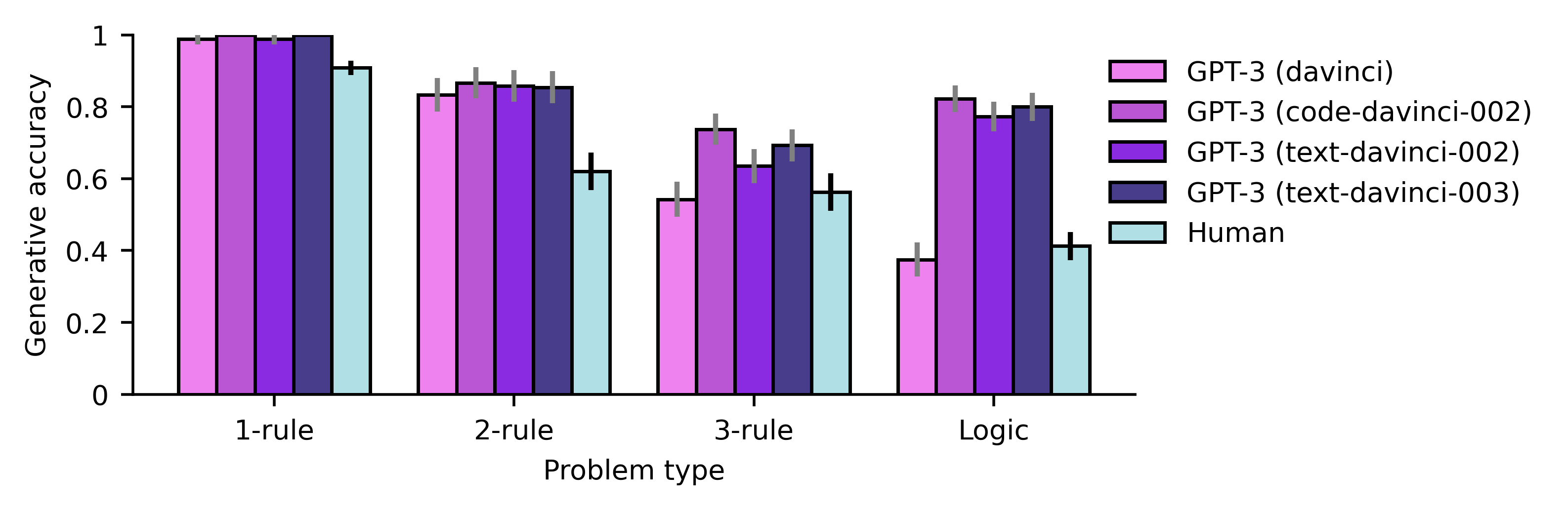} 
    \subcaption{}
\end{subfigure}
\caption{\textbf{Matrix reasoning results for all GPT-3 variants.} Zero-shot results on Digit Matrices for four GPT-3 model variants: davinci, code-davinci-002, text-davinci-002, and text-davinci-003. Results reflect generative accuracy for major problem types, including transformation problems with between one and three rules, and logic problems. Human results reflect average performance for N=40 participants. Black error bars represent standard error of the mean across participants. Summary of human results is plotted here for comparison with GPT-3, individual participant data are shown in Main Text Figure 3a. Gray error bars represent 95\% binomial confidence intervals for average performance across multiple problems.} 
\label{digit_mat_all_GPT}
\end{figure}

\captionsetup{labelfont=bf,font=small}
\begin{figure}[h!]
\centering
\begin{subfigure}[t]{1\linewidth}\vskip 0pt
    \centering
    \includegraphics[width=\linewidth]{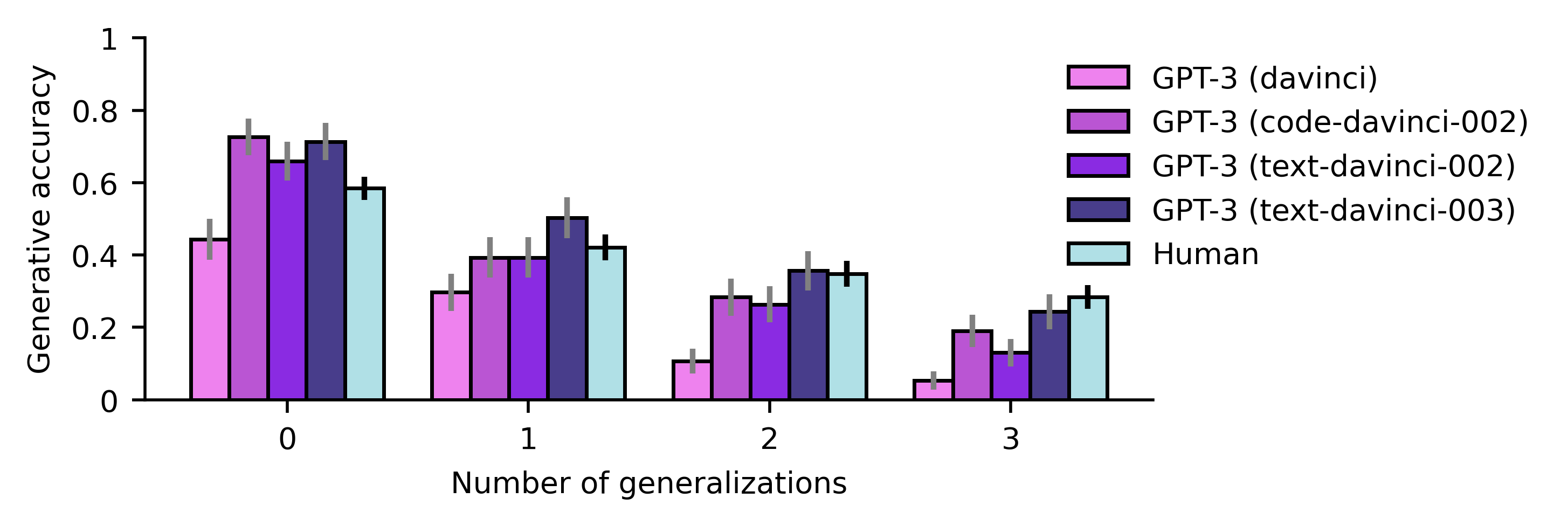} 
    \subcaption{}
\end{subfigure}
\caption{\textbf{Letter string analogy results for all GPT-3 variants.} Letter string analogy results for four GPT-3 model variants: davinci, code-davinci-002, text-davinci-002, and text-davinci-003. Results reflect generative accuracy as a function of the number of generalizations between source and target. Human results reflect average performance for N=57 participants. Black error bars represent standard error of the mean across participants. Summary of human results is plotted here for comparison with GPT-3, individual participant data are shown in Main Text Figure 6a. Gray error bars represent 95\% binomial confidence intervals for average performance across multiple problems.} 
\label{letterstring_all_GPT}
\end{figure}

\captionsetup{labelfont=bf,font=small}
\begin{figure}[h!]
\centering
\begin{subfigure}[t]{1\linewidth}\vskip 0pt
    \centering
    \includegraphics[width=\linewidth]{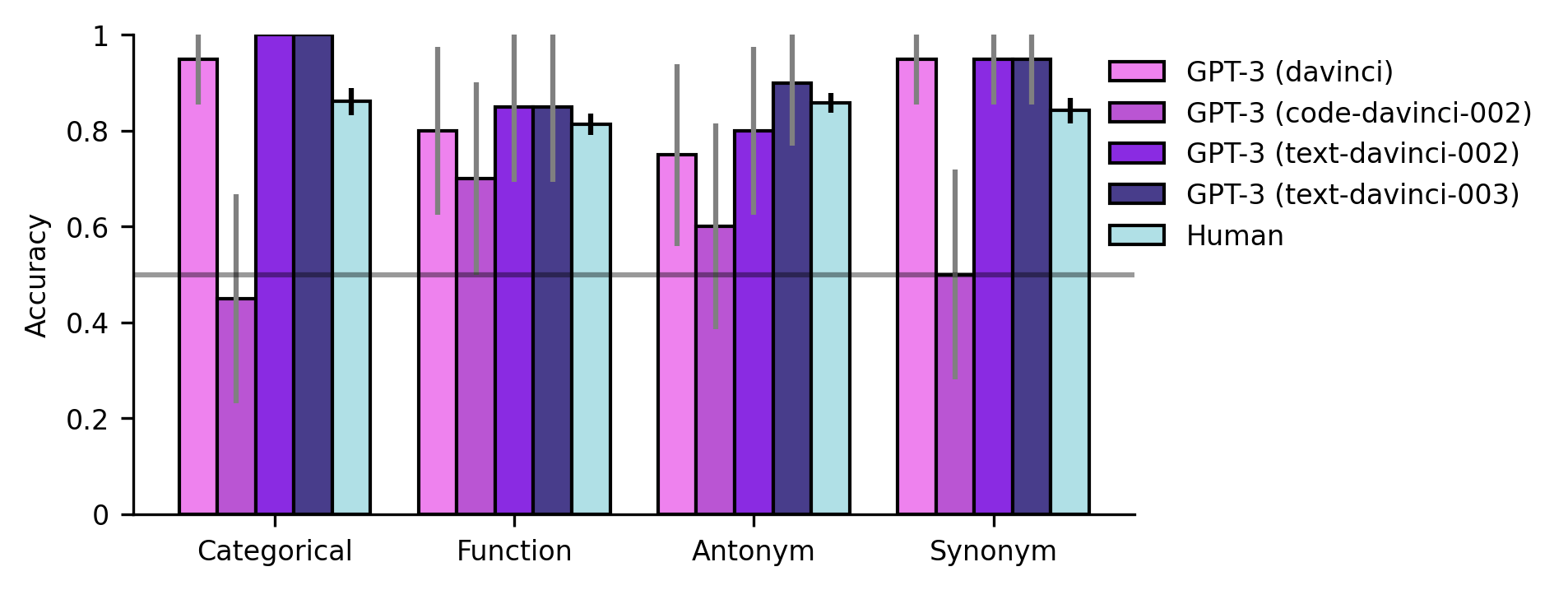} 
    \subcaption{}
\end{subfigure}
\caption{\textbf{Four-term verbal analogy results for all GPT-3 variants.} Results on UCLA Verbal Analogy Test (VAT) for four GPT-3 model variants: davinci, code-davinci-002, text-davinci-002, and text-davinci-003. Results reflect multiple-choice accuracy for problems involving different relation categories. Gray horizontal line represents chance performance. Human results reflect average performance for N=57 participants. Black error bars represent standard error of the mean across participants. Summary of human results is plotted here for comparison with GPT-3, individual participant data are shown in Main Text Figure 7a. Gray error bars represent 95\% binomial confidence intervals for average performance across multiple problems.} 
\label{UCLA_VAT_all_GPT}
\end{figure}

\captionsetup{labelfont=bf,font=small}
\begin{figure}[h!]
\centering
\begin{subfigure}[t]{0.02\textwidth}
    \textbf{\fontfamily{phv}\selectfont{a}}
\end{subfigure}
\begin{subfigure}[t]{.5\linewidth}\vskip 0pt
    \includegraphics[width=\linewidth]{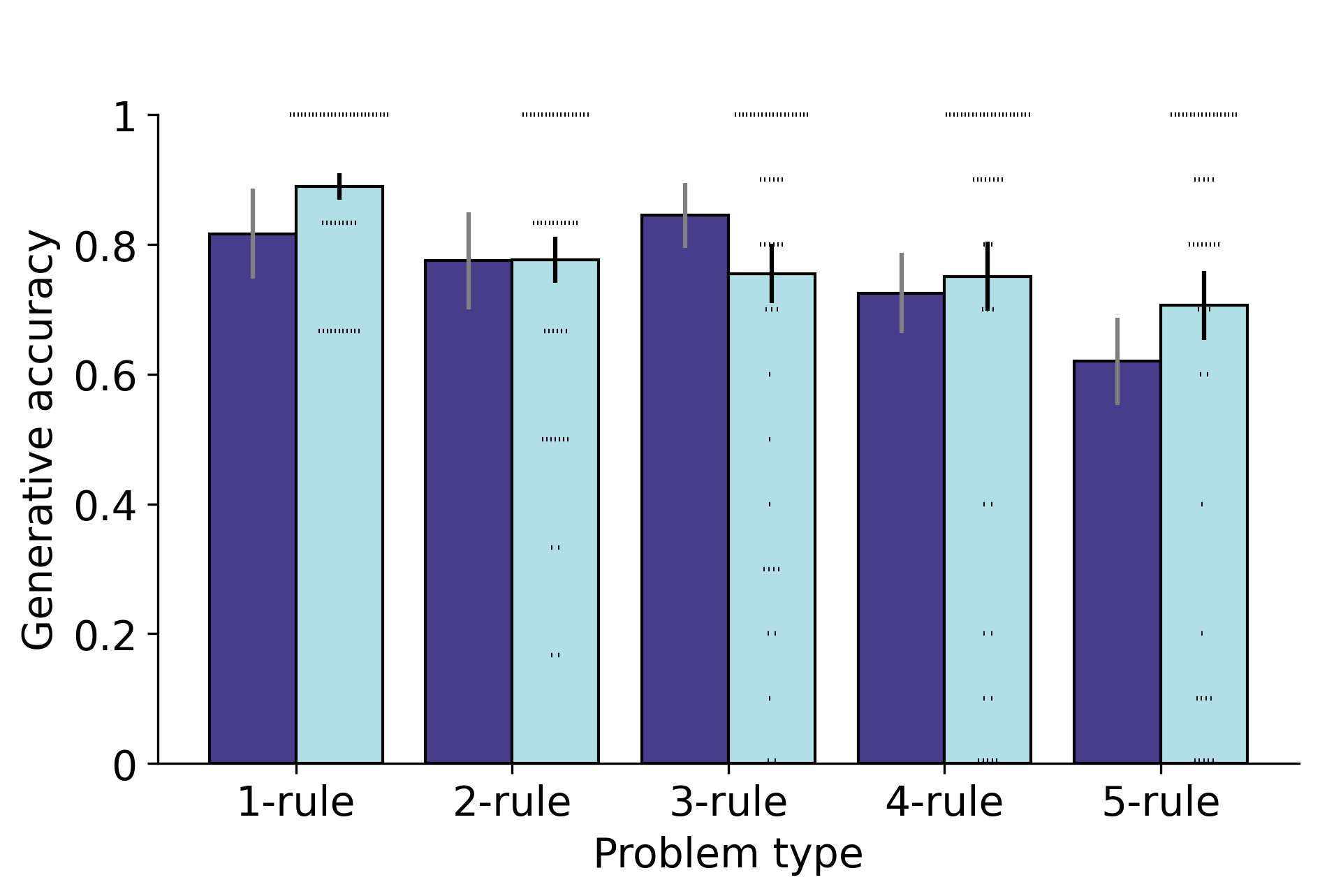} 
    \subcaption{}
    \label{prog_5rule_gen}
\end{subfigure}
\begin{subfigure}[t]{0.02\textwidth}
    \textbf{\fontfamily{phv}\selectfont{b}}
\end{subfigure}
\begin{subfigure}[t]{.413\linewidth}\vskip 0pt
    \includegraphics[width=\linewidth]{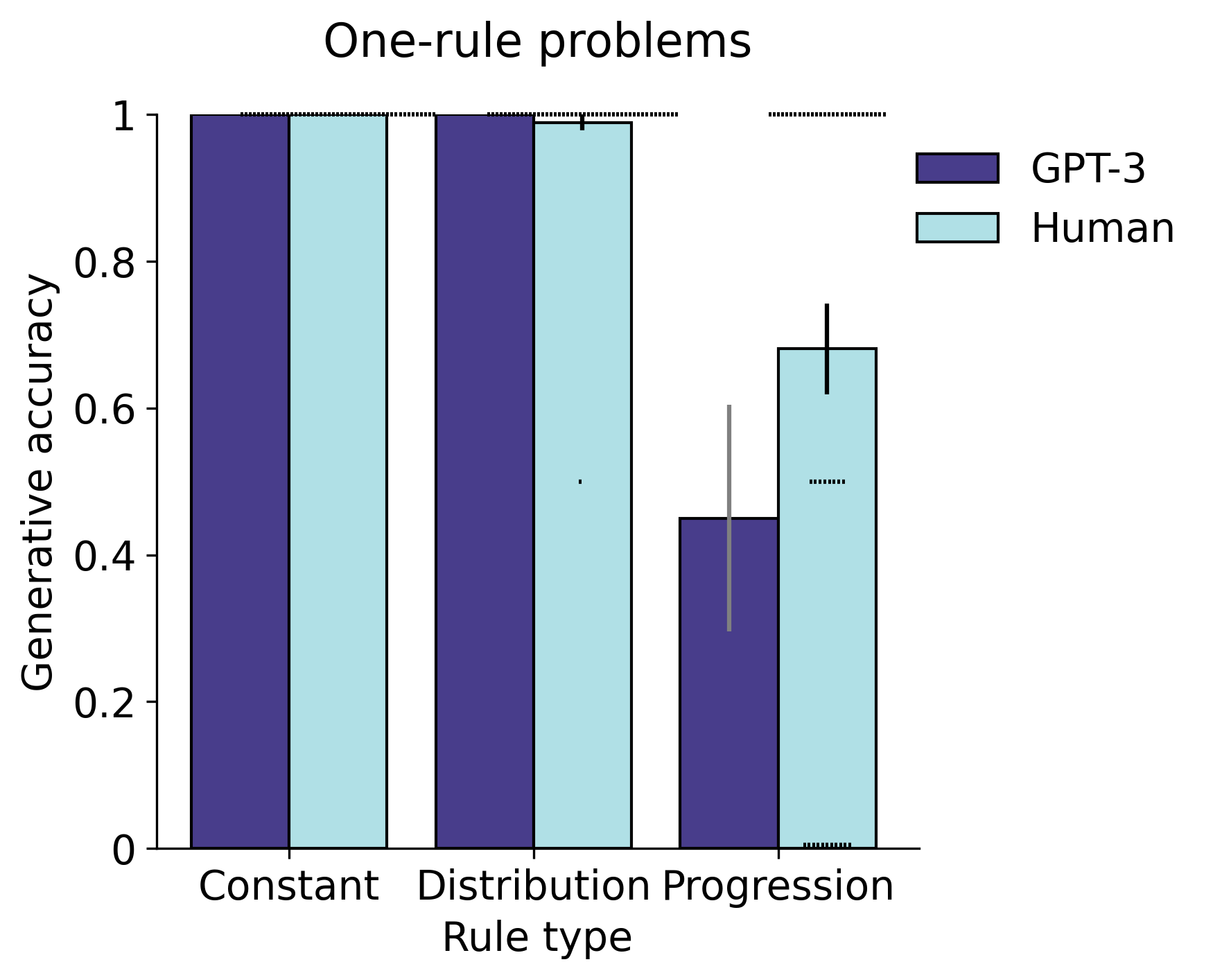} 
    \subcaption{}
    \label{prog_5rule_1rule_gen}
\end{subfigure}
\caption{\textbf{GPT-3 shows human-like contextual effects.} In a separate experiment, we presented both GPT-3 and human participants (N=47, UCLA undergraduates) with Digit Matrix problems in order of increasing complexity (easy-to-hard: one-rule problems, followed by two-rule problems, and so on). \textbf{(a)} Both GPT-3 and human participants were able to generalize the structure inferred from few-rule problems to more complex many-rule problems, resulting in very little decrease in performance even for five-rule problems (compare with the decrease in performance from one- to three-rule problems seen in Main Text Figure 3). \textbf{(b)} One-rule problems were also presented in order of increasing complexity, beginning with constant problems, followed by distribution-of-3 problems, followed by progression problems. Interestingly, this actually impaired performance on progression problems relative to zero-shot (or shuffled) presentation. This was likely due to a tendency to mistake the progression rule for the distribution-of-3 rule in the previously presented problems (which only differ in terms of a single digit). Both GPT-3 and human participants showed this effect. Human results reflect average performance for N=47 participants. Black error bars represent standard error of the mean across participants. Each dot represents accuracy for a single participant. Gray error bars represent 95\% binomial confidence intervals for average performance across multiple problems.} 
\label{prog_5rule}
\end{figure}

\captionsetup{labelfont=bf,font=small}
\begin{figure}[h!]
\centering
\begin{subfigure}[t]{0.02\textwidth}
    \textbf{\fontfamily{phv}\selectfont{a}}
\end{subfigure}
\begin{subfigure}[t]{.4\linewidth}\vskip 0pt
    \includegraphics[width=\linewidth]{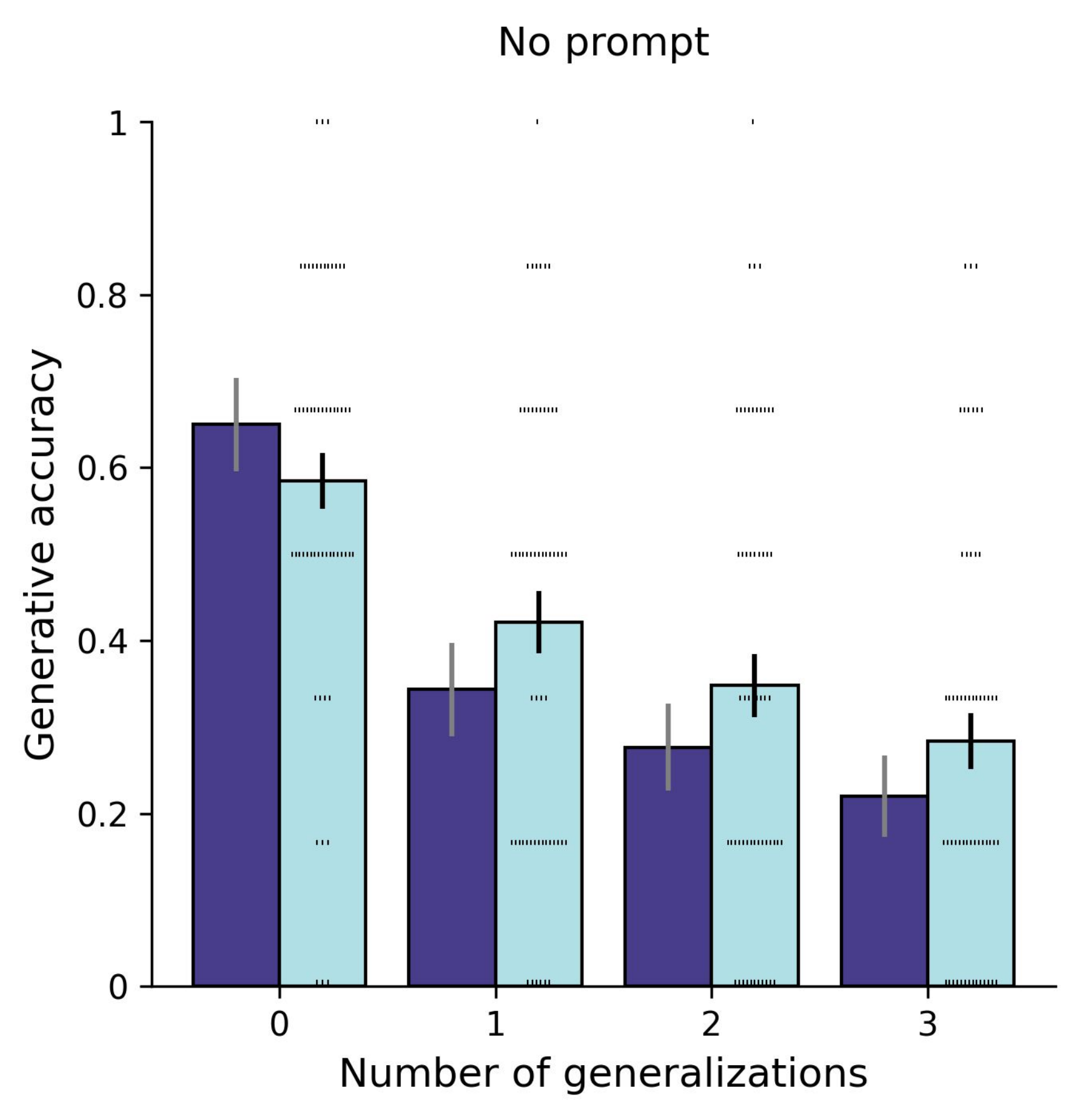} 
    \subcaption{}
    \label{letterstring_all_gen_noprompt}
\end{subfigure}
\begin{subfigure}[t]{0.02\textwidth}
    \textbf{\fontfamily{phv}\selectfont{b}}
\end{subfigure}
\begin{subfigure}[t]{.5\linewidth}\vskip 0pt
    \includegraphics[width=\linewidth]{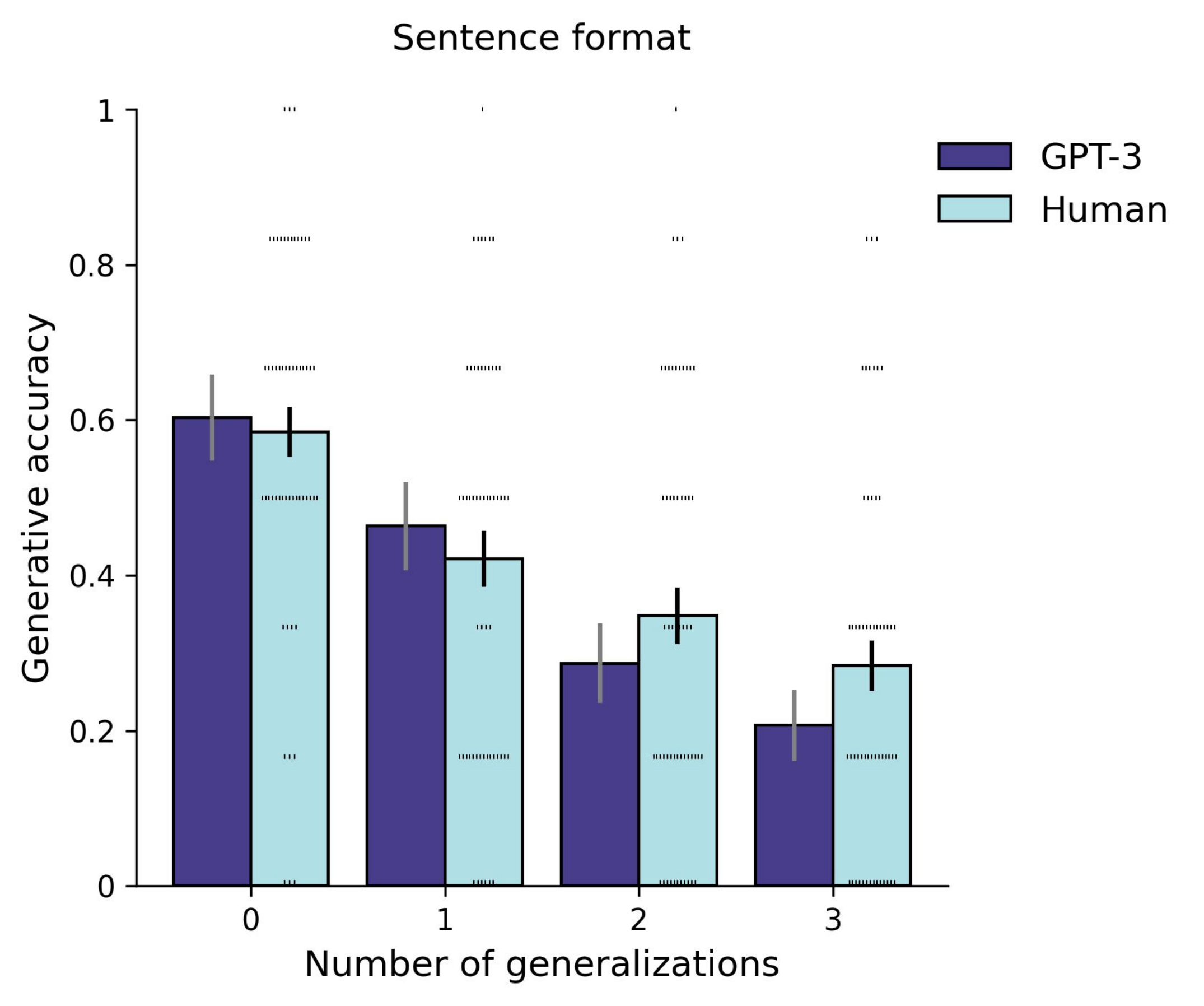} 
    \subcaption{}
    \label{letterstring_all_gen_sentence}
\end{subfigure}
\caption{\textbf{Letter string analogy results with alternative formatting.} \textbf{(a)} GPT-3 performed worse when letter string problems were presented without prompt, in comparison with results (shown in Main Text Figure 6a) when same problems were formatted in the standard way (using brackets to demarcate the analogs, and with the prompt “Let’s try to complete the pattern”). \textbf{(b)} GPT-3 also performed worse when problems were presented in the form of a sentence (e.g., `If a b c d changes to a b c e, then i j k l should change to'). Human results reflect average performance for N=57 participants (UCLA undergraduates). Black error bars represent standard error of the mean across participants. Each dot represents accuracy for a single participant. Gray error bars represent 95\% binomial confidence intervals for average performance across multiple problems.} 
\label{letterstring_results_alt}
\end{figure}

\captionsetup{labelfont=bf,font=small}
\begin{figure}[h!]
\centering
\begin{subfigure}[t]{.4\linewidth}\vskip 0pt
    \includegraphics[width=\linewidth]{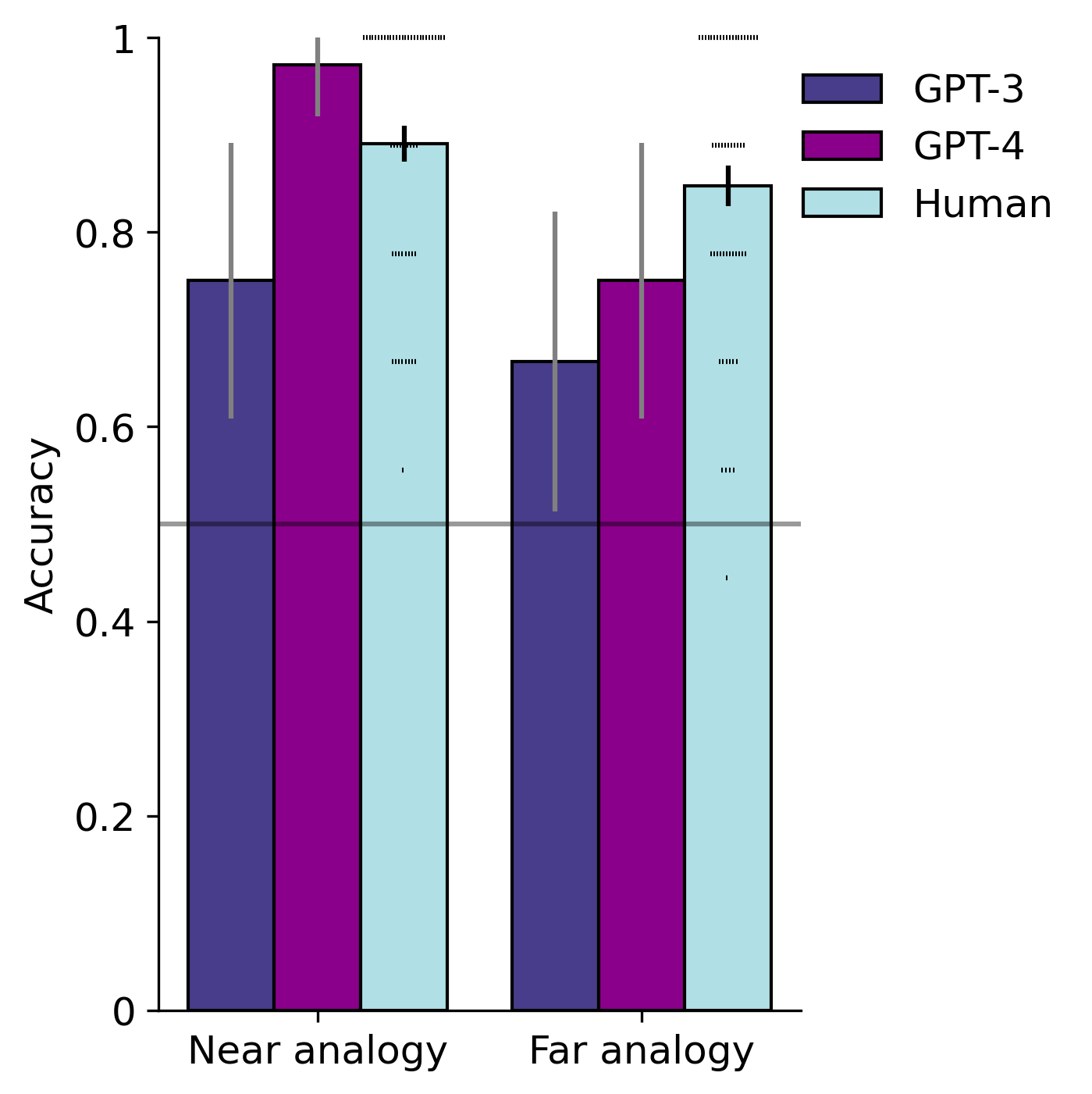} 
    \subcaption{}
\end{subfigure}
\caption{\textbf{Story analogy results including test of GPT-4.} Results for identification of analogies between stories, including both GPT-3 and GPT-4. Whereas GPT-3 was only reliably able to perform the task in the near analogy condition, GPT-4 displayed much more robust performance. GPT-4 showed nearly perfect performance in the near analogy condition, and displayed some sensitivity to higher-order relations in the far analogy condition (two-sided binomial test: $p = 0.0039$; null hypothesis is chance-level performance of 0.5). Human results reflect average performance for N=54 participants (UCLA undergraduates). Black error bars represent standard error of the mean across participants. Each dot represents accuracy for a single participant. Gray error bars represent 95\% binomial confidence intervals for average performance across multiple problems. Gray horizontal line represents chance performance.} 
\label{story_analogy_results_gpt4}
\end{figure}

\pagebreak

\section{Solutions to example matrix reasoning problems}
\label{example_mat_solutions}

The solution to the example visual matrix reasoning problem in Main Text Figure 2a is option 5. The problem is defined by a constant rule (applied to the number of shapes in each cell), and two distribution-of-3 rules (one applied to color, and one applied to shape). The solution to the example Digit Matrix problem in Main Text Figure 2b is option 7. This problem is also defined by a constant rule (applied to the digits in the center of each cell), and two distribution-of-3 rules (one applied to the digits on the left side of each cell, and one applied to the digits on the right side). 

\section{GPT-3 model variants}
\label{GPT3_variant_discussion}

Since the initial release of GPT-3~\cite{brown2020language}, OpenAI has released a number of updates to the original base model. The largest version (175B parameters) of the base model, davinci, was trained exclusively on next-token prediction using a web-based corpus of text data. Code-davinci-002 was further trained on next-token prediction using a dataset of publicly available code from GitHub~\cite{chen2021evaluating}. Text-davinci-002 and text-davinci-003 were both fine-tuned to respond appropriately to prompts~\cite{ouyang2022training}. Text-davinci-002 was initialized with code-davinci-002, and then fine-tuned using supervised learning based on a set of example prompts and responses. Text-davinci-003 was initialized with text-davinci-002, and then further fine-tuned using reinforcement learning from human feedback (RLHF). In RLHF, a reward model (a separate neural network) is first trained to predict human ratings for pairs of human-generated prompts and language-model responses, and this reward model is then used to fine-tune the primary language model through reinforcement learning. More details on the different model variants and training objectives can be found at \href{https://platform.openai.com/docs/model-index-for-researchers}{https://platform.openai.com/docs/model-index-for-researchers}.

We evaluated all four of these variants on Digit Matrices (Figure~\ref{digit_mat_all_GPT}), letter string analogies (Figure~\ref{letterstring_all_GPT}), and four-term verbal analogies (Figure~\ref{UCLA_VAT_all_GPT}). Text-davinci-003 displayed the best overall performance, but other model variants performed well on a subset of tasks. For instance, code-davinci-002 performed well on the Digit Matrices and letter string problems. These task domains both involve simple alphanumeric characters and highly regular relational structure, similar to computer code. It therefore seems likely that code-davinci-002’s strong performance on these tasks was a consequence of having been trained on code. By contrast, code-davinci-002 performed very poorly on four-term verbal analogies (near chance performance), whereas the original davinci model performed relatively well on these problems. This suggests that code-davinci-002's ability to model synthetic code-like structures may have come at the cost of the ability to process more real-world relational concepts. Text-davinci-002, and especially text-davinci-003, seem to have combined both of these abilities, perhaps as a result of prompt training, though these models may have also received additional fine-tuning on the original language modeling task~\cite{ouyang2022training}. Finally, it seems likely that prompt training improved text-davinci-002 and text-davinci-003’s ability to perform tasks without the need for few-shot task demonstrations, therefore making it easier to evaluate these models’ latent capabilities in a zero-shot setting.

We also performed an initial investigation of GPT-4 on the story analogy problems from Gentner et al.~\cite{gentner1993roles} (Supplementary Figure~\ref{story_analogy_results_gpt4}). GPT-4 showed significant improvement on this task relative to GPT-3, more reliably identifying the target story that shared higher-order relations with the source, and providing more precise explanations (Section~\ref{gpt4_story_example}). We were not able to test GPT-4 on the other analogy tasks due to a lack of API access. Very little is known about the details of GPT-4, but it is likely that this improvement stems at least in part from increased scale of both model and training set~\cite{gpt4_tech_report}.

\section{Presence of test materials in GPT-3's training data}
\label{test_materials_in_training_data}

Given the massive and uncurated nature of GPT-3's training data, it is important to consider the likelihood that our test materials were included in this training data, and the possibility therefore that GPT-3 may have memorized some of these materials (thus undermining their use as a test of zero-shot reasoning).

The Digit Matrices dataset was created specifically for the purposes of our study and therefore certainly was not included in GPT-3's training data. Furthermore, this problem format itself is, to the best of our knowledge, completely novel, and it is thus extremely unlikely that GPT-3 has been trained on similar problems.

The letter string problem set that we used was also created specifically for the purposes of our study, and so could not have been included in GPT-3's training data. It is possible that GPT-3 has been trained on other letter string analogy problems, as these problems are discussed on a number of webpages (e.g., \href{https://cogsci.indiana.edu/lap.html}{https://cogsci.indiana.edu/lap.html}). However, we were not able to obtain any evidence that GPT-3 knows about these problems, e.g., by asking it to describe or give examples of such problems.

The UCLA Verbal Analogy Test~\cite{lu2019emergence}, and the four-term verbal analogy sets from Sternberg and Nigro~\cite{sternberg1980developmental} and Jones et al.~\cite{jones2022differential} are all available in a downloadable supplement in a recently published paper~\cite{ichien2020verbal}, though to our knowledge they are not published directly in the form of webpages, and are thus unlikely to be included in web crawl data such as the Common Crawl dataset on which GPT-3 was trained. The dataset of SAT problems from Turney et al.~\cite{turney2003combining} is not publicly available, but has been distributed to a relatively small number of researchers. It is possible that any of these datasets were deliberately included in GPT-3's training data (i.e., as a supplement to the web crawl data). However, we were not able to find any evidence that GPT-3 had memorized any of these problems, e.g., by prompting GPT-3 to complete these problems based only on the source analog (the A and B terms alone).

The story analogy materials from Gentner et al.~\cite{gentner1993roles} were also included in the downloadable supplement associated with~\cite{ichien2020verbal}, though to our knowledge have not been published directly to any webpages. We were also not able to obtain any evidence that GPT-3 had memorized any of these stories, e.g., by asking it to complete partial stories.

The analogy between the general story and the radiation problem is discussed in the original publication from Gick and Holyoak~\cite{gick1980analogical}, and is discussed on other webpages as well (e.g., \href{http://cognitivepsychology.wikidot.com/problem-solving:analogy}{http://cognitivepsychology.wikidot.com/problem-solving:analogy}). We found evidence that GPT-3 is familiar with this study (it was able to identify the year of publication, and journal in which it was published), but we did not find any evidence that GPT-3 had memorized the details of the analogy itself. GPT-3 was also able to solve the radiation problem via an analogy with lesser known source stories~\cite{holyoak1987surface}. It seems unlikely that GPT-3's success on these problems is due to prior exposure, but these results should be interpreted with caution. The other materials used to evaluate analogical problem solving are also discussed in the associated publication~\cite{holyoak1984development}, but we did not find any evidence that GPT-3 is familiar with this study, or that it had memorized these materials.

\section{Story analogy results}
\label{story_analogy_examples}

The following section contains examples of GPT-3's responses to the story analogy problems. GPT-3's responses are presented in bold text.

\subsection{Near analogy}

The following contains an example from the near analogy condition. GPT-3 identified the correct answer (Story A).

\begin{quote}

\noindent Consider the following story: 

\vspace{3mm}

\noindent Story 1: Karla, an old hawk, lived at the top of a tall oak tree. One afternoon, she saw a hunter on the ground with a bow and some crude arrows that had no feathers. The hunter took aim and shot at the hawk but missed. Karla knew the hunter wanted her feathers so she glided down to the hunter and offered to give him a few. The hunter was so grateful that he pledged never to shoot at a hawk again. He went off and shot deer instead. 

\vspace{3mm}

\noindent Now consider two more stories: 

\vspace{3mm}

\noindent Story A: Once there was an eagle named Zerdia who nested on a rocky cliff. One day she saw a sportsman coming with a crossbow and some bolts that had no feathers. The sportsman attacked but the bolts missed. Zerdia realized that the sportsman wanted her tailfeathers so she flew down and donated a few of her tailfeathers to the sportsman. The sportsman was pleased. He promised never to attack eagles again. 

\vspace{3mm}

\noindent Story B: Once there was an eagle named Zerdia who donated a few of her tailfeathers to a sportsman so he would promise never to attack eagles. One day Zerdia was nesting high on a rocky cliff when she saw the sportsman coming with a crossbow. Zerdia flew down to meet the man, but he attacked and felled her with a single bolt. As she fluttered to the ground Zerdia realized that the bolt had her own tailfeathers on it.

\vspace{3mm}

\noindent Which of Story A and Story B is a better analogy to Story 1? 
\noindent Is the best answer Story A, Story B, or both are equally analogous?

\vspace{3mm}

\noindent \textbf{Story A is the better analogy to Story 1. Story B is not as analogous because it has a different ending. In Story 1, the hunter is grateful for the hawk's feathers and promises not to shoot at hawks again. In Story B, the hunter attacks the eagle and kills her with her own feathers.}

\end{quote}

\subsection{Far analogy}

The following contains an example from the far analogy condition. GPT-3 identified the correct answer (Story A).

\begin{quote}

\noindent Consider the following story: 

\vspace{3mm}

\noindent Story 1: Karla, an old hawk, lived at the top of a tall oak tree. One afternoon, she saw a hunter on the ground with a bow and some crude arrows that had no feathers. The hunter took aim and shot at the hawk but missed. Karla knew the hunter wanted her feathers so she glided down to the hunter and offered to give him a few. The hunter was so grateful that he pledged never to shoot at a hawk again. He went off and shot deer instead. 

\vspace{3mm}

\noindent Now consider two more stories: 

\vspace{3mm}

\noindent Story A: Once there was a small country called Zerdia that learned to make the world's smartest computer. One day Zerdia was attacked by its warlike neighbor, Gagrach. But the missiles were badly aimed and the attack failed. The Zerdian government realized that Gagrach wanted Zerdian computers so it offered to sell some of its computers to the country. The government of Gagrach was very pleased. It promised never to attack Zerdia again. 

\vspace{3mm}

\noindent Story B: Once there was a small country called Zerdia that learned to make the world's smartest computer. Zerdia sold one of itssupercomputers to its neighbor, Gagrach, so Gagrach would promise never to attack Zerdia. But one day Zerdia was overwhelmed by a surprise attack from Gagrach. As it capitulated the crippled government of Zerdia realized that the attacker's missiles had been guided by Zerdian supercomputers. 

\vspace{3mm}

\noindent Which of Story A and Story B is a better analogy to Story 1? 
\noindent Is the best answer Story A, Story B, or both are equally analogous?

\vspace{3mm}

\noindent \textbf{The best answer is Story A. Story A is a better analogy to Story 1 because it follows a similar pattern of a protagonist offering something of value to an antagonist in exchange for a promise of peace. In Story 1, Karla offers her feathers to the hunter in exchange for a promise not to shoot at hawks again. In Story A, Zerdia offers its computers to Gagrach in exchange for a promise not to attack Zerdia again.}

\end{quote}

\subsection{GPT-4}
\label{gpt4_story_example}

The following contains an example response from GPT-4 in the far analogy condition. GPT-4 both identifies the correct answer (Story A), and offers a precise explanation of the analogy.

\begin{quote}

\noindent Consider the following story: 

\vspace{3mm}

\noindent Story 1: William was a patient in a psychiatric hospital who was confined indoors almost all the time. He could never pass the monthly room inspections so he hated them. He spent most of his time daydreaming about food. A few days before the April inspection William's room was still a mess since he had done nothing but daydream. To provide William with an incentive, the nurse promised him some gingerbread from the cookie shop if he scrubbed his room and put it in order once and for all. William was overjoyed. But there was no longer enough time for him to put it in order. As a result, he did not pass the inspection and did not get any gingerbread. William sulked all day and slammed his door so hard the plaster cracked, but he still didn't get any gingerbread.

\vspace{3mm}

\noindent Now consider two more stories: 

\vspace{3mm}

\noindent Story A: Karen always did poorly in high school so she despised it. But she loved vacations. She spent most of her time dreaming about going to Hawaii. Not long before the end of her fourth year Karen was not doing at all well in her classes because she had spent all her time daydreaming. To motivate her, Karen's father promised her that if she did well enough during the next few weeks to graduate from high school he would pay for her trip to Hawaii. This made Karen extremely happy. But she was too far behind in her classes. Consequently she failed too many and did not go to Hawaii.

\vspace{3mm}

\noindent Story B: Karen disliked high school so she always had trouble getting passing grades in her classes. Vacations were her favorite part of the school year. Not long before the end of her fourth year of school, Karen was not doing well at all because she despised school. To motivate her, Karen's father promised her a trip to Hawaii if she would just pass enough of her classes to graduate. This made Karen extremely happy. During the last few weeks of school she spent most of her time dreaming about Hawaii and preparing for her trip there. Consequently, she failed to graduate and she did not get to go.

\vspace{3mm}

\noindent Which of Story A and Story B is a better analogy to Story 1? 
\noindent Is the best answer Story A, Story B, or both are equally analogous?

\vspace{3mm}

\noindent \textbf{The best answer is Story A. Both Story A and Story 1 share the following similarities:
\begin{enumerate}
    \item The main characters (William and Karen) are not successful in their respective situations (room inspection and graduating high school) due to spending too much time daydreaming.
    \item An incentive is provided to motivate the main characters (gingerbread for William and a trip to Hawaii for Karen).
    \item Despite being motivated by the incentive, both characters fail to achieve the desired outcome because they are too far behind in their tasks (cleaning the room and catching up on classes).
\end{enumerate}
Story B is less analogous to Story 1 because Karen spends her time daydreaming and preparing for the trip instead of focusing on her classes, which leads to her failure to graduate. The cause of her failure is different from that in Story 1, where William fails due to the lack of time to complete the task.}

\end{quote}

\section{Analogical problem-solving results}
\label{problem_solving_results}

The following sections contain all of the materials used to evaluate analogical reasoning in natural language problem-solving. GPT-3's responses are presented in bold text.

\subsection{Solution to radiation problem in isolation}
\label{radiation_prob_only}

The following contains GPT-3's proposed solution to the radiation problem when presented in isolation.

\begin{quote}

\noindent Target problem: 

\vspace{3mm}

\noindent Suppose you were a doctor faced with a patient suffering from a malignant stomach tumor. The tumor is inoperable, and the patient will die unless it is destroyed. There is a kind of ray that will destroy the tumor if the rays reach it at a high intensity. But unfortunately, such high-intensity rays will also destroy the healthy tissue they pass through on the way to the tumor. At lower intensities the rays will not damage the healthy tissue, but neither will they remove the tumor. How can the doctor use rays to destroy the tumor, while at the same time sparing the healthy tissue? 

\vspace{3mm}

\noindent Solution:

\vspace{3mm}

\noindent \textbf{The doctor can use a technique called brachytherapy, which involves placing a radioactive source directly into or near the tumor. This allows the doctor to deliver a high dose of radiation directly to the tumor, while sparing the healthy tissue around it. The radiation source can be placed in the tumor using a catheter, or it can be placed in a sealed container and placed near the tumor. This technique is often used to treat tumors that are inoperable or too close to vital organs to be safely removed.}

\end{quote}

\subsection{Solution to radiation problem following general story}
\label{radiation_prob_general_story}

The following contains GPT-3's proposed solution to the radiation problem when first presented with the general story, along with responses to followup questions that explicitly probe the analogy between the source story and the target problem.

\begin{quote}

\noindent Source story:

\vspace{3mm}

\noindent An evil dictator controlled a fortress situated in the center of a small country. Many roads radiated out from the fortress like spokes on a wheel. A general raised an army at the border, vowing to capture the fortress and overthrow the dictator. The general was about to send his entire army down one road to capture the fortress, when he learned that the dictator had mined each road so that although small groups could still pass, a large army would set off an explosion. The general then had a clever idea: he divided his army into small groups and dispatched each group to the head of a different road. Upon his signal, each group charged down a different road. All the groups passed safely to the fortress, where the entire army attacked the fortress in full strength. In this way the general captured the fortress and overthrew the dictator.

\vspace{3mm}

\noindent Target problem: 

\vspace{3mm}

\noindent Suppose you were a doctor faced with a patient suffering from a malignant stomach tumor. The tumor is inoperable, and the patient will die unless it is destroyed. There is a kind of ray that will destroy the tumor if the rays reach it at a high intensity. But unfortunately, such high-intensity rays will also destroy the healthy tissue they pass through on the way to the tumor. At lower intensities the rays will not damage the healthy tissue, but neither will they remove the tumor. How can the doctor use rays to destroy the tumor, while at the same time sparing the healthy tissue?

\vspace{3mm}

\noindent Solution:

\vspace{3mm}

\noindent \textbf{The doctor could divide the rays into small groups and send each group down a different path to the tumor. By using a lower intensity for each group, the rays will not damage the healthy tissue, but when they all reach the tumor at the same time, the combined intensity will be enough to destroy it. The doctor could also use imaging technology to map out the paths of the rays and ensure that they all reach the tumor at the same time.}

\vspace{3mm}

\noindent Can you identify an analogy between the source story and the target problem? 

\vspace{3mm}

\noindent \textbf{The analogy between the source story and the target problem is that both involve a situation in which a large force needs to be divided into smaller groups in order to achieve a desired outcome. In the source story, the general divides his army into small groups in order to avoid setting off the mines and capture the fortress. In the target problem, the doctor needs to divide the rays into smaller groups in order to destroy the tumor without damaging the healthy tissue.}

\vspace{3mm}

\noindent Who or what in the target problem is most analogous to the general in the source story? 

\vspace{3mm}

\noindent \textbf{The doctor in the target problem is most analogous to the general in the source story. Both are faced with a difficult situation and must come up with a clever solution in order to achieve their desired outcome. In the source story, the general divides his army into small groups in order to avoid setting off the mines and capture the fortress. In the target problem, the doctor needs to divide the rays into smaller groups in order to destroy the tumor without damaging the healthy tissue.}

\vspace{3mm}

\noindent Who or what in the target problem is most analogous to the evil dictator in the source story? 

\vspace{3mm}

\noindent \textbf{The malignant stomach tumor in the target problem is most analogous to the evil dictator in the source story. Both are obstacles that must be overcome in order to achieve the desired outcome. In the source story, the evil dictator must be overthrown in order for the general to capture the fortress. In the target problem, the malignant stomach tumor must be destroyed in order for the patient to be saved.}

\vspace{3mm}

\noindent Who or what in the target problem is most analogous to the army in the source story?

\vspace{3mm}

\noindent \textbf{The rays in the target problem are most analogous to the army in the source story. Both are forces that must be divided into smaller groups in order to achieve the desired outcome. In the source story, the army is divided into small groups in order to avoid setting off the mines and capture the fortress. In the target problem, the rays must be divided into smaller groups in order to destroy the tumor without damaging the healthy tissue.}

\end{quote}

\subsection{Solution to radiation problem following alternative source analogs}
\label{radiation_prob_other_source}

We also tested GPT-3 using alternative materials for the source analog taken from a different study~\cite{holyoak1987surface}. These stories each involve a solution based on convergence, but differ from the general story used in the original study of Gick and Holyoak~\cite{gick1980analogical}. GPT-3 was also able to identify the convergence solution to the radiation problem when presented with these alternative source analogs. The following contains GPT-3's response to the radiation problem when first presented with an analogous story about a student in a physics lab who uses lasers to repair a lightbulb.

\begin{quote}

\noindent Source story:

\vspace{3mm}

\noindent In a physics lab at a major university, a very expensive lightbulb which would emit precisely controlled quantities of light was being used in some experiments. Ruth was the research assistant responsible for operating the sensitive lightbulb. One morning she came into the lab and found to her dismay that the lightbulb no longer worked. She realized that she had forgotten to turn it off the previous night. As a result the lightbulb overheated, and the filament inside the bulb had broken into two parts. The surrounding glass bulb was completely sealed, so there was no way to open it. Ruth knew that the lightbulb could be repaired if a brief, high-intensity laser beam could be used to fuse the two parts of the filament into one. Furthermore, the lab had the necessary equipment to do the job. However, a high-intensity laser beam would also break the fragile glass surrounding the filament. At lower intensities the laser would not break the glass, but neither would it fuse the filament. So it seemed that the lightbulb could not be repaired, and a costly replacement would be required. Ruth was about to give up when she had an idea. She placed several lasers in a circle around the lightbulb, and administered low-intensity laser beams from several directions all at once. The beams all converged on the filament, where their combined effect was enough to fuse it. Since each spot on the surrounding glass received only a low-intensity beam from one laser, the glass was left intact. Ruth was greatly relieved that the lightbulb was repaired, and she then went on to successfully complete the experiment.

\vspace{3mm}

\noindent Target problem: 

\vspace{3mm}

\noindent Suppose you were a doctor faced with a patient suffering from a malignant stomach tumor. The tumor is inoperable, and the patient will die unless it is destroyed. There is a kind of ray that will destroy the tumor if the rays reach it at a high intensity. But unfortunately, such high-intensity rays will also destroy the healthy tissue they pass through on the way to the tumor. At lower intensities the rays will not damage the healthy tissue, but neither will they remove the tumor. How can the doctor use rays to destroy the tumor, while at the same time sparing the healthy tissue?

\vspace{3mm}

\noindent Solution:

\vspace{3mm}

\noindent \textbf{The doctor could use multiple low-intensity rays from several directions to converge on the tumor. The combined effect of the rays would be enough to destroy the tumor, while each spot on the healthy tissue would only receive a low-intensity beam from one ray, thus leaving it intact.}
    
\end{quote}

The following contains GPT-3's response to the radiation problem when first presented with a different story involving ultrasound instead of lasers. GPT-3 was still able to use this story to identify the convergence solution, despite the fact that it was designed to have lower surface similarity to the radiation problem.

\begin{quote}

\noindent Source story:

\vspace{3mm}

\noindent In a physics lab at a major university, a very expensive lightbulb which would emit precisely controlled quantities of light was being used in some experiments. Ruth was the research assistant responsible for operating the sensitive lightbulb. One morning she came into the lab and found to her dismay that the lightbulb no longer worked. She realized that she had forgotten to turn it off the previous night. As a result the lightbulb overheated, and the two wires in the filament inside the bulb fused together. The surrounding glass bulb was completely sealed, so there was no way to open it. Ruth knew that the lightbulb could be repaired if a brief, high-intensity ultrasound wave could be used to jar apart the fused parts. Furthermore, the lab had the necessary equipment to do the job. However, a high-intensity ultrasound wave would also break the fragile glass surrounding the filament. At lower intensities the ultrasound wave would not break the glass, but neither would it jar apart the fused parts. So it seemed that the lightbulb could not be repaired, and a costly replacement would be required. Ruth was about to give up when she had an idea. She placed several ultrasound machines in a circle around the lightbulb, and administered low-intensity ultrasound waves from several directions all at once. The waves all converged on the filament, where their combined effect was enough to jar apart the fused parts. Since each spot on the surrounding glass received only a low-intensity wave from one ultrasound machine, the glass was left intact. Ruth was greatly relieved that the lightbulb was repaired, and she then went on to successfully complete the experiment.

\vspace{3mm}

\noindent Target problem: 

\vspace{3mm}

\noindent Suppose you were a doctor faced with a patient suffering from a malignant stomach tumor. The tumor is inoperable, and the patient will die unless it is destroyed. There is a kind of ray that will destroy the tumor if the rays reach it at a high intensity. But unfortunately, such high-intensity rays will also destroy the healthy tissue they pass through on the way to the tumor. At lower intensities the rays will not damage the healthy tissue, but neither will they remove the tumor. How can the doctor use rays to destroy the tumor, while at the same time sparing the healthy tissue?

\vspace{3mm}

\noindent Solution:

\vspace{3mm}

\noindent \textbf{The doctor could use multiple low-intensity rays from several directions all at once. The rays would converge on the tumor, where their combined effect would be enough to destroy it. Since each spot on the healthy tissue would receive only a low-intensity ray from one source, the healthy tissue would be left intact.}
    
\end{quote}

\subsection{Solution to radiation problem with distracting stories}
\label{radiation_prob_distracting_stories}

The following contains GPT-3's proposed solution to the radiation problem when first presented with both the general story and two distracting (i.e., non-analogous) stories. GPT-3 did not identify the convergence solution in this case, instead proposing the same solution as it did when presented with the radiation problem in isolation.

\begin{quote}

\noindent Story \#1 -- The Wine Merchants:

\vspace{3mm}

\noindent One day a rich man found that his wine cellar was empty. So he sent out messengers to announce a generous offer. The first person to bring the rich man a barrel of wine would be given a brick of solid gold. However, the offer would expire at sundown. Two wine merchants heard the news. Each had a horse-drawn cart loaded with large barrels of wine. They both set out for the duke’s palace at once. An hour before sundown they came to a place where the bridge had been washed out by a raging river. The first merchant drove his horses and cart into the flood in a desperate attempt to reach the other side. But the horses were already exhausted and could not fight the current. The cart overturned, and the horses, wine, and driver were washed away. The second merchant tried a different tactic. He poured the wine out of all but one of his barrels, and lashed them together to form a raft; then he loaded the one full barrel, a horse, and himself on top. He set the raft adrift and floated downstream. In a few minutes the raft came to rest on the shore in front of the town where the rich man lived. The merchant disembarked, loaded the wine barrel on the horse, and led it to the rich man’s house. He arrived just as the sun was setting, and collected the gold brick as a reward for his efforts. 

\vspace{3mm}

Story \#2 -- The General:

\vspace{3mm}

\noindent An evil dictator controlled a fortress situated in the center of a small country. Many roads radiated out from the fortress like spokes on a wheel. A general raised an army at the border, vowing to capture the fortress and overthrow the dictator. The general was about to send his entire army down one road to capture the fortress, when he learned that the dictator had mined each road so that although small groups could still pass, a large army would set off an explosion. The general then had a clever idea: he divided his army into small groups and dispatched each group to the head of a different road. Upon his signal, each group charged down a different road. All the groups passed safely to the fortress, where the entire army attacked the fortress in full strength. In this way the general captured the fortress and overthrew the dictator. 

\vspace{3mm}

\noindent Story \#3 -- The Identical Twins:

\vspace{3mm}

\noindent Once there were identical twins who were continually playing pranks on their family, friends, and teachers. The annual school picnic was always a big event for the twins. There were races and other athletic events in which the twins won lots of prizes. One year a new student arrived who was a star runner. The twins wanted to win the main event: the 2-mile race through the woods behind the school. So they secretly devised a plan which would enable them to outdo the newcomer. The day of the race arrived. Each runner was to pick his own path through the woods to a clearing, where a teacher stood posted to determine the winner. One twin entered the race, while the other excused himself on the grounds that he had hurt his leg in an earlier broadjumping event. The race began and the students rushed into the woods. The twin rushed into the woods and waited until the others had passed out of sight. Then he went back to the school using a path hidden from the picnic area. Shortly after, the other twin, who had been hiding behind a rock near the finish line of the race, burst out and ran into the clearing ahead of the other runners. The teacher named him the winner and marveled at the speed of his running. Next year the twins switched places and thereafter maintained their status on this event. 

\vspace{3mm}

\noindent Target problem: 

\vspace{3mm}

\noindent Suppose you were a doctor faced with a patient suffering from a malignant stomach tumor. The tumor is inoperable, and the patient will die unless it is destroyed. There is a kind of ray that will destroy the tumor if the rays reach it at a high intensity. But unfortunately, such high-intensity rays will also destroy the healthy tissue they pass through on the way to the tumor. At lower intensities the rays will not damage the healthy tissue, but neither will they remove the tumor. How can the doctor use rays to destroy the tumor, while at the same time sparing the healthy tissue? 

\vspace{3mm}

\noindent Solution:

\vspace{3mm}

\noindent \textbf{The doctor can use a technique called brachytherapy. This involves placing a radioactive source directly into the tumor. The radiation from the source will be of a high enough intensity to destroy the tumor, but the radiation will not travel far enough to damage the healthy tissue. The doctor can also use a technique called stereotactic radiotherapy. This involves using a machine to deliver a high-intensity beam of radiation to the tumor from outside the body. The beam is focused on the tumor, so that the healthy tissue is spared.}

\end{quote}

\subsection{Solution to radiation problem with distracting stories and prompt}
\label{radiation_prob_distracting_stories_prompt}

When presented with distracting stories, GPT-3 did not propose a solution to the radiation problem based on the analogy with the general story. We therefore provided a prompt to encourage GPT-3 to consider the previously presented stories when generating its solution. This same prompt was used in the original human behavioral study~\cite{gick1980analogical}, where it was found to increase the likelihood that participants would identify the analogous convergence solution. The source stories and target problem were presented in the exact same manner, but the prompt used to elicit GPT-3's solution was changed. When presented with this prompt, GPT-3 correctly identified the convergence solution, and specifically identified the general story as the relevant source.

\begin{quote}

\noindent Solution (in solving this problem you may find that one of the stories you read before will give you a hint for a solution of this problem):

\vspace{3mm}

\noindent \textbf{The doctor could use the same tactic as the general in the second story. He could divide the rays into small groups and send each group down a different path to the tumor. The rays would be of low intensity, so they would not damage the healthy tissue, but when they all reached the tumor at the same time, the combined intensity would be enough to destroy it.}

\end{quote}

\subsection{Solution to gumball problem in isolation}
\label{gumball_prob_only}

The following contains GPT-3's proposed solution to the gumball problem (from~\cite{holyoak1984development}) when presented in isolation (enumeration is part of the response provided by GPT-3).

\begin{quote}

\noindent Target problem:

\vspace{3mm}

\noindent You are seated in front of a table. Two bowls are on the table, one within your reach and one farther away. The closer bowl contains a number of small gumballs, and the farther one is empty. Also on the table are an aluminum walking cane, a large rectangular sheet of heavy paper (posterboard), a hollow cardboard tube long enough to reach the farther bowl, scissors, string, tape, paper clips, and rubber bands. Using the materials provided, how can you transfer the balls from the filled to the empty bowl without leaving your seat? 

\vspace{3mm}

\noindent Solution:

\vspace{3mm}

\textbf{
\begin{enumerate}
    \item Cut the posterboard into a long strip and tape it to the end of the cardboard tube.
    \item Tie one end of the string to the end of the tube and the other end to the handle of the cane.
    \item Place the tube in the bowl of gumballs and use the cane to pull the tube and gumballs towards you.
    \item Once the tube is close enough, use the paper clips to secure the tube to the edge of the table.
    \item Use the rubber bands to secure the tube to the cane.
    \item Pull the cane towards you, using the string, and the gumballs will be transferred to the empty bowl.
\end{enumerate}
}

\end{quote}

\subsection{Solution to gumball problem following magic staff story}
\label{gumball_prob_magic_staff_story}

The following contains GPT-3's proposed solution to the gumball problem when first presented with the `magic staff story'. This story involves a magical genie who has moved from his old home (a bottle) to a new home (a different bottle), and wants to transport his collection of jewels between the bottles. To do so, he uses a magic staff to pull the new bottle over next to the old bottle and transfers the jewels by hand. When first presented with this story, children most often identify a solution to the gumball problem that involves using the aluminum walking cane to pull the empty bowl over next to the bowl with the gumballs and then transferring the gumballs by hand. GPT-3 was not able to discover this solution, but was able to identify the high-level analogy between the source story and the target problem (though not the specific analogy between the magic staff and the walking cane).

\begin{quote}

\noindent Source story:

\vspace{3mm}

\noindent Once upon a time there lived a magical genie. He was a very old, wise, and rich genie indeed. One day while he was polishing his home, which was actually a bottle, he decided he would like to find an even bigger and better home to live in. So he began searching far and wide for another bottle. Finally he found the perfect home. It was larger, prettier, and not too far away from his old bottle. The genie was very excited and began moving his belongings right away. But now the genie had a problem. He had a great many beautiful and very precious jewels in his old home. He had to somehow get all the jewels from his old bottle to the new bottle without dropping or losing a single jewel. 

\vspace{3mm}

\noindent After thinking a bit, the genie came up with a wonderful idea. He began searching for his magic staff, or wand. He then commanded his staff to stretch itself from his old home to his new home. Next, the genie tugged and pulled on his magical staff until at last he pulled the new bottle right up next to his old bottle. At once, the genie began gathering his jewels together in his old home and simply dropped them carefully into his new home right next to him. When all his jewels were safely tucked away in his new home, the genie settled in happily. He invited his friend to come in and admire his new home. I’m sure you can still find the genie sitting in his new bigger and better bottle with all his jewels and smiling contentedly even today! 

\vspace{3mm}

\noindent Target problem:

\vspace{3mm}

\noindent You are seated in front of a table. Two bowls are on the table, one within your reach and one farther away. The closer bowl contains a number of small gumballs, and the farther one is empty. Also on the table are an aluminum walking cane, a large rectangular sheet of heavy paper (posterboard), a hollow cardboard tube long enough to reach the farther bowl, scissors, string, tape, paper clips, and rubber bands. Using the materials provided, how can you transfer the balls from the filled to the empty bowl without leaving your seat? 

\vspace{3mm}

\noindent Solution:

\vspace{3mm}

\noindent \textbf{Take the cardboard tube and cut it in half. Place one half of the tube in the bowl with the gumballs and the other half in the empty bowl. Secure the two halves together with the rubber bands. Take the posterboard and cut it into strips. Tape the strips together to form a long rope. Tie one end of the rope to the handle of the walking cane and the other end to the rubber bands. Place the walking cane on the table and hold it in place. Gently pull the cane towards you, and the gumballs will travel through the tube and into the empty bowl. When all the gumballs have been transferred, remove the tube and rope and enjoy your gumballs!}

\vspace{3mm}

\noindent Can you identify an analogy between the source story and the target problem?

\vspace{3mm}

\noindent \textbf{The source story and the target problem both involve transferring something from one place to another. In the source story, the genie is transferring his jewels from one bottle to another, and in the target problem, the gumballs are being transferred from one bowl to another. In both cases, the characters must use creative problem-solving to figure out how to move the items without losing any.}

\vspace{3mm}

\noindent Does this analogy suggest a solution to the target problem? 

\vspace{3mm}

\noindent \textbf{Yes, the analogy suggests a solution to the target problem. The genie used a magical staff to stretch from one bottle to the other, and the target problem suggests using a rope and a walking cane to transfer the gumballs from one bowl to the other.}

\vspace{3mm}

\noindent Who or what in the target problem is most analogous to the genie's old home in the source story? 

\vspace{3mm}

\noindent \textbf{The bowl with the gumballs is most analogous to the genie's old home in the source story. Both contain something valuable that needs to be transferred to a new location.}

\vspace{3mm}

\noindent Who or what in the target problem is most analogous to the genie's new home in the source story? 

\vspace{3mm}

\noindent \textbf{The empty bowl is most analogous to the genie's new home in the source story. Both are the destination for something valuable that needs to be transferred from an old location.}

\vspace{3mm}

\noindent Who or what in the target problem is most analogous to the jewels in the source story? 

\vspace{3mm}

\noindent \textbf{The gumballs are most analogous to the jewels in the source story. Both are valuable items that need to be transferred from one place to another.}

\vspace{3mm}

\noindent Who or what in the target problem is most analogous to the magic staff in the source story? 

\vspace{3mm}

\noindent \textbf{The rope and walking cane are most analogous to the magic staff in the source story. Both are used to transfer something valuable from one place to another.}

\end{quote}

\subsection{Solution to gumball problem following magic carpet story}
\label{gumball_prob_magic_carpet_story}

The following contains GPT-3's proposed solution to the gumball problem when first presented with the `magic carpet story'. This story has the same setup as the magic staff story -- a magical genie wants to transport his jewels from his old bottle to his new bottle -- but the genie's solution in this story is to roll up his magic carpet and use it as a bridge between the two bottles. When first presented with this story, children most often identify a solution to the gumball problem that involves rolling up the posterboard and rolling the gumballs through it into the empty bowl. It is interesting to note that children do this despite having access to an already formed cardboard tube, likely due to the analogy between the rolled up posterboard and the rolled up magic carpet. GPT-3 identified essentially this same solution, but added some unnecessary steps, including first cutting the posterboard in half. GPT-3 correctly identified the analogy between the source story and the target problem.

\begin{quote}

\noindent Source story:

\vspace{3mm}

\noindent Once upon a time there lived a magical genie. He was a very old, wise, and rich genie indeed. One day while he was polishing his home, which was actually a bottle, he decided he would like to find an even bigger and better home to live in. So he began searching far and wide for another bottle. Finally he found the perfect home. It was larger, prettier, and not too far away from his old bottle. The genie was very excited and began moving his belongings right away. But now the genie had a problem. He had a great many beautiful and very precious jewels in his old home. He had to somehow get all the jewels from his old bottle to the new bottle without dropping or losing a single jewel. 

\vspace{3mm}

\noindent After thinking a bit, the genie came up with a wonderful idea. He searched for his magic carpet. Then he commanded it to roll itself up into a long hollow tube. Next the genie commanded his flying carpet to place one end at his old home and the other end at his new home so that it formed a sort of hollow bridge between the two bottles. Then, the genie very carefully took one jewel from inside his old home and placed it into the opening of his carpet. At once, the jewel began tumbling and rolling through the carpet tube until it reached his new home and plopped safely inside. The genie grinned happily and began rolling all his jewels through the carpet into his new home. In fact, I’m sure you can still find him sitting in his new, bigger and better bottle with all his jewels and smiling contentedly even today! 

\vspace{3mm}

\noindent Target problem:

\vspace{3mm}

\noindent You are seated in front of a table. Two bowls are on the table, one within your reach and one farther away. The closer bowl contains a number of small gumballs, and the farther one is empty. Also on the table are an aluminum walking cane, a large rectangular sheet of heavy paper (posterboard), a hollow cardboard tube long enough to reach the farther bowl, scissors, string, tape, paper clips, and rubber bands. Using the materials provided, how can you transfer the balls from the filled to the empty bowl without leaving your seat? 

\vspace{3mm}

\noindent Solution:

\vspace{3mm}

\noindent \textbf{Take the posterboard and cut it into two pieces. Take one of the pieces and roll it up into a tube. Secure the tube with the rubber bands, tape, and paper clips. Place one end of the tube into the bowl with the gumballs and the other end into the empty bowl. Then, use the walking cane to push the gumballs through the tube until they reach the empty bowl.}

\vspace{3mm}

\noindent Can you identify an analogy between the source story and the target problem?

\vspace{3mm}

\noindent \textbf{The source story and the target problem both involve transferring items from one place to another. In the source story, the genie transfers jewels from one bottle to another using a flying carpet. In the target problem, the person transfers gumballs from one bowl to another using a posterboard tube. In both cases, the items are transferred without leaving the original location.}

\vspace{3mm}

\noindent Does this analogy suggest a solution to the target problem? 

\vspace{3mm}

\noindent \textbf{Yes, the analogy suggests a solution to the target problem. The solution is to use a posterboard tube to transfer the gumballs from one bowl to the other. This is similar to how the genie used a flying carpet to transfer the jewels from one bottle to the other.}

\vspace{3mm}

\noindent Who or what in the target problem is most analogous to the genie's old home in the source story? 

\vspace{3mm}

\noindent \textbf{The bowl with the gumballs is most analogous to the genie's old home in the source story. Both contain items that need to be transferred to a new location.}

\vspace{3mm}

\noindent Who or what in the target problem is most analogous to the genie's new home in the source story? 

\vspace{3mm}

\noindent \textbf{The empty bowl is most analogous to the genie's new home in the source story. Both are the destination for items that need to be transferred from an old location.}

\vspace{3mm}

\noindent Who or what in the target problem is most analogous to the jewels in the source story? 

\vspace{3mm}

\noindent \textbf{The gumballs in the target problem are most analogous to the jewels in the source story. Both are items that need to be transferred from one location to another.}

\vspace{3mm}

\noindent Who or what in the target problem is most analogous to the magic staff in the source story? 

\vspace{3mm}

\noindent \textbf{The posterboard tube in the target problem is most analogous to the magic carpet in the source story. Both are used to transfer items from one location to another.}

\end{quote}


\end{document}